\documentclass{article}
\usepackage{iclr2027_conference}
\usepackage{times}
\iclrfinalcopy 
\usepackage{amsmath,amssymb,amsthm,mathtools,booktabs,graphicx,array,caption}
\usepackage{makecell}
\usepackage[inline]{enumitem}
\usepackage{import}
\usepackage{xifthen}
\usepackage{pdfpages}
\usepackage{transparent}
\usepackage{xcolor}
\usepackage{float}
\usepackage{microtype}
\usepackage[hidelinks]{hyperref}

\usepackage{array}
\usepackage{booktabs}

\numberwithin{equation}{section}
\newtheorem{theorem}{Theorem}[section]
\newtheorem{proposition}[theorem]{Proposition}
\newtheorem{corollary}[theorem]{Corollary}
\newtheorem{lemma}[theorem]{Lemma}

\newcommand{\N}{\mathcal N}
\newcommand{\R}{\mathbb{R}}
\newcommand{\E}{\mathbb{E}}

\newcommand{\dd}{\,\mathrm d}
\providecommand{\PP}{\mathcal P}
\providecommand{\norm}[1]{\left\|#1\right\|}
\providecommand{\Id}{\operatorname{Id}}
\providecommand{\Lip}{\operatorname{Lip}}
\providecommand{\Cov}{\operatorname{Cov}}
\providecommand{\tr}{\operatorname{tr}}
\providecommand{\D}{\mathfrak D}

\newcommand{\euler}{\texttt{euler}}
\newcommand{\sqg}{\texttt{sqg}}
\newcommand{\waves}{\texttt{het-waves}}
\newcommand{\scalar}{\texttt{passive-scalar}}
\newcommand{\nlkg}{\texttt{nonlinear-kg}}

\graphicspath{{figs/}}

\title{Generative Forecasting from Coarse \newline Observations: Does Cascading Help?}

\author{Alexandra Spitzer$^1$, Giovanni Abati$^1$, Lorenzo Baldassari$^1$, Till Muser$^1$,\\ \textbf{Cheng Shi}$^2$ \textbf{\& Ivan Dokmani\'c}$^1$\\
$^1$Department of Mathematics and Computer Science, University of Basel, Switzerland\\
$^2$D\'epartement d'Informatique, \'Ecole Normale Sup\'erieure, PSL University, Paris, France}

\begin{document}
\maketitle

\begin{abstract}
We study probabilistic forecasting of dynamical systems such as fluids and waves from coarse observations. Since the unseen fine scales influence the dynamics, the same coarse initial state can evolve into a whole distribution of futures. Generative forecasters may produce visually plausible samples, but it is hard to tell whether they sample the right conditional distribution, since trajectory data contain a single future for each initial state. We address this problem by generating reference data in a hierarchical way: for each coarse initial state, we generate many different fine-scale components and numerically evolve the full states. 
Using these conditional ensembles, we show that standard
diffusion models and stochastic interpolants can make
substantial distributional errors when trained over the
full forecast horizon with finite data and model capacity,
despite being exact in principle.
We then show that cascading shorter forecasts greatly reduces these errors for the turbulent flows and waves we study, but we also find and characterize cases in which it hurts. Since the coarse process is not Markov, cascading discards information about the unresolved fine scales even with exact transitions. To understand this, we propose a solvable analytic model which shows that the optimal cascade depth decreases as the coupling between the unresolved and observed scales gets stronger, and increases as the dependence of the forecast on the observation becomes more oscillatory. Consistent with our experiments, cascading is detrimental when the direct forecast is ``easy'' to learn, for example Gaussian as in the passive scalar case. We release our datasets as a first benchmark for probabilistic forecasting with ground-truth conditional ensembles.
\end{abstract}

\section{Introduction}
\label{sec:introduction}

Suppose we observe a turbulent flow on a coarse grid and want to predict what it will look like some time later. The grid resolves the large scales, but the unobserved fine (subgrid) scales also influence the dynamics. In 1969, Lorenz argued that in a flow with many scales, a fine-scale error propagates to ever larger ones. Two states that differ only in smallest details become as different as two random  states, in finite time~\citep{l69,palmer2014real}. Weather forecasting is a classic example of this ``butterfly effect''~\citep{l63}. The same coarse observation $X_0$ is therefore consistent with many futures $X_T$, one for each subgrid realization, and the forecast is the conditional distribution $P(X_T\mid X_0)$~\citep{chorin2000optimalb,chorin2002optimal}.

A modern approach is to learn $P(X_T \mid X_0)$ with conditional generative models such as diffusion models~\citep{song2021scorebasedgenerativemodelingstochastic} and stochastic interpolants (SIs)~\citep{albergo2023building,albergo2025stochastic}\footnote{A more detailed discussion is in Appendix~\ref{app:related}.} trained on pairs of initial and final observed states. These models generate samples of future weather and turbulent flows~\citep{Price2025,li2023seeds,pmlr-v235-chen24n,kohl2026benchmarking} which resemble real physical states, but do they actually sample from the right conditional distribution~\citep{Theis2016a}? Testing this is a challenging problem. One reason is that the conditional distribution describes all futures consistent with a given observation but available data usually only contain one trajectory per initial state.

Here we address this challenge. We design an experiment that lets us show that common generative forecasters can make substantial distributional mistakes. We then study a way to reduce these mistakes by better using the available data: forecasting over shorter time intervals and composing the results. Practical forecasters already do this, but mostly as a matter of convenience. Our goal is to understand when and why it helps, and when it can actually hurt.

The experiment works as follows. For five spatio--temporal systems with different dynamical characteristics we first generate coarse states and then complete each coarse state with many independently drawn fine-scale components (Section~\ref{sec:problem}). The full states are identical under coarse graining, but the  dynamics (computed by a numerical solver) evolve them into different coarse futures. For each coarse initial state this yields a reference ensemble from the conditional distribution, whose statistical properties can be studied empirically. We test a standard conditional SI and a conditional diffusion model trained directly on initial--final pairs $(X_0,X_T)$, and find that they can be substantially wrong. \autoref{fig:failure} shows an example for Euler flow (second and third rows). Their samples have visually correct local statistics, but the global structure is completely wrong.

Diffusion models and SIs generate samples via stochastic dynamics in an auxiliary ``diffusion'' time. This dynamics is in general very different from the physical dynamics of the system over the forecasting period. A forecaster trained only on pairs $(X_0,X_T)$ has to learn in one fell swoop everything the physical dynamics does between $0$ and $T$. But the physical dynamics is a composition of many short steps. As intermediate observations at times $0<t<T$ are often available~\citep{cachay2023dyffusion,rozet2023scorebaseddataassimilation}, it is natural to ask whether they can be leveraged to improve conditional forecasting. Instead of learning $P(X_T\mid X_0)$ directly, one can learn the conditional transitions $P(X_{t_{i+1}}\mid X_{t_i})$ over $S$ shorter intervals and compose them into a \emph{cascade} of depth $S$. In a cascade of SIs, the generative path is then forced to agree with the physical evolution at some intermediate times. In the Euler example, such a cascaded SI approach is much better than direct and it effectively reproduces the conditional distribution (last row of~\autoref{fig:failure}).

In practice, the depth of a cascade is set by a fixed time step: GenCast rolls out a twelve-hour transition over fifteen days~\citep{Price2025}, and conditional diffusion models and SIs for fluid simulation use the time step of the training data~\citep{pmlr-v235-chen24n,kohl2026benchmarking,MUCKE2026107098}.
Cascading can, however, also discard information even with exact transitions. Only the full state evolves by composing short steps. After coarse graining, the unresolved fine scales carry information from one interval to the next, so that the coarse process is not Markov. To our knowledge, there is no analysis of how cascading affects the learned distribution. The questions we ask here are: \emph{is it better to forecast directly or with a cascade, what properties of the physical system determine the answer, and how deep should a cascade be?}

We propose a solvable model to answer these questions (Section~\ref{sec:cascade}). The optimal cascade depth results from the interplay between easier learning and memory loss, and depends on the specific system under study. If the conditional distribution is easy to learn over the full horizon, cascading can be detrimental. This is the case for a scalar field passively advected by a steady flow, whose conditional distribution stays Gaussian at every horizon, with a mean affine in the observation and a covariance that does not depend on it (Section~\ref{sec:results}).

Our main contributions are the following.
\begin{itemize}[leftmargin=*,itemsep=2pt,topsep=2pt]
\item \emph{An experimental design that enables proper evaluation of the conditional distribution.} New datasets for five physical systems, with different simulated futures for each observed initial state, allow us to compare learned and true conditional distributions on every observation (Sections~\ref{sec:problem} and~\ref{sec:design}). To our knowledge, these are the first datasets designed for this notoriously challenging problem; we release them as an openly accessible benchmark. We find that diffusion models and SIs trained directly over the forecast horizon sample substantially wrong conditional distributions in four of the five systems, despite producing visually plausible samples.

\item \emph{A theoretical explanation.} For smooth dynamics, we show that, through nonlinear interactions, the conditional mean leaves a fixed approximation class as the horizon grows, and that, under partial observation, a fine cascade of exact transitions loses the conditional spread of the forecast. We construct a stylized model of a rotating system driven by a hidden Ornstein--Uhlenbeck process where we compute the error of a simple SI exactly and prove that it is minimized at a finite depth $S^\star\geq2$; with the rotation switched off, every cascade is worse than the exact direct forecast (Section~\ref{sec:cascade}).

\item \emph{An experimental test.} At a fixed number of supervised transitions, cascading reduces the error of the direct forecasters by factors between $2.3$ and $5.7$, with the smallest error at an intermediate depth, for nonlinear flows and linear waves alike. For the passive scalar, every added stage increases the error, consistent with the analysis (Section~\ref{sec:results}).
\end{itemize}

\section{Conditional forecasting from coarse observations}
\label{sec:problem}

We consider a physical system whose full state $Y_t$ evolves according to $Y_t=\mathcal G_t(Y_0)$. A forecaster sees only the coarse state $X_t=\mathcal H(Y_t)$. At the initial time, we write $Y_0=(X_0,Z_0)$, where $Z_0$ contains the fine-scale components. Since the observation $X_0=x_0$ generally does not determine $Z_0$, different full initial states consistent with the same observation can lead to different observed futures. The forecasting task is therefore to learn the conditional distribution
\begin{equation}
P(X_T\mid X_0=x_0)
\label{eqn:targetdist}
\end{equation}
for every observation $x_0$. In our systems, $Y_t$ consists of one or two fields on a two-dimensional periodic domain, simulated on grids ranging
from $128^2$ to $1024^2$ points. The observation map $\mathcal H$ keeps only their Fourier modes with $0<|k|\leq8$ (196 real coordinates per field), while
the unresolved initial modes extend to higher wavenumbers
that depend on the system (Appendix~\ref{app:datasets}).

\textbf{Why trajectory data are not enough.}
A generative forecaster learns an approximation $\widehat P(X_T\mid X_0=x_0)$ from which it can draw samples. To measure its accuracy, we would like to compare the learned and the true conditional distributions for each observation, using some metric $D$ on probability distributions, and then average over observations,
\begin{equation}
\mathcal E(\widehat P)=\E_{x_0\sim P(X_0)}\Bigl[D\bigl(P(X_T\mid X_0=x_0),\,\widehat P(X_T\mid X_0=x_0)\bigr)\Bigr].
\label{eq:conditional_error}
\end{equation}
Estimating $\mathcal E$ requires many samples from $P(X_T\mid X_0=x_0)$ for each of many observations $x_0$. But data usually contain only one future per observed initial state, with all observed states being different. Proper scoring rules can rank forecasters on such data~\citep{Gneiting01032007}, but they cannot determine how far they are from the true conditional. Averaging final spectra or other statistics over initial conditions only tests the marginal distribution $P(X_T)$. Comparing the joint distributions of $(X_0,X_T)$ could in principle work because they determine the conditionals. The issue is that a forecaster that generates the ``correct'' forecast for a slightly perturbed observation would have a nearly correct joint distribution, but the forecasts could be completely wrong if they are sensitive to the initial state (which is often the interesting regime).

\begin{figure}[t]
\centering
\includegraphics[width=0.7\linewidth, trim=0 2.3mm 0 1.5mm,clip]{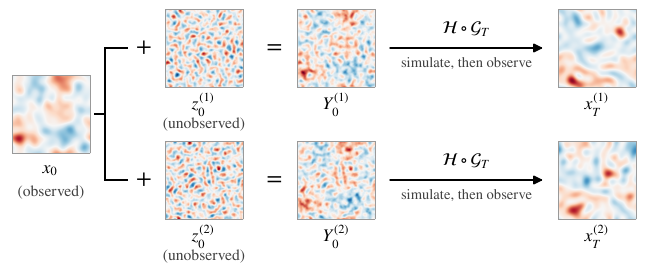}
\caption{\textbf{Hierarchical data generation.} Many unobserved fine initial states $z_0^{(i)}$ are combined with the same observed coarse state $x_0$. The solver evolves each full initial state $Y_0^{(i)}$, and observing the result at time $T$ gives a reference ensemble $x_T^{(1)},x_T^{(2)},\ldots$ for $P(X_T\mid X_0=x_0)$.}
\label{fig:method}
\end{figure}

\textbf{Hierarchical data generation.}
When the system can be simulated, this problem can be addressed by sampling the initial data hierarchically. For each conditioning ensemble $e$, we first sample the coarse scales $x_0^{(e)}\sim P(X_0)$, and then many different fine scales for this coarse state,
\begin{equation}
z_0^{(e,i)}\sim P\bigl(Z_0\mid X_0=x_0^{(e)}\bigr),\qquad Y_0^{(e,i)}=\bigl(x_0^{(e)},z_0^{(e,i)}\bigr),\qquad i=1,\ldots,n.
\label{eq:hierarchical}
\end{equation}
A numerical solver sees the full initial state, coarse and fine, and evolves it into (generally) diverging future states, because the fine scales influence the evolution of the coarse ones. But the forecaster only sees the coarse scales, which are the same for every member of the ensemble. The observed endpoints $X_T^{(e,i)}=\mathcal H(\mathcal G_T(Y_0^{(e,i)}))$ are thus independent samples from $P(X_T\mid X_0=x_0^{(e)})$, and they form a reference ensemble against which any forecast from $x_0^{(e)}$ can be compared (Figure~\ref{fig:method}). Each conditioning ensemble belongs as a whole to the training, validation, or test split, so every test observation is new to the forecaster.

\begin{figure}[t]
\centering
\includegraphics[width=0.7\linewidth,trim=0 1.5mm 0 1mm,clip]{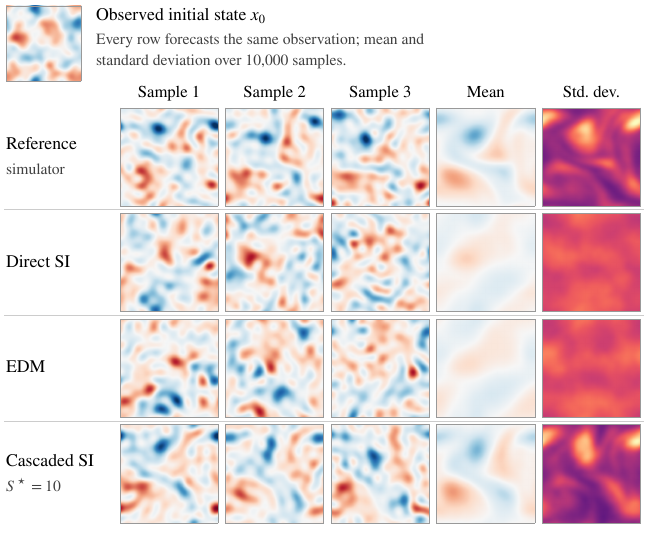}
\caption{\textbf{Direct forecasters can sample the wrong conditional distribution.} One held-out Euler observation (top left). Each row shows three samples of the vorticity at time $T$, followed by the conditional mean and the pointwise standard deviation over $10{,}000$ samples. Reference: the simulator, run from $10{,}000$ fine-scale completions of the observation. Direct SI and EDM: generative forecasters trained on pairs $(X_0,X_T)$. Cascaded SI: the cascaded forecaster of Section~\ref{sec:cascade} at the validation-selected depth $S^\star=10$, discussed in Section~\ref{sec:results}.}
\label{fig:failure}
\end{figure}

\textbf{Direct forecasters can sample the wrong distribution.}
Using our reference ensembles, we can check whether a learned forecaster samples the right conditional distribution. This leads to our first major finding. Figure~\ref{fig:failure} shows results for a two-dimensional Euler flow, where the forecaster observes a coarse-grained initial vorticity. We trained two strong generative forecasters on pairs $(X_0,X_T)$ from the training ensembles: a conditional SI~\citep{pmlr-v235-chen24n}, which we call Direct SI, and a conditional elucidated diffusion model (EDM)~\citep{NEURIPS2022_a98846e9}. We then drew $10{,}000$ samples from each model for a held-out observation. We find that both models generate plausible individual samples. However, a closer look reveals that the large-scale structure of their forecasts is wrong. Indeed, the reference mean has a distinct large-scale geometry, and the reference standard deviation is strongly heterogeneous, while the means of the two forecasters are ``washed out'', and their standard deviations are nearly uniform. The same happens for waves in heterogeneous media (Appendix~\ref{app:qualitative}).

\section{Direct and cascaded forecasts}
\label{sec:cascade}

As in most forecasting datasets, our datasets contain observations at intermediate times $0 < t < T$. A direct forecaster does not use them: it learns $P(X_T\mid X_0)$ from the pairs $(X_0,X_T)$ alone. A cascaded forecaster divides the forecast horizon into $S$ intervals, with $t_i=iT/S$, and learns the conditional distributions $K_i(x,\cdot)=P(X_{t_{i+1}}\in\cdot\mid X_{t_i}=x)$ of the transitions between consecutive observed states, for $i=0,\ldots,S-1$. Let $\widehat K_i$ denote the learned approximation of $K_i$. A cascaded forecast from $x_0$ applies these approximations in turn: starting from $\widehat X_0=x_0$, it samples
\begin{equation}
\widehat X_{t_{i+1}}\sim\widehat K_i\bigl(\widehat X_{t_i},\cdot\bigr),\qquad i=0,\ldots,S-1.
\label{eq:cascaded_forecast}
\end{equation}
Each transition is thus conditioned only on the state generated by the previous one, and the forecast $\widehat X_T$ is distributed according to the composition $\widehat K_0\widehat K_1\cdots\widehat K_{S-1}$ of the learned transitions. We implement both forecasters with stochastic interpolants (Section~\ref{sec:design}). The direct one is the Direct SI of Figure~\ref{fig:failure}, and the cascaded one is Cascaded SI. The two coincide for $S=1$.

Consider first an ideal setting, in which the model class of the forecaster contains the true conditional distributions, training reaches the population optimum, and sampling is exact. In this setting, Direct SI samples $P(X_T\mid X_0)$ exactly, even though the state is only partially observed, since~\eqref{eqn:targetdist} is a well-defined conditional distribution. Cascaded SI need not be exact even in this ideal setting, because the composition $K_0K_1\cdots K_{S-1}$ of the exact transitions generally differs from $P(X_T\mid X_0)$. In practice, however, the transitions are learned from finite data with a model of finite capacity, which is why shorter transitions can help. We analyze the two effects in turn and then show how they trade off.

\textbf{Why can shorter transitions be easier to learn?}
\label{sec:approximation}
Over a long horizon, the future observation can depend on the initial one in a nonlinear or oscillatory way, which a model of finite capacity may not be able to represent. Over a short horizon this dependence may be simpler. We make this precise for the conditional mean $m_t(x)=\E[X_t\mid X_0=x]$, since an error in the conditional mean implies an error in the distribution (Proposition~\ref{prop:mean-endpoint}). We model limited capacity by a fixed class $\mathcal F$ of functions of the observation, such as the polynomials of a given degree. For smooth dynamics, the distance from $m_t$ to $\mathcal F$ grows like $c_nt^n/n!$ for short horizons, where $t^n$ is the first power in the expansion of $m_t$ whose coefficient lies outside $\mathcal F$ (Theorem~\ref{thm:first-escape}). For quadratic dynamics, such as Galerkin truncations of the Euler equations, and quadratic $\mathcal F$, this coefficient is cubic in the observation and appears at order $t^2$ (Theorem~\ref{thm:coarse-cubic}). Repeated nonlinear interactions then create a dependence that a quadratic map cannot represent.

It can however be represented by a composition of short maps from the same class. When the full state is observed, composing $S$ fitted polynomial maps of a fixed degree over intervals of length $\delta=T/S$ converges to the true map over $[0,T]$ as $\delta\to0$, whereas a single fitted polynomial keeps a positive error (Theorem~\ref{thm:composition}).

\textbf{Why can too many short transitions hurt?}
\label{sec:memory}
Even when the short transitions are exact, they can incur a memory loss. Along a true physical trajectory, the fine scales at time $t_i$ are shaped by the unobserved initial state and by the evolution up to $t_i$, and they continue to act on the coarse scales after $t_i$. The earlier observation $X_0$ therefore has more information about the fine scales at time $t_i$ than the current observation $X_{t_i}$ and the distribution of $X_{t_{i+1}}$ given $X_{t_i}$ changes when we also condition on $X_0$. In other words, the observed process is not Markov. A cascade replaces the actual fine scales, at each of its $S$ stages, by a fresh draw from those compatible with the current observation. This loss of memory under coarse graining is classical in optimal prediction and in the Mori--Zwanzig formalism, where the history of the observed variables enters their reduced dynamics as a memory term~\citep{chorin2000optimalb,chorin2002optimal,ma2019}.

An exact transition maps the distribution of $X_{t_i}$ to that of $X_{t_{i+1}}$, so a cascade of exact transitions, averaged over $X_0$, produces the correct distribution of the state at every time of the partition. What can be wrong is the dependence of the forecast on $X_0$ (Appendix~\ref{app:forgets}). This cannot be detected by pooling statistics over initial conditions, such as energy spectra (Section~\ref{sec:problem}). For smooth deterministic dynamics, a cascade of exact transitions converges, as the transitions shrink, to a deterministic flow driven by the conditional mean velocity of the observed state, whose error is at least the conditional standard deviation of the true forecast (Theorem~\ref{thm:memory-limit}). To see why, consider the size of the terms. Over a transition of length $\delta$, the unobserved components push the observed state by an amount of order $\delta$. Along a true trajectory, these pushes persist and add up to a spread of order $T$, whereas the cascade draws them anew at each stage, so that their fluctuations average out and only their mean survives. Retaining the initial observation $X_0$ at every stage
eliminates this memory error for exact transitions
(Proposition~\ref{prop:anchoring}). However, the resulting
transitions may be harder to learn, as we show both
theoretically and experimentally in the rotation--OU model
(Appendices~\ref{app:anchored-ou-theory}--\ref{app:ou-anchoring-experiment}).

\textbf{Why can an intermediate depth be best?}
\label{sec:theory}
The two effects imply a trade-off. To analyze it, we construct a model in which the target distribution, the distribution generated by a specific SI at every depth, and the distance between them can all be computed. The model has two parameters that can be varied independently: $\omega$ sets how complicated the dependence of the forecast on the observation is, and $\rho$ sets how much of the uncertainty is coupled to the observed state.

\label{sec:rotation-model}
We consider a two-dimensional observed state $Y_t\in\R^2$ that rotates with an angular velocity controlled by an observed variable $c\sim\operatorname{Unif}[-1,1]$, and that is driven in addition by two sources of uncertainty: an unobserved Ornstein--Uhlenbeck (OU) process $H_t\in\R^2$, which is correlated over time and plays the role of the unobserved fine scales, and independent white noise. With independent $Y_0\sim\N(0,s_0^2I_2)$ and $H_0\sim\N(0,I_2)$ and independent Brownian motions $B$ and $W$, we consider
\begin{equation}
\begin{aligned}
\dd H_t &= -\kappa H_t\dd t+\sqrt{2\kappa}\dd B_t,\\
\dd Y_t &= \bigl(\omega cJY_t+\beta_\rho R(\omega ct)H_t\bigr)\dd t+\sigma_\rho\dd W_t.
\end{aligned}
\label{eq:ou-dynamics}
\end{equation}
where $J=\left(\begin{smallmatrix}0&-1\\1&0\end{smallmatrix}\right)$ and $R(\theta)=e^{\theta J}$ is the rotation by $\theta$. Without the OU forcing and the white noise, the rotation drift alone would carry $Y_0=y$ to $Y_T=R(\omega cT)y$. As $\omega$ increases, the map $c\mapsto R(\omega cT)y$ becomes more oscillatory and harder to approximate. The hidden process $H_t$ is stationary and forgets its value at rate $\kappa$. Its values therefore remain correlated across consecutive transitions, and this correlation is discarded by a cascade conditioned only on $(c,Y_{t_i})$.

The strengths $\beta_\rho$ and $\sigma_\rho$ are chosen so that, for fixed $q>0$ and $0\leq\rho<1$, the hidden OU process contributes $\rho qI_2$ to the covariance at time $T$ and the white noise contributes $(1-\rho)qI_2$ (Appendix~\ref{sec:rotation-ou-tradeoff}). Expressing the dynamics in the rotating coordinates $R(-\omega ct)Y_t$ then gives
\begin{equation}
Y_T\mid(Y_0=y,c)\sim\N\bigl(R(\omega cT)y,\,qI_2\bigr)=:K_T^{\mathrm{OU}}(y,c).
\label{eq:ou-target}
\end{equation}
The target conditional distribution is thus the same for every $\rho$: changing $\rho$ changes how uncertainty is carried through time, but not the conditional distribution at time $T$. This lets us study how hidden-state memory affects cascading without also changing the difficulty of the direct forecasting problem.

Over one transition of length $\delta=T/S$, the exact conditional distribution is Gaussian, $Y_{t_{i+1}}\mid(Y_{t_i}=y,c)\sim\N(r_iR(\omega c\delta)y,\,q_iI_2)$, with scalars $r_i$ and $q_i$ determined by the model. We compare Direct SI and Cascaded SI for an SI whose only approximation is in the conditional mean. It keeps the exact Gaussian form of each transition, with $r_i$ and $q_i$ known, and approximates the rotation $R(\omega c\delta)y$ by a polynomial $p(c,y)$ of total degree at most two, which allows us to compute the approximation error explicitly. For this family, minimizing the SI training loss amounts to fitting $p$ by least squares, and exact sampling generates $\N(r_ip(c,y),q_iI_2)$ (Proposition~\ref{prop:gaussian-transfer}). Direct SI fits the full rotation with one such polynomial; Cascaded SI fits the rotation over one transition and composes $S$ fitted transitions. Since $y$ is centered and isotropic, the best fit has the form $M(c)y$ with $M$ affine in $c$, so the conditional mean of the cascade, $A_SM(c)^Sy$ with $A_S=\prod_ir_i$, is a polynomial of degree $S$ in $c$.

\emph{A cascade can therefore represent a more complicated dependence on $c$, even though each individual transition uses the same quadratic approximation class.} In our experiments, we follow the same idea. Rather than increasing the capacity of the neural network, we learn shorter transitions and compose them.

We measure the error of the generated distribution $\widehat K_S(y,c)$ by~\eqref{eq:conditional_error} with the squared Wasserstein-2 metric $D=W_2^2$, that is, by $\mathcal E_S(\rho)=\E_{Y_0,c}\,W_2^2(K_T^{\mathrm{OU}}(Y_0,c),\widehat K_S(Y_0,c))$.

\label{sec:optimal-depth}
For Direct SI, the error is entirely due to approximation, $\mathcal E_1(\rho)=2s_0^2[1-j_0(\omega T)^2-3j_1(\omega T)^2]>0$, with $j_0(\nu)=\sin\nu/\nu$ and $j_1(\nu)=(\sin\nu-\nu\cos\nu)/\nu^2$, and it does not depend on $\rho$. As $S\to\infty$, the fitted rotations become exact, and only the memory error remains. We denote its limit by $\mathcal C_{\rho,\infty}$ (Eq.~\eqref{eq:rotation-ou-fine-limit}). $\mathcal C_{\rho,\infty}$ is a consequence of the parameter $\rho$, the fraction of the endpoint variance carried by the persistent hidden process. It vanishes at $\rho=0$ and grows like $\rho^2$ for small $\rho$.

\begin{theorem}
\label{thm:intermediate}
Fix $T,\kappa,q,s_0>0$ and $\omega\ne0$ in~\eqref{eq:ou-dynamics}. For the SI model defined above, which approximates the transition mean with a quadratic function:
\begin{enumerate}[itemsep=0pt,topsep=2pt]
\item If $\rho=0$, then $\mathcal E_S(0)=O(S^{-2})$ as $S\to\infty$.
\item For every fixed $0<\rho<1$, there is a $\Lambda_\rho>0$ such that $\mathcal E_S(\rho)=\mathcal C_{\rho,\infty}-\Lambda_\rho T/S+O(S^{-2})$.
\item There is a $\rho_*>0$ such that, for each $0<\rho<\rho_*$, a global minimizer over positive integer depths exists at a finite $S^\star\geq2$, and $\mathcal E_{S^\star}(\rho)<\min\{\mathcal E_1(\rho),\mathcal C_{\rho,\infty}\}$.
\end{enumerate}
\end{theorem}

The theorem shows that these two effects can produce a finite optimal cascade depth: without hidden forcing, deep cascades become exact, whereas with it, the asymptotic error is nonzero. We further prove that, as $\rho\downarrow0$, $\rho^2S^\star(\rho)$ converges to a constant proportional to $\omega^4$ (Theorem~\ref{thm:rotation-ou-scaling}): the weaker the contribution of the hidden process and the faster the rotation, the deeper the optimal cascade. Without the rotation ($\omega=0$), the quadratic fit is exact, and so is Direct SI; for $\rho>0$, any cascade performs worse, increasingly so with depth in our evaluations (Appendix~\ref{sec:rotation-ou-tradeoff}).

\textbf{Connection between analysis and experiment.}
This model shares two key features with the PDE experiments: the dependence of the forecast on the conditioning variable becomes more complicated as the horizon grows, and an unobserved component carries information across transitions. Its dynamics are linear in $(Y_t,H_t)$ for fixed $c$, so its behavior requires neither nonlinearity nor chaos. It suggests three things: cascading helps when the forecast depends on the observation in a complicated way; very deep cascades lose accuracy when unobserved components act on the observed ones strongly and persistently, and the stronger they act, the shallower the best cascade; and cascading does not help when the direct forecast is simple, as at $\omega=0$. In the next section we experimentally study how these two effects play out
across five PDE systems.

\section{Experimental design}
\label{sec:design}

\begin{figure}[t]
\centering
\includegraphics[width=0.7\linewidth, trim=0 1.5mm 0 2.3mm,clip]{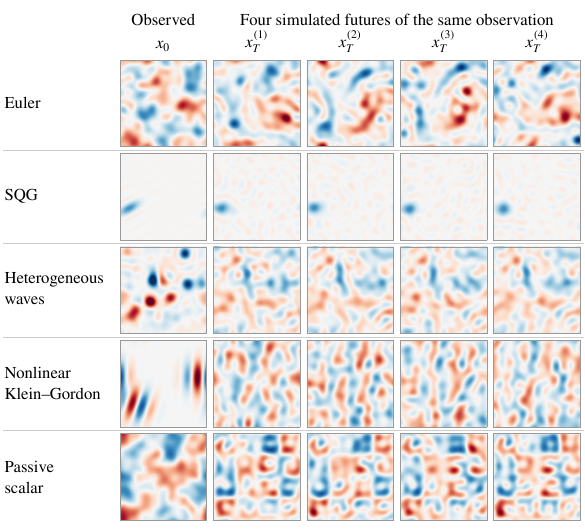}
\caption{\textbf{Experimental systems.} Each row shows one observed initial state and four simulated futures from its conditioning ensemble; the futures differ only in their unobserved initial fine scales. The fields shown are the observed vorticity, buoyancy, displacement, displacement, and concentration.}
\label{fig:dataset_overview}
\end{figure}

\textbf{The experimental dynamical systems.}
We select five physical systems to investigate conditional forecasting under different physical dynamics and forms of unobserved initial uncertainty (Figure~\ref{fig:dataset_overview} and Table~\ref{tab:systems}). They comprise nonlinear fluid dynamics, wave propagation, and linear transport. In particular, we include both nonlinear and linear dynamics to investigate whether the benefits of cascading extend beyond systems with nonlinear interactions or chaotic sensitivity to initial conditions. Appendix~\ref{app:datasets} gives the equations, initial distributions, numerical solvers, and convergence checks.

\emph{Euler} (\euler{}). The vorticity of an inviscid fluid is transported by a velocity field that is determined by the vorticity itself. The resulting nonlinear interactions couple spatial scales, so the unobserved fine-scale vorticity influences the observed coarse evolution.

\emph{Surface quasi-geostrophic flow} (\sqg{}). The surface buoyancy of a rapidly rotating, stratified fluid is transported in the same way~\citep{held:qgs}, but the velocity depends more strongly on fine-scale variations of the buoyancy than in Euler flow. The unobserved scales therefore have a stronger influence on the flow that transports the observed ones.

\emph{Heterogeneous waves} (\waves{}). Damped linear waves propagate in a medium with a fixed, spatially varying wave speed, which scatters them. The state comprises the displacement and the velocity. The observed initial displacement consists of the coarse scales of randomly placed shapes on a random background; the unobserved fine-scale displacement is confined to the interiors of the shapes, so its distribution depends on the observation, and the unobserved initial velocity consists of three Gaussian kicks at random locations. The forecaster observes the coarse displacement and a velocity proxy computed from two consecutive displacement snapshots, which is zero at $t=0$.

\emph{Nonlinear Klein--Gordon} (\nlkg{}). A scalar field undergoes dispersive wave propagation with a cubic restoring force. As in the wave system, the state comprises the field and its time derivative, but the forecaster observes the coarse scales of both. The cubic term couples Fourier modes, so the fine scales influence the coarse ones through a local interaction instead of transport.

\emph{Passive scalar} (\scalar{}). A scalar is advected by a steady cellular flow and diffuses. Its evolution is linear, and its unobserved fine scales are Gaussian and independent of the observation. The conditional distribution of the future is therefore Gaussian, with a mean that depends affinely on the observation and a covariance that does not depend on it (Proposition~\ref{prop:linear-control}). Among our systems, this is the simplest conditional distribution to learn.

\begin{table}[t]
\centering
\footnotesize
\setlength{\tabcolsep}{3pt}
\renewcommand{\arraystretch}{1.0}
\caption{\textbf{Properties of the five systems relevant for forecasting.}
``Independent'' means independent of $x_0$.
The nonlinear systems are expected to be chaotic, but we do not
show it formally or experimentally. Cascading helps when $S^\star > 1$.}
\label{tab:systems}
\begin{tabular}{@{}
  >{\raggedright\arraybackslash}m{0.135\linewidth}
  >{\raggedright\arraybackslash}m{0.185\linewidth}
  >{\centering\arraybackslash}m{0.060\linewidth}
  >{\centering\arraybackslash}m{0.080\linewidth}
  >{\raggedright\arraybackslash}m{0.190\linewidth}
  >{\raggedright\arraybackslash}m{0.200\linewidth}
  >{\centering\arraybackslash}m{0.045\linewidth}
  @{}}
\toprule
System & Dynamics & Linear & Chaotic
& Unobserved initial component
& Distribution of $X_T$ given $x_0$
& $S^\star$ \\
\midrule
Euler
& transport by own velocity
& no & expected
& normalized Gaussian, independent
& nonlinear in $x_0$
& 10 \\
SQG
& as Euler, stronger fine-scale coupling
& no & expected
& normalized Gaussian, independent
& nonlinear in $x_0$
& 10 \\
Het.\ waves
& damped waves, scattering
& yes & no
& depends on $x_0$; velocity kicks
& affine mean, cov.\ depends on $x_0$
& 5 \\
Nonlinear Klein--Gordon
& dispersive waves, cubic force
& no & expected
& normalized, independent
& nonlinear in $x_0$
& 20 \\
Passive scalar
& fixed-flow transport, diffusion
& yes & no
& Gaussian, independent
& Gaussian, affine mean, fixed cov.
& 1 \\
\bottomrule
\end{tabular}
\end{table}

\textbf{Data.}
For each system, we generate 80 training, 10 validation, and 10 test conditioning ensembles, with distinct observed initial states. Each training ensemble contains 200 trajectories stored at 41 equally spaced times to accommodate different cascade depths. An important feature of our benchmark is that the \emph{test} datasets are large: each test observation comes with $10{,}000$ reference futures, which allows statistically meaningful comparisons of conditional distributions.

\textbf{Forecasters.}
\label{sec:methods}
Direct SI learns a single conditional stochastic interpolant from $X_0$ to $X_T$~\citep{pmlr-v235-chen24n}. It connects the two states by a stochastic path in an auxiliary time $\tau\in[0,1]$, learns the drift of this path by regression, and generates a forecast by integrating the resulting stochastic differential equation from $X_0$. Cascaded SI learns the transitions from $X_{t_i}$ to $X_{t_{i+1}}$ in the same way, with one network shared by all transitions and conditioned on their physical start and end times. EDM~\citep{NEURIPS2022_a98846e9}, a strong diffusion-model baseline, is trained on the same endpoint pairs as Direct SI. For each system, all methods use the same U-Net, with channel widths $64$, $128$, and $256$ and circular convolutions, acting on a $32\times32$ reconstruction of the observed modes. Hyperparameters, checkpoints, and the cascade depth $S^\star\in\{1,2,5,10,20,40\}$ are selected on the validation ensembles by the conditional energy distance~\citep{SZEKELY20131249}, and the number of sampler steps is calibrated on validation data. The test ensembles are used only for the final evaluation (Appendix~\ref{app:training}).

\textbf{A fixed number of training constraints.}
A single trajectory provides one endpoint pair for Direct SI or EDM, but $S$ transition pairs for a cascade of depth $S$. To compare learning the forecast horizon directly with learning it through shorter transitions, we therefore fix the number of supervised pairs at $B=16{,}000$: Direct SI and EDM use one endpoint pair from each of the 200 trajectories of every training ensemble, and Cascaded SI at depth $S$ uses all $S$ transitions of $200/S$ trajectories per ensemble. Every run uses $30{,}000$ optimizer updates.  A deeper cascade receives no additional pairs or updates, it is trained on fewer distinct trajectories, and its forecasts require $S$ sequential sampler integrations. \emph{We emphasize that this is a harsh constraint.} In this way we equalize the number of training constraints but not the variety of the training data, whereas in reality the intermediate observations are often simply available. Our goal is to ensure that we focus on the tradeoff between approximation and Markovianity. On Euler, increasing the number of distinct training
observations by a factor of 10 at a fixed number of supervised
pairs improves Direct SI, but its csFID-10k remains nearly
three times that of Cascaded SI. Cascading also helps
when both forecasters are trained on a single trajectory
per initial observation
(Appendices~\ref{app:training-diversity}--\ref{app:single-trajectory}).

\textbf{Evaluation.}
For each test observation $x_0^{(e)}$, we draw $N=10{,}000$ samples from the forecaster and compare them with the $N$ reference samples of the simulator. Because a forecast can reproduce realistic local patterns at the wrong locations (Figure~\ref{fig:failure}), we use spatial FID~\citep{pmlr-v139-nash21a}. Unlike the standard FID~\citep{fid}, the features of spatial FID are sensitive to the spatial layout of the images. We verify that this makes the metric more sensitive to spatial misalignment on our data: for example, it penalizes a simple flip of the Euler fields much more strongly than standard FID (Appendix~\ref{app:sfid}).
With $(m_e,C_e)$ and $(\widehat m_e,\widehat C_e)$ the means and covariances of $\phi$ over the reference and the generated samples, we compute for each test observation the Fr\'echet distance
\begin{equation}
d_e=\|m_e-\widehat m_e\|^2+\tr\Bigl[C_e+\widehat C_e-2\bigl(C_e^{1/2}\widehat C_eC_e^{1/2}\bigr)^{1/2}\Bigr],
\label{eq:csfid}
\end{equation}
an instance of $D$ in~\eqref{eq:conditional_error}. We report its average over the ten test observations, which we call csFID-10k, with the standard error of the mean (SEM) over the test observations, for one training seed. For the two wave systems, the features are computed from the displacement (Appendices~\ref{app:evaluation}--\ref{app:sfid}).

\begin{table}[h!]
\centering\footnotesize
\setlength{\tabcolsep}{4pt}
\caption{\textbf{Conditional forecasting error.} Test csFID-10k (lower is better), mean $\pm$ SEM over the ten test observations, for one training seed. $S^\star$ is selected on validation data; for the passive scalar, $S^\star=1$, so Direct SI and Cascaded SI are the same fitted model. $^{\dagger}$The dagger signifies that the best value of one hyperparameter lay at the edge of the searched range (Appendix~\ref{app:training}).}
\begin{tabular}{@{}lcccc@{}}
\toprule
& \multicolumn{3}{c}{csFID-10k} & \\
\cmidrule(lr){2-4}
System & Direct SI & EDM & Cascaded SI & $S^\star$ \\
\midrule
Euler  & $57.26\pm3.35$   & $56.32\pm3.74$   & $\mathbf{16.32\pm1.14}$ & 10 \\
SQG    & $138.48\pm23.09$ & $118.62\pm21.67$ & $\mathbf{51.24\pm8.32}$ & 10 \\
Heterogeneous waves  & $115.86\pm6.76$  & $123.04\pm7.07$ & $\mathbf{36.66\pm1.78}$ & 5 \\
Nonlinear Klein--Gordon & $98.48\pm6.18$ & $92.37\pm5.43^{\dagger}$ & $\mathbf{16.24\pm1.55}$ & 20 \\
Passive scalar & $8.67\pm0.28$    & $\mathbf{7.88\pm0.31}$ & $8.67\pm0.28$ & 1 \\
\bottomrule
\end{tabular}
\label{tab:results}
\end{table}

\section{Results}
\label{sec:results}

\begin{figure}[t]
\centering
\includegraphics[width=0.9\linewidth, trim=0 5mm 0 2.8mm,clip]{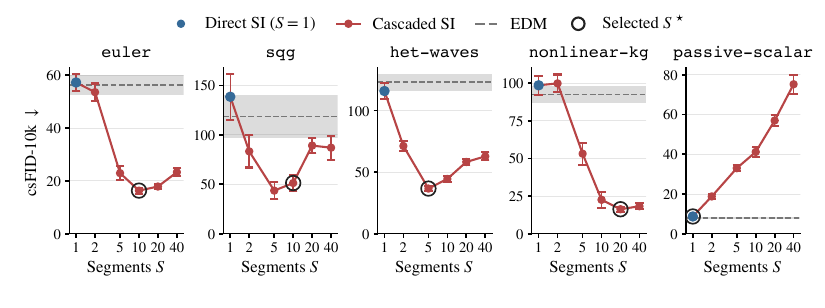}
\caption{\textbf{Intermediate-depth cascades improve conditional forecasting in four of five systems.} Test csFID-10k versus cascade depth $S$ (values in Table~\ref{tab:depth_values}). $S=1$ is Direct SI; outlined markers: validation-selected $S^\star$; dashed lines: EDM; error bars: $\pm1$ SEM over the ten test observations.}
\label{fig:depth_sweep}
\end{figure}

\textbf{Cascading recovers the conditional distribution.}
\label{sec:posterior_results}
The last row of Figure~\ref{fig:failure} shows the cascaded forecast for the same Euler observation as the other rows. The cascaded forecast correctly captures the spatial structure of the reference conditional mean and the regions of high and low standard deviation, which the direct forecasts fail to reproduce. The same holds for heterogeneous waves, whose direct forecasts are far too variable (Appendix~\ref{app:qualitative}). Table~\ref{tab:results} quantifies the improvement over all test observations. On Euler, Cascaded SI at the validation-selected depth $S^\star=10$ reduces csFID-10k by a factor of about $3.5$ relative to both Direct SI and EDM, and it is better than both on each of the ten test observations.
We make the following observations:
\begin{enumerate}[leftmargin=*,itemsep=1pt,topsep=2pt]
\item \emph{Cascading helps in all three nonlinear systems.} The improvement on Euler extends to SQG and Klein--Gordon, by factors of $2.3$ and $5.7$ relative to the better direct forecaster.
\item \emph{Neither nonlinearity nor chaos is necessary for cascading to help.} The heterogeneous-wave dynamics are linear and nonchaotic, and the conditional mean is affine in the observation, yet Cascaded SI reduces csFID-10k by more than a factor of three relative to both Direct SI and EDM. The rotation model of Section~\ref{sec:theory} is linear as well.
\item \emph{For the passive scalar, cascading is detrimental.} Validation selects $S^\star=1$, and every added stage increases the error, from $8.7$ at $S=1$ to $75$ at $S=40$ (Figure~\ref{fig:depth_sweep}); EDM achieves the lowest score. This is what the rotation model does at $\omega=0$: the direct forecast is simple, and a cascade has little to gain. The passive scalar is linear like the waves, so linearity alone does not decide the outcome. Its conditional covariance does not depend on the observation, whereas that of the waves does, through the masked fine scales; the contrast suggests that this dependence is what the direct forecasters cannot capture.
\end{enumerate}
Whether cascading is beneficial therefore cannot be explained simply by the underlying physical dynamics. The results point instead to the difficulty of learning the conditional forecast distribution over the full horizon, which depends on the dynamics, the observation, and the distribution of the unobserved scales. This difficulty is reflected in the generalization
behavior of the forecasters. In the four systems where
cascading helps, continued training improves 
Direct SI's fit to the training observations but worsens its
forecasts on unseen observations. Cascaded SI achieves
substantially lower test errors, whereas on the passive
scalar both direct forecasters generalize well
(Appendix~\ref{app:train-test}).

\textbf{The best depth is finite and $> 1$.}
\label{sec:depth}
Figure~\ref{fig:depth_sweep} shows the test error at each cascade depth $S$, with the model, checkpoint, and sampler resolution selected on validation data separately for each depth. For the other four systems, the best test performance occurs at an intermediate depth: $S=5$ for heterogeneous waves and SQG, $S=10$ for Euler, and $S=20$ for Klein--Gordon (for SQG, validation selects $S^\star=10$). Shorter transitions are thus not always better, and the preferred depth depends on the forecasting problem and
horizon (Appendix~\ref{app:horizon-depth}). Under the fixed budget of $16{,}000$ supervised transitions and $30{,}000$ optimizer updates, however, deeper cascades are also trained on fewer distinct trajectories, integrate their samplers more often, and compound learning errors, so the deterioration at large depths need not be caused by memory loss alone. Indeed, additionally conditioning every transition on $X_0$
reduces the forecasting error on the passive scalar but
not on Euler (Appendix~\ref{app:anchoring}).
In both systems, the error still increases at large depths,
consistent with our analysis of learned anchored transitions
in Appendix~\ref{app:anchored-ou-theory}.

The optimal depths seem to agree with the theory. Among other differences, the fine scales drive the velocity more strongly in SQG than in Euler flow, and on SQG we find that a shallower cascade is better, $S=5$ against $S=10$; the optimal gain is similarly smaller ($2.3$ against $3.5$). In the rotation model, a stronger contribution of the hidden process yields a shallower optimum, although we have not measured the memory of the two systems. The wave system illustrates the relevance of how the intermediate states in the cascades are structured. Composing deterministic next-state regressors that use the displacement alone, but not the velocity, reduces the correlation with the reference conditional mean from $0.73$ after one transition to $0.11$ after five; including the velocity proxy, initialized at zero as in the benchmark, improves it from $0.65$ to $0.87$ (Table~\ref{tab:wave_state_ablation} in Appendix~\ref{app:waves}).

\section{Conclusion}
\label{sec:conclusion}

We studied why generative forecasters can sample the wrong distribution of future states given coarse observations of a physical system, and when cascading transitions over shorter time intervals helps. In our experiments, cascaded SIs recovered conditional distributions when direct forecasters failed to do so, for both linear and nonlinear dynamics, but cascading was detrimental for a scalar passively advected and diffused by a prescribed flow. Our analytical model explains both effects by the trade-off between the approximation error of long transitions and the memory error of repeated conditioning on a coarse-grained state.

\textbf{Open questions.} The theory does not predict the depth selected in the PDE experiments; this will require further work, for example measuring how strongly and persistently the unobserved scales drive the observed ones. Cascading increases inference cost, and its benefit may depend on model capacity.

\subsubsection*{Code and data availability.}
We have released the
\href{https://huggingface.co/coarse-grained-forecasting-benchmark/datasets}
{benchmark datasets},
\href{https://huggingface.co/coarse-grained-forecasting-benchmark/forecasting-models/tree/main/code}
{our code}, and
\href{https://huggingface.co/coarse-grained-forecasting-benchmark/forecasting-models}
{trained model checkpoints}.

\subsubsection*{Acknowledgments}
We gratefully acknowledge support from the European Research Council Consolidator Grant 101232533 (PhaseShift). We also acknowledge computing resources provided by sciCORE, the scientific computing center of the University of Basel, and by the Swiss National Supercomputing Centre (CSCS).

\bibliographystyle{iclr2027_conference}
\bibliography{iclr2027_conference,arxiv_refactored_additions}

\clearpage
\appendix
\raggedbottom

\section{Related work}
\label{app:related}

\textbf{Generative models for probabilistic forecasting.}
Recent work has used conditional diffusion models~\citep{batzolis2021conditional, baldassari2023conditional} and stochastic interpolants~\citep{albergo2025stochastic} to represent distributions over future states rather than produce a single deterministic forecast. \cite{pmlr-v235-chen24n} formulate forecasting directly as sampling from the conditional distribution of a future state given the current state, using stochastic interpolants to define a generative path between them. Related diffusion-based approaches have been developed for weather forecasting, including SEEDS~\citep{li2023seeds} and GenCast~\citep{Price2025}, and for complex physical systems and turbulent flows~\citep{kohl2026benchmarking,MUCKE2026107098}. Our work considers the same probabilistic forecasting objective, but focuses on how the physical time interval used for learning affects the accuracy of the resulting conditional distribution.

\textbf{Using intermediate times and composing forecasts.}
Several forecasting methods make use of intermediate physical states or generate trajectories by repeatedly applying learned transitions. GenCast, for example, learns a conditional distribution over a fixed physical time step and samples successive states autoregressively~\citep{Price2025}. Conditional diffusion models have similarly been applied autoregressively to turbulent-flow simulation~\citep{kohl2026benchmarking}. Deterministic weather models have compared step lengths: Pangu-Weather composes models trained for several lead times so as to keep the number of steps small, since errors accumulate with every step~\citep{bi2023accurate}, and Stormer averages forecasts made with different step lengths~\citep{nguyen2024scaling}. For conditional diffusion models of PDEs, \citet{shysheya2024conditional} find that sampling a trajectory autoregressively forecasts better than sampling it all at once. DYffusion~\citep{cachay2023dyffusion} is particularly related to our setting, since it uses intermediate physical times to build a dynamics-informed diffusion process. These works show that intermediate temporal information can be useful for forecasting, and they evaluate forecasts against single realized futures. We instead compare learning $P(X_T\mid X_0)$ directly with learning and composing transitions over intervals of length $T/S$, at a fixed final time $T$ and against reference ensembles of the conditional distribution. We vary $S$ while keeping the number of supervised training pairs fixed. This exposes a non-monotone dependence on cascade depth: making the transitions shorter can improve the forecast, but dividing the interval increasingly finely need not continue to help, and for a passive scalar every added stage makes the forecast worse.

\textbf{Iterated and direct multi-step forecasts.}
In time-series analysis, the choice between iterating a one-step model and estimating a separate model for each horizon has long been studied~\citep{marcellino2006comparison,chevillon2007direct}. For linear autoregressions, it is a trade-off between estimation variance and bias: iterated forecasts are more efficient when the one-step model is correctly specified, while direct forecasts are more robust to misspecification. The two effects in our analysis play analogous roles: a transition over a short interval is easier to approximate than one over the full horizon, which is the approximation effect, and a one-step model of the observed process is misspecified whenever this process is not Markov, which is the memory effect. We study these effects for generative forecasts of the full conditional distribution of high-dimensional physical fields, made with models of limited capacity, rather than for point forecasts with linear models.

\textbf{Partial observation and memory.}
The second effect in our analysis is specific to forecasting from incomplete observations. Projection of a dynamical system onto a subset of its variables generally produces dynamics with memory: unresolved variables can retain information about the past that is not contained in the current observation. This phenomenon is classical in optimal prediction and the Mori--Zwanzig formalism~\citep{chorin2000optimalb,chorin2002optimal}. It has a direct consequence for cascaded probabilistic forecasts. Even if each short transition $P(X_{t_{i+1}}\mid X_{t_i})$ were known exactly, their composition need not equal $P(X_T\mid X_0)$ when the observed process is not Markov. Our theoretical analysis makes this loss of information explicit and studies its competition with the approximation error of learning longer transitions. In the rotation--OU model, these two effects produce a finite optimal cascade depth.

\textbf{Evaluating conditional forecast distributions.}
Evaluating generative models is difficult because realistic individual samples do not by themselves imply that the correct distribution has been learned~\citep{Theis2016a}. Probabilistic forecasting methods therefore use ensemble-based scores, distributional statistics, and diagnostics of forecast uncertainty~\citep{li2023seeds,Price2025,kohl2026benchmarking}. Our benchmark is designed specifically for conditional evaluation under partial observation. For each observed initial state $x_0$, we generate many full initial states that share $x_0$ but differ in their hidden components, and evolve each of them with the physical simulator. The resulting reference ensemble gives direct samples from $P(X_T\mid X_0=x_0)$. This allows us to distinguish a model that generates plausible states from one that recovers the conditional distribution associated with the particular observation supplied to the forecaster.

\section{Experiments}

\subsection{Dataset construction and numerical solvers}
\label{app:datasets}

For each observed initial state, we simulate several trajectories by
varying the hidden initial conditions. We call this set of trajectories
an \emph{ensemble}. All trajectories in an ensemble therefore start from
the same observation but can reach different final states (Figure~\ref{fig:dataset_overview}).

The benchmark uses 80 ensembles for training, 10 for validation, and
10 for testing, with different observed initial states in each split.
Each test ensemble provides $10{,}000$ simulated endpoints for comparison
with the model's predictions.

The model observes only Fourier modes $0<|k|\leq8$, represented by
196 real coordinates per scalar field. Table~\ref{tab:dataset_specifications}
summarizes the simulation settings and observed variables.

\begin{table}[h!]
\centering\small
\setlength{\tabcolsep}{4pt}
\caption{\textbf{Benchmark dataset specifications.}
Parentheses give the number of real coordinates observed by the model.
Each benchmark trajectory contains 41 equally spaced snapshots.
Horizons are reported in physical time.}
\begin{tabular}{@{}lccl@{}}
\toprule
Dataset & Simulation grid & Horizon $T$ & Observed variables \\
\midrule
\euler & $1024^2$ & 1 & Vorticity (196) \\
\sqg & $256^2$ & 20  & Buoyancy (196) \\
\waves & $512^2$ & 1 & Displacement and velocity proxy (392) \\
\nlkg & $128^2$ & 12 &  Displacement and velocity (392) \\
\scalar & $512^2$ & 1 & Concentration (196) \\

\bottomrule
\end{tabular}

\label{tab:dataset_specifications}
\end{table}

\subsubsection{Two-dimensional Euler}
On the unit periodic square, vorticity evolves according to
\begin{equation}
\partial_t\omega+u\cdot\nabla\omega=0,\qquad
u=\nabla^\perp\Delta^{-1}\omega.
\end{equation}
Here $\nabla^\perp=(-\partial_y,\partial_x)$.
The initial field has independently drawn low and high Fourier bands.
Filtered Gaussian white noise with spectral amplitude proportional to
$|k|^{-2}$ on $1<|k|\leq8$ determines the condition; an independent draw
with amplitude proportional to $|k|^{-3}$ on $8<|k|\leq64$ is added for
each member. The bands are normalized separately to velocity RMS values
1 and 0.2.

We use the double-precision \emph{azeban} pseudospectral solver
\cite{azeban}, with a $1024^2$ grid, $1536^2$ padding, smooth-cutoff
spectral viscosity ($s=1$, $\epsilon=0.4$), and adaptive SSP--RK3
integration at CFL 0.5. The 41 states are archived every 0.025 through
$T=1$.
A comparison of $1024^2$ and $2048^2$ simulations for one initial-condition
ensemble supports distributional convergence of the observed endpoint states at the benchmark resolution.

\subsubsection{Surface quasi-geostrophic flow}
On the periodic square $[0,2\pi)^2$, surface buoyancy obeys
\begin{equation}
\partial_t b+J(\psi,b)=0,\qquad
\widehat\psi(k)=\frac{\widehat b(k)}{|k|}\quad(k\ne0).
\end{equation}
Here $J(f,g)=\partial_x f\,\partial_y g-\partial_y f\,\partial_x g$.
Each condition contains one to four signed elliptical vortices,
with uniform centers and orientations, amplitude magnitudes in
$[0.7,1.2]$, aspect ratios in $[1.5,4]$, and widths in
$[2\pi/9,2\pi/6]$. Centers are separated by at least 1.2.
Projection onto $0<|k|\leq8$ defines the observed initial state.
Each ensemble member adds a real Gaussian Fourier field on
$8<|k|\leq32$, with coefficient variance proportional to
$|k|^{-5/3}$, normalized to spatial RMS 0.10.

We use \texttt{pyqg.SQGModel} \citep{pyqg}
(version \texttt{0.7.3.dev23}, commit \texttt{9abad33}),
with $f_0=N=H=1$, $\beta=0$, and no linear drag.
The solver uses a $256^2$ Fourier grid, third-order
Adams--Bashforth stepping with $\Delta t=0.005$, and its built-in
high-wavenumber filter, which dissipates buoyancy variance.
We archive 41 states every 0.5 through $T=20$ and rescale
forecast time to $[0,1]$. A comparison of $256^2$ and $512^2$ simulations, with the time step
halved at the finer resolution, supports distributional convergence
of the observed endpoint states.

\subsubsection{Heterogeneous waves}
\label{app:waves}
Displacement $u$ and velocity $v$ evolve on the unit periodic square:
\begin{equation}
\partial_tu=v,\qquad
\partial_tv=\nabla\cdot(c^2\nabla u)-0.2v,\qquad
c^2(x,y)=1+0.5\sin(2\pi x)\sin(2\pi y).
\end{equation}
The medium is fixed. Each condition combines an anisotropic Gaussian
background (correlation lengths 0.06 and 0.03, scale 0.2) with 18 random
boxes, triangles, or disks. Shapes overwrite earlier values where they
overlap; centers and orientations are uniform, sizes lie in $[0.08,0.20]$,
and amplitudes in $[-3,3]$. Smoothing with Gaussian width $2/128$ and
projection onto $0<|k|\leq8$ define the shared initial displacement.

Each ensemble member adds masked Gaussian noise projected onto
$8<|k|\leq16$. The mask depends on the condition's shape interiors;
before masking, the band filter gives expected spatial mean square
$3.3^2$. Initial velocity consists of three Gaussian kicks with uniform
centers, width 0.03, and independent strengths $\mathcal N(0,165^2)$,
projected onto $|k|>8$. Both perturbations therefore leave the observed
initial state unchanged.

A double-precision PyTorch solver uses centered spatial differences
and RK4 on a $512^2$ grid with $\Delta t=2.5\times10^{-4}$.
We archive 41 states every 0.025 through $T=1$.
The observed state combines coarse displacement with the velocity proxy
\begin{equation}
\hat v_{t_n}=P_8\frac{u_{t_n}-u_{t_{n-1}}}{0.025}\quad(n>0),
\qquad \hat v_0=0,
\end{equation}
where $P_8$ extracts the 196 real observed Fourier coordinates.
Training targets use this proxy; during generation, the model predicts
both channels and passes them to the next segment.
An independent \texttt{py-pde} implementation \citep{Zwicker2020} verifies
the spatial operator. Time-step and grid-refinement checks support
numerical accuracy of the observed states.

\textbf{Wave state diagnostic.}
\label{app:wave_ablation}
We compare deterministic regressors using
displacement alone or the augmented state $(u,\hat v)$. They have nearly
equal parameter counts and use mean-squared error, 8,000 training updates,
and learning rates $3\times10^{-4}$ and $10^{-3}$, with checkpoint selection
on eight development conditions. Each depth receives $B=16{,}000$
supervised transitions. We recursively compose the predictions and measure
their correlation with the simulated conditional endpoint mean on ten
validation conditions (Table~\ref{tab:wave_state_ablation}). The supplied
initial proxy uses a forward difference and therefore gives additional
initial information. Setting that proxy to zero at inference tests the
same fitted augmented model without this information.

\begin{table}[h!]
\centering\small
\caption{\textbf{Wave state diagnostic.} Mean correlation with the true
conditional endpoint mean over ten validation conditions. Both augmented columns use the same fitted model. The benchmark initializes the proxy to zero and subsequently carries generated values.}
\begin{tabular}{@{}cccc@{}}
\toprule
& & \multicolumn{2}{c}{Displacement and velocity proxy} \\
\cmidrule(lr){3-4}
$S$ & Displacement only & Initial proxy supplied & Initial proxy zero \\
\midrule
1 & 0.731 & 0.760 & 0.647 \\
5 & 0.108 & 0.949 & 0.871 \\
10 & 0.094 & 0.953 & 0.888 \\
20 & 0.024 & 0.946 & 0.848 \\
\bottomrule
\end{tabular}
\label{tab:wave_state_ablation}
\end{table}

\subsubsection{Nonlinear Klein--Gordon}
\label{app:nlkg_dataset}
On the periodic square $[0,2\pi)^2$, displacement and velocity obey
the defocusing nonlinear Klein--Gordon equation:
\begin{equation}
\partial_tu=v,\qquad
\partial_tv=\Delta u-u-64u^3.
\end{equation}
Each condition consists of two travelling Gaussian wave packets,
projected onto $0<|k|\leq8$ in both fields.
Centers, directions, and phases are uniform; amplitudes lie in
$[0.5,1]$, carrier wavenumber magnitudes in $[3,5]$, and the
envelope width is 0.8.
Each ensemble member adds hidden displacement and velocity on
$8<|k|\leq24$. Their Fourier coefficients are proportional to
$g_k/\sqrt{1+|k|^2}$ and $g'_k$, respectively, with independent
complex Gaussian draws and reality enforced.
Both hidden fields are scaled by the same member-specific factor
to give displacement RMS 0.20.
We observe 196 real Fourier coordinates of $u$ and 196 of $v/4$.

Dedalus 3.0.5 \citep{Dedalus} uses a $128^2$ Fourier grid,
factor-two dealiasing for the cubic term, and the third-order
IMEX scheme RK443 with $\Delta t=0.0025$.
States are archived every 0.1; retaining every third state gives
41 snapshots through $T=12$. Spatial and time-step refinement checks support numerical convergence.

\subsubsection{Passive scalar}
On the unit periodic square, concentration evolves under a fixed flow:
\begin{align}
\partial_t q+v\cdot\nabla q&=\kappa\Delta q,\\
v(x,y)&=\bigl(V\sin(2\pi Kx)\cos(2\pi Ky),
             -V\cos(2\pi Kx)\sin(2\pi Ky)\bigr),
\end{align}
with $K=2$, $V=0.5$, and $\kappa=6.184154\times10^{-6}$.
Each condition is drawn on a $256^2$ generator grid from Gaussian white
noise filtered with amplitude $|k|^{-2}$ on $1<|k|\leq8$ and normalized
to spatial RMS 1. Each ensemble member adds independent Gaussian noise
filtered with amplitude $|k|^{-3}$ on $8<|k|\leq64$.
A fixed multiplier gives expected spatial mean square 1.14491,
without normalizing individual members. The hidden field is therefore
Gaussian, and linear evolution preserves a Gaussian conditional
endpoint distribution.

A double-precision Fourier--Galerkin solver uses integrating-factor
RK4 on a $512^2$ grid with $768^2$ padding and $\Delta t=0.005/6$.
We archive states every 0.005 through $T=1$ and retain every fifth
state for the 41-snapshot benchmark.
Independent simulations of five trajectories with Dedalus
\citep{Dedalus} agree with the observed dataset states as the
Dedalus time step is refined.

\subsection{Training, selection, and numerical sampling}
\label{app:training}

\textbf{Supervision and architecture.}
The matched-budget comparison uses 80 canonical training conditions and
200 candidate trajectories per condition. At depth
$S\in\{1,2,5,10,20,40\}$, $200/S$ trajectories supply all $S$ consecutive
transitions, keeping the budget at $B=16{,}000$. Direct SI is the $S=1$
member of this family. Normalization is fitted on training
data and shared across depths and methods.

The U-Net acts on $32\times32$ reconstructions of the observed modes,
with channel widths $64,128,256$, circular $3\times3$ convolutions,
SiLU activations, two average-pooling stages, and skip connections.
Its output is projected back onto the observed modes. SI receives the
interpolated state, conditioning state, auxiliary time, and physical
segment start/end times. Euler uses 1,515,265 parameters. Passive scalar
and waves also receive spatial-coordinate features because their flow
or medium is spatially inhomogeneous. The wave model predicts both
displacement and velocity-proxy channels.

\textbf{Objectives and samplers.}
For a normalized pair $(a,b)$, draw
$\tau\sim\operatorname{Unif}[0,1]$ and $z\sim\mathcal N(0,I)$, and set
\begin{equation}
I_\tau=(1-\tau)a+\tau b+\sigma(1-\tau)\sqrt\tau\,z,
\qquad R_\tau=b-a-\sigma\sqrt\tau\,z.
\end{equation}
The drift regresses $R_\tau$ from $(I_\tau,\tau,a,t_i,t_{i+1})$ using
the sum of $\tfrac12\|b_\theta\|^2-b_\theta\cdot R_\tau$ for antithetic
draws $z,-z$. Euler--Maruyama samples
\begin{equation}
dZ_\tau=b_\theta(Z_\tau,\tau,a,t_i,t_{i+1})\,d\tau
       +\sigma(1-\tau)\,dW_\tau,\qquad Z_0=a.
\end{equation}
At the next physical segment, $a$ is replaced by the generated endpoint.
SI gradient norms are clipped at 1.

EDM uses the conditional denoising objective and preconditioning of
\citet{NEURIPS2022_a98846e9}, with $\sigma_{\rm data}=1$,
$P_{\rm std}=1.2$, and a searched log-noise location $P_{\rm mean}$.
Its deterministic Heun sampler uses $\sigma_{\min}=0.002$,
$\sigma_{\max}=80$, and schedule exponent 7; $n$ numerical steps require
$2n-1$ network evaluations. EDM uses no gradient clipping or
learning-rate ramp-up.

\textbf{Optimization and selection.}
Models train for 30,000 Adam updates with batch size 256, constant learning
rate, no weight decay, $(\beta_1,\beta_2)=(0.9,0.999)$, and
$\epsilon=10^{-8}$. Training uses float32, without mixed precision or
TF32. Exponential-moving-average weights have a 500,000-example half-life
and ramp-up ratio 0.05 and are used for validation and test.
Checkpoints are scored every 1,000 updates and at updates
100, 200, 300, 500, and 750.

Checkpoint and hyperparameter selection minimize the mean validation
energy-distance U-statistic~\citep{SZEKELY20131249}; $S^\star$ minimizes it across depths.
For generated states $A=\{a_i\}_{i=1}^n$ and true states
$R=\{r_j\}_{j=1}^m$, we use
\begin{equation}
\widehat{\mathrm{ED}}_U(A,R)=
\frac{2}{nm}\sum_{i,j}\|a_i-r_j\|
-\frac{1}{n(n-1)}\sum_{i\ne i'}\|a_i-a_{i'}\|
-\frac{1}{m(m-1)}\sum_{j\ne j'}\|r_j-r_{j'}\|.
\end{equation}
For sampler-resolution calibration, we instead use the corresponding
V-statistic
\begin{equation}
\widehat{\mathrm{ED}}_V(A,R)=
\frac{2}{nm}\sum_{i,j}\|a_i-r_j\|
-\frac{1}{n^2}\sum_{i,i'}\|a_i-a_{i'}\|
-\frac{1}{m^2}\sum_{j,j'}\|r_j-r_{j'}\|,
\end{equation}
which includes self-pairs and is more stable for relative convergence
checks near zero.

Selection uses 100 true and 100 generated states per condition in
normalized model coordinates, with float64 accumulation. Exact ties
select the earlier checkpoint. One training seed is used; sampling random
numbers are shared across configurations within each condition.

Initial learning rates are $\{10^{-4},3\times10^{-4},10^{-3}\}$,
SI noise scales are $\{0.25,0.5,1,2,4,8,16,32\}$, and EDM uses
$P_{\rm mean}\in\{-3.2,-2.8,\ldots,-0.4\}$.
One boundary-extension round is allowed. At the lower or upper edge,
learning rates extend by $\{3\times10^{-5},10^{-5}\}$ or
$\{3\times10^{-3},10^{-2}\}$, SI noise by $\{0.125,0.0625\}$ or
$\{64,128\}$, and $P_{\rm mean}$ by $\{-3.6,-4\}$ or $\{0,0.4\}$.
Unresolved search boundaries retain a flag.

\textbf{Sampler resolution.}
For each depth and method, a fixed probe is evaluated using
$\widehat{\mathrm{ED}}_V$ on the step ladder
$20$, $40$, $80$, $160$, $320$, and $640$. The upper resolution of the first doubling with
relative score change below $5\%$ is retained. Probes use the final
checkpoint at learning rate $3\times10^{-4}$, with SI noise 8 or EDM
$P_{\rm mean}=-1.6$. The selected model is checked again at twice its
assigned resolution. A failed check permits one resolution escalation
and validation reselection; a repeated failure is flagged. Sampler
calibration is therefore separate from model selection by
$\widehat{\mathrm{ED}}_U$.

To verify that the direct baselines are not limited by sampler resolution, we reevaluate the selected Direct SI and EDM checkpoints using the same number of network evaluations (NFE) as Cascaded SI. As shown in Table \ref{tab:matched_nfe}, EDM performance remains essentially unchanged. Direct SI improves slightly on waves and NLKG (1–2\%), but remains substantially worse than Cascaded SI. On Euler and SQG, increasing the sampler resolution does not improve performance.

\begin{table}[h!]
\centering
\caption{\textbf{Matched inference cost.} Test csFID-10k for Direct SI and EDM at the original 40 sampling steps and at approximately the same number of network evaluations (NFE) as Cascaded SI. Standard errors across the ten test conditions are omitted for readability. The 40-step values were re-generated for this comparison from the Table~\ref{tab:results} checkpoints and agree with Table~\ref{tab:results} to within 0.24 csFID (different GPU hardware). Cascaded SI values are those of Table~\ref{tab:results}.}
\label{tab:matched_nfe}
\small
\begin{tabular}{lccc}
\toprule
System
& Direct SI (40 $\rightarrow$ matched)
& EDM (40 $\rightarrow$ matched)
& Cascaded SI (NFE) \\
\midrule
Euler
& $57.02 \rightarrow 57.11$
& $56.21 \rightarrow 56.24$
& $\mathbf{16.32}$ (800) \\
SQG
& $138.48 \rightarrow 140.45$
& $118.50 \rightarrow 118.56$
& $\mathbf{51.24}$ (1600) \\
Waves
& $115.83 \rightarrow 114.56$
& $123.02 \rightarrow 123.05$
& $\mathbf{36.66}$ (200) \\
NLKG
& $98.27 \rightarrow 96.10$
& $92.32 \rightarrow 92.26$
& $\mathbf{16.24}$ (6400) \\
\bottomrule
\end{tabular}
\end{table}

\textbf{Selected configurations.}
Table~\ref{tab:selected_settings} reports the frozen selections for all
five systems. Sampler steps are reported per segment; network function
evaluations (NFE) count the full forecast.

\begin{table}[t]
\centering\footnotesize
\setlength{\tabcolsep}{2.5pt}
\caption{Selected configurations and sampling cost.}
\begin{tabular}{@{}llrrrrrr@{}}
\toprule
System & Method & $S$ & Learning rate & Noise parameter & Update & Sampler steps & NFE \\
\midrule
\euler
 & Direct SI   & 1  & $10^{-4}$        & $\sigma=8$              & 5,000  & 40 & 40 \\
 & EDM         & 1  & $10^{-3}$        & $P_{\rm mean}=-1.6$      & 300    & 40 & 79 \\
 & Cascaded SI & 10 & $3\times10^{-4}$ & $\sigma=8$              & 17,000 & 80 & 800 \\
 \addlinespace
\sqg
 & Direct SI   & 1  & $10^{-4}$        & $\sigma=16$         & 4,000  & 40  & 40 \\
 & EDM         & 1  & $3\times10^{-4}$ & $P_{\rm mean}=-3.2$ & 4,000  & 40  & 79 \\
 & Cascaded SI & 10 & $10^{-3}$        & $\sigma=16$         & 11,000 & 160 & 1,600 \\
\addlinespace
\waves
 & Direct SI   & 1 & $10^{-3}$        & $\sigma=32$             & 6,000  & 40 & 40 \\
 & EDM         & 1 & $3\times10^{-5}$ & $P_{\rm mean}=-0.4$     & 3,000  & 40 & 79 \\
 & Cascaded SI & 5 & $3\times10^{-3}$ & $\sigma=2$              & 30,000 & 40 & 200 \\
 \addlinespace
\nlkg
 & Direct SI   & 1  & $10^{-3}$ & $\sigma=128$     & 15,000 & 40  & 40 \\
 & EDM         & 1  & $10^{-3}$ & $P_{\rm mean}=-4$ & 17,000 & 40  & 79 \\
 & Cascaded SI & 20 & $10^{-4}$ & $\sigma=1$       & 30,000 & 320 & 6,400 \\
\addlinespace
\scalar
 & Direct/Cascaded SI & 1 & $3\times10^{-3}$ & $\sigma=1$          & 6,000  & 40 & 40 \\
 & EDM                & 1 & $10^{-3}$        & $P_{\rm mean}=-1.2$ & 16,000 & 40 & 79 \\
\bottomrule
\end{tabular}
\label{tab:selected_settings}
\end{table}

\textbf{Hyperparameter boundary checks.}
Three selected configurations initially reached the boundary
of their hyperparameter search range. We therefore extended
the searches for nonlinear Klein--Gordon Direct SI and
heterogeneous-wave EDM using validation data only.

For Direct SI, extending the noise range to
$\sigma = 256, 512$ and testing higher learning rates did not
improve the selected configuration ($\mathrm{ED}_U = 10.651$).
For wave EDM, the best additional configuration reached
$\mathrm{ED}_U = 10.678$, compared with $10.685$ for the
original selection. This difference is smaller than the
observed across-seed standard deviation ($0.028$), so we
retained the original configuration under our predefined
extension criterion.

The nonlinear Klein--Gordon EDM search remains unresolved.
Validation errors varied little across the original
50 configurations, and additional training modifications
did not improve performance.

\subsection{Conditional distributional evaluation}
\label{app:evaluation}
csFID-10k reconstructs unnormalized coarse fields on a $128\times128$
grid. Signed values are clipped symmetrically at the training-fitted
0.9999 quantile of absolute amplitude, linearly mapped to uint8, and
replicated into three identical channels. There is no colormap or
per-image normalization. For heterogeneous waves and nonlinear Klein--Gordon, csFID evaluates
displacement only, although the forecasting state also includes a
velocity channel.
Images undergo one bicubic resize to $299\times299$ in
\texttt{clean-fid} 0.1.35~\citep{parmar2022aliasedresizingsurprisingsubtleties}. The first seven channels of Inception's
\texttt{mixed\_6/conv} $17\times17$ activation map are flattened in
spatial--channel order, giving 2023 features without spatial pooling.
Pooled cFID uses the same preprocessing and Inception weights with
2048-dimensional \texttt{pool3} features.

For each condition, the Fr\'echet distance is computed between the feature
means and covariances of 10,000 true and 10,000 generated endpoints.
The conditional score averages these distances with equal condition
weights. Reported uncertainty is the sample standard deviation of
condition-level scores divided by the square root of the number of
conditions. For the main comparison, this is SEM over ten test conditions
for one training seed.
Secondary state-space diagnostics use fixed subsets of 200 true and
200 generated endpoints unless otherwise stated.

Numerical values from the depth sweep presented in Section~\ref{sec:depth} are reported in Table~\ref{tab:depth_values}.

\begin{table}[tbp]
\centering\small
\setlength{\tabcolsep}{4pt}
\caption{\textbf{Conditional forecasting performance across cascade depths.}
Test csFID-10k, mean $\pm$ SEM across ten conditioning ensembles for one
training seed. Bold indicates the lowest SI score within each system; EDM is
shown for comparison. For SQG at $S=20$, one nonfinite generated endpoint was excluded from one
conditioning ensemble, leaving 9,999 generated samples for that score.}
\begin{tabular}{@{}lccccc@{}}
\toprule
Segments $S$ & \euler & \sqg & \waves & \nlkg & \scalar \\
\midrule
1 (Direct SI) & $57.26\pm3.35$ & $138.48\pm23.09$ & $115.86\pm6.76$ & $98.48\pm6.18$ & $\mathbf{8.67\pm0.28}$ \\
2 & $53.60\pm3.28$ & $83.25\pm16.31$ & $71.23\pm3.91$ & $99.75\pm5.89$ & $18.75\pm1.17$ \\
5 & $22.94\pm2.55$ & $\mathbf{43.57\pm8.79}$ & $\mathbf{36.66\pm1.78}$ & $53.10\pm7.30$ & $32.97\pm1.61$ \\
10 & $\mathbf{16.32\pm1.14}$ & $51.24\pm8.32$ & $44.51\pm2.34$ & $22.50\pm5.40$ & $41.14\pm2.36$ \\
20 & $17.86\pm0.87$ & $89.08\pm7.56$ & $58.28\pm2.58$ & $\mathbf{16.24\pm1.55}$ & $56.95\pm2.79$ \\
40 & $23.29\pm1.50$ & $86.98\pm12.00$ & $62.81\pm3.25$ & $18.23\pm2.05$ & $75.21\pm4.74$ \\
\midrule
EDM & $56.32\pm3.74$ & $118.62\pm21.67$ & $123.04\pm7.07$ & $92.37\pm5.43$ & $7.88\pm0.31$ \\
\bottomrule
\end{tabular}
\label{tab:depth_values}
\end{table}

\subsection{Validation of the spatial feature metric}
\label{app:sfid}

The conditioning field fixes spatial registration: realistic structures at
incorrect locations can still constitute an incorrect conditional forecast.
We therefore use spatial FID (sFID) \citep{pmlr-v139-nash21a}, which retains
intermediate spatial Inception features, alongside pooled FID as a diagnostic.
The representations have 2023 and 2048 coordinates, respectively; all image
preprocessing and Inception weights are shared. Appendix~\ref{app:evaluation}
specifies their construction and conditional aggregation.

\textbf{Agreement with state-space diagnostics.}
On Euler, pooled cFID separates Direct SI and EDM more strongly than csFID does
(Table~\ref{tab:fid_sfid_diagnostics}). The paired EDM-minus-Direct difference
over the ten test conditions is $-9.35\pm1.31$ for cFID and
$-0.94\pm0.90$ for csFID (mean $\pm$ SEM). The closer csFID scores agree with
the similar sliced Wasserstein distances, conditional-mean errors, and spread
ratios of these two endpoint models. Cascaded SI has the lowest cFID, csFID, sliced Wasserstein distance,
and conditional-mean error.

\begin{table}[h!]
\centering\small
\caption{\textbf{Euler metric comparison.} Mean $\pm$ SEM over ten test
conditions. Feature metrics use $10{,}000$ true and generated endpoints per
condition. State-space diagnostics use 200 of each in normalized Fourier
coordinates: sliced $W_2$ uses 100 fixed random projections, mean error is the
Euclidean distance between ensemble means, and spread is the ratio of mean
pairwise distances (generated/true, excluding self-pairs). Lower is better
except for spread, whose target is one.}
\begin{tabular}{@{}lccc@{}}
\toprule
Metric & Cascaded SI ($S=10$) & Direct SI & EDM \\
\midrule
cFID & $8.96\pm0.89$ & $24.30\pm2.08$ & $14.94\pm1.21$ \\
csFID & $16.32\pm1.14$ & $57.26\pm3.35$ & $56.32\pm3.74$ \\
Sliced $W_2$ & $0.188\pm0.012$ & $0.390\pm0.026$ & $0.389\pm0.025$ \\
Conditional-mean error & $1.94\pm0.17$ & $5.22\pm0.34$ & $5.17\pm0.32$ \\
Spread ratio & $0.931\pm0.013$ & $1.060\pm0.023$ & $1.065\pm0.021$ \\
\bottomrule
\end{tabular}

\label{tab:fid_sfid_diagnostics}
\end{table}

\textbf{Sensitivity to spatial rearrangement.}
We split each true $10{,}000$-member endpoint bank into disjoint sets $A,B$ of
5000 samples and compare $d(A,B)$ with $d(F(A),B)$. Transformations act exactly
on Fourier coefficients, before image construction. A translation of $d$
pixels shifts both coordinates by $d$ on the $128\times128$ reconstruction.
Euler averages ten resplits within each condition; passive scalar uses one
split. Table~\ref{tab:spatial_sensitivity} reports ratios of across-condition
means to the corresponding truth-to-truth mean. On Euler, spatial features
respond more strongly relative to this baseline. Pooled features also respond strongly on passive scalar, so their
sensitivity to spatial rearrangement is dataset-dependent. These probes measure sensitivity, not a calibrated
distance between spatial transformations. Resplits reuse the same bank and
are not independent ensemble replicates.

\begin{table}[h!]
\centering\small
\caption{\textbf{Response to spatial perturbations.} Each entry is the ratio
of the perturbed score to the truth-to-truth score, using means over ten
conditioning states. Passive scalar has only two distinct
outputs because Cascaded SI selects $S^\star=1$.}
\begin{tabular}{@{}lrrrr@{}}
\toprule
& \multicolumn{2}{c}{\euler} & \multicolumn{2}{c}{\scalar} \\
\cmidrule(lr){2-3}\cmidrule(lr){4-5}
Transformation & cFID & csFID & cFID & csFID \\
\midrule
Translation $(2,2)$ px & $1.04$ & $1.16$ & $2.52$ & $4.57$ \\
Translation $(16,16)$ px & $2.62$ & $5.68$ & $28.50$ & $30.80$ \\
Rotation $90^\circ$ & $3.65$ & $7.55$ & $11.17$ & $26.07$ \\
Reflection in $x$ & $2.84$ & $7.65$ & $9.21$ & $25.24$ \\
\bottomrule
\end{tabular}
\label{tab:spatial_sensitivity}
\end{table}

\textbf{Sample-size stability.}
Recomputing both metrics on nested ensembles of $N=2000,5000,10000$ preserves
the ordering of the mean scores on Euler and passive scalar. On Euler, the
EDM/Direct-SI score ratios are $0.995,0.988,0.984$ for csFID and
$0.706,0.646,0.615$ for cFID. Thus the smaller separation under spatial
features persists throughout this range.

\subsection{Additional qualitative conditional forecasts}
\label{app:qualitative}

\begin{figure}[t]
\centering
\includegraphics[width=\linewidth]{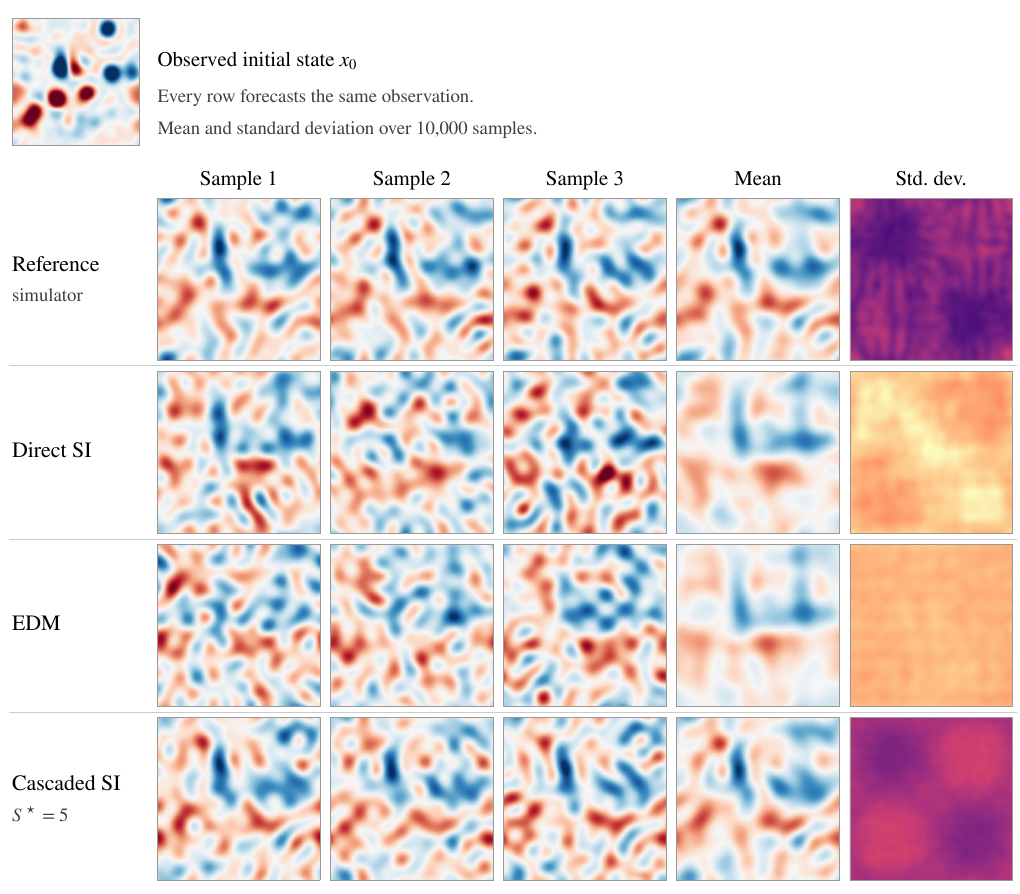}
\caption{\textbf{Heterogeneous waves: direct and cascaded forecasts of one held-out observation.}
The displayed field is the displacement; the forecasters also predict a velocity proxy. Cascaded SI uses the validation-selected depth $S^\star=5$. Means and standard deviations are estimated from $10{,}000$ samples.}
\label{fig:waves_qualitative}
\end{figure}

\begin{figure}[t]
\centering
{\def\svgwidth{\columnwidth}
 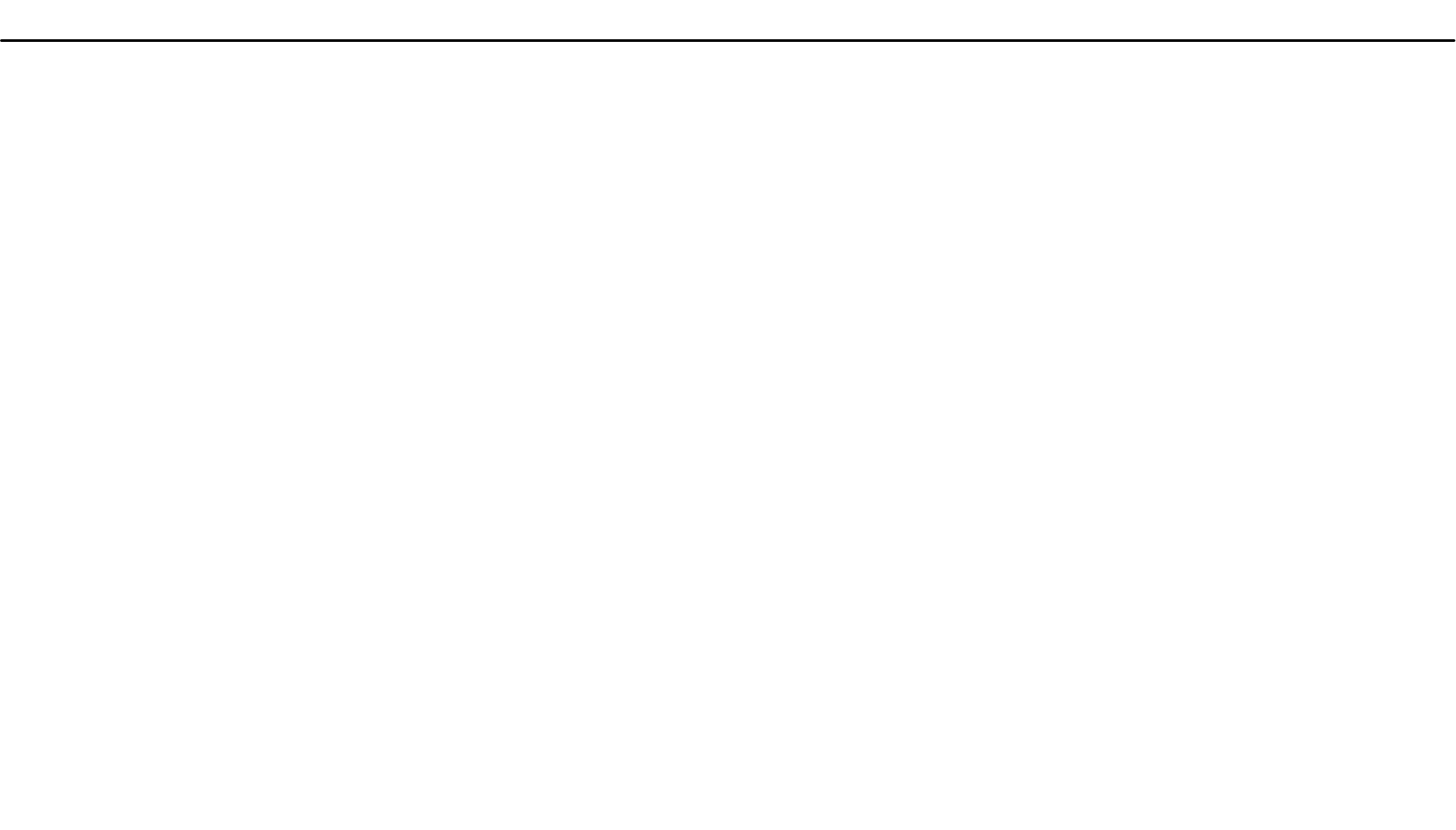}
\caption{\textbf{Additional conditional endpoint distributions.}
Test conditions for SQG, nonlinear Klein--Gordon, and passive scalar.
For each method, three endpoint samples are shown together with the conditional
mean and pointwise standard deviation estimated from 10,000 generated
endpoints; reference statistics are computed from 10,000 simulator-generated
endpoints from the same conditioning state.}
\label{fig:qualitative_appendix}
\end{figure}

Figure~\ref{fig:waves_qualitative} shows the comparison of direct and cascaded forecasts for one test condition of the heterogeneous-wave system, in the layout of Figure~\ref{fig:failure}. Figure~\ref{fig:qualitative_appendix} shows an example test condition for SQG, nonlinear Klein--Gordon, and passive scalar. For SQG and nonlinear
Klein--Gordon, Cascaded SI outperforms Direct SI and EDM by a large amount. For passive scalar, validation selects $S^\star=1$, so Direct SI and Cascaded SI coincide.

\subsection{Additional quantitative experiments}
\label{app:additional-experiments}

\subsubsection{Effect of training-condition diversity.}
\label{app:training-diversity}
We conduct an additional experiment on Euler at $T=1$. Keeping the
training budget fixed at $16{,}000$ supervised pairs, we
compare Direct SI trained on $80$ initial conditions with
$200$ trajectories each against $800$ initial conditions
with $20$ trajectories each. As a reference, we include
Cascaded SI with $S=10$, trained on $80$ initial conditions
with $20$ trajectories each and $10$ transitions per
trajectory. The cascade depth is fixed rather than
reoptimized for this experiment.

Architecture and optimization follow the main benchmark
(U-Net, Adam, batch size $256$, $30{,}000$ updates, and the
same EMA and checkpoint schedule). For each configuration,
we search learning rates
$\{3\times10^{-5},10^{-4},3\times10^{-4},10^{-3},
3\times10^{-3}\}$
and SI noise scales
$\{0.125,0.25,0.5,0.75,1,1.5,2,3,4,6\}$,
with one boundary extension to $\{9,13.5\}$.
This gives $60$ hyperparameter configurations per training
allocation.
Checkpoints are selected by biased validation energy distance
$\widehat{\mathrm{ED}}_V$, using $100$ generated samples
against the full reference ensemble per validation condition,
and choosing the earliest checkpoint within $0.5\%$ of
the minimum. Sampling uses $80$ steps for Direct SI and
$160$ per segment for Cascaded SI, with convergence
verified by doubling. Hyperparameters are selected using
seed 0, and only the selected configuration is retrained
with seed 1. Test evaluation uses the same csFID-10k metric,
image adapter, and reference ensembles as Table~\ref{tab:results}.

\begin{table}[h!]
\caption{\textbf{Effect of training-condition diversity on Euler.}
Results are averaged over two training seeds within each
of ten test conditions and reported as mean $\pm$ SEM
across conditions.}
\label{tab:euler_training_diversity}
\centering
\small
\begin{tabular}{@{}lcc@{}}
\toprule
Method & Training allocation & csFID-10k $\downarrow$ \\
\midrule
Direct SI   & $80 \times 200$ &
$59.88 \pm 3.39$ \\
Direct SI   & $800 \times 20$ &
$49.11 \pm 3.52$ \\
Cascaded SI & $80 \times 20 \times 10$ &
$\mathbf{16.63} \pm 0.81$ \\
\bottomrule
\end{tabular}
\end{table}

Increasing the number of distinct initial conditions by a factor of ten improves Direct SI by $18\%$, with a paired csFID reduction
of $10.8\pm2.3$ across the ten test conditions. Nevertheless, Cascaded SI achieves approximately
three times lower csFID, despite using ten times fewer original
initial conditions and physical trajectories ($1{,}600$
versus $16{,}000$). Its depth $S=10$ was not reoptimized
for this experiment, so a different depth might yield even better results.

\subsubsection{Training with one trajectory per initial condition}
\label{app:single-trajectory}

To investigate whether the advantage of cascading depends on
having multiple training trajectories per initial condition,
we conduct an additional experiment on Euler with 16,000
distinct training initial conditions and exactly one trajectory
each. Both Direct SI and Cascaded SI use the same 16,000
trajectories and receive 16,000 supervised pairs, matching
the training-data budget of the main benchmark. Direct SI
uses one full-horizon endpoint pair per trajectory, while
Cascaded SI uses one short transition per trajectory, balanced
across the $S$ segments.

We follow the main benchmark protocol for architecture,
optimization, hyperparameter and checkpoint selection, and
sampler calibration. Normalization is fitted on the new
training data, while validation and test conditions and
reference ensembles remain unchanged. We evaluate depths
$S\in\{1,2,5,10\}$, selecting $S^\star=10$ on validation.

\begin{table}[t]
    \centering
    \caption{Euler forecasting performance when training with
    one trajectory per initial condition. Test csFID-10k
    (mean $\pm$ SEM across ten test conditions for one training
    seed). Bold indicates the validation-selected depth.}
    \label{tab:single-trajectory}
    \begin{tabular}{lr}
        \toprule
        Model & csFID-10k $\downarrow$ \\
        \midrule
        Direct SI ($S=1$)    & $44.20\pm3.29$ \\
        Cascaded SI ($S=2$)  & $33.77\pm2.76$ \\
        Cascaded SI ($S=5$)  & $17.29\pm1.25$ \\
        Cascaded SI ($S=10$) & $\mathbf{14.70\pm0.48}$ \\
        \bottomrule
    \end{tabular}
\end{table}

As shown in Table~\ref{tab:single-trajectory}, Cascaded SI
achieves roughly three times lower csFID-10k than Direct SI
and outperforms it on all ten test conditions. Notably, it
achieves a score comparable to the main benchmark
(14.70 versus 16.32).

\subsubsection{Training versus test conditions}
\label{app:train-test}

We evaluated the models in Table~\ref{tab:results} on all 80 training conditions
and 10 test conditions, using the validation-selected checkpoint
and the final checkpoint after 30,000 updates. We used the
original samplers and unbiased energy distance
($\mathrm{ED}_U$), with 100 generated and 100 reference samples
per condition.

\begin{table}[h!]
\centering
\caption{\textbf{Training and test performance.}
Results are for the original Table~\ref{tab:results} models. Small negative values can occur with the unbiased  $\mathrm{ED}_U$ metric.}
\label{tab:train-test}
\small
\begin{tabular}{llrrrr}
\toprule
& & \multicolumn{2}{c}{Validation-selected}
  & \multicolumn{2}{c}{30k updates} \\
\cmidrule(lr){3-4}\cmidrule(lr){5-6}
System & Model & Train & Test & Train & Test \\
\midrule
Euler
 & Direct SI   & 1.09 & 1.58 & -0.02 & 2.01 \\
 & EDM         & 1.07 & 1.50 & -0.02 & 1.86 \\
 & Cascaded SI & 0.12 & 0.19 &  0.13 & 0.22 \\
\addlinespace
SQG
 & Direct SI   & 5.25 & 6.96 & 0.06 & 7.73 \\
 & EDM         & 4.42 & 6.28 & 0.64 & 6.40 \\
 & Cascaded SI & 0.85 & 2.55 & 0.41 & 2.73 \\
\addlinespace
Waves
 & Direct SI   & 7.23 & 10.06 & 0.90 & 16.31 \\
 & EDM         & 9.72 & 10.69 & 0.38 & 18.46 \\
 & Cascaded SI & 1.93 &  2.84 & 1.93 &  2.84 \\
\addlinespace
NLKG
 & Direct SI   & 8.53 & 9.98 & 0.96 & 11.22 \\
 & EDM         & 8.15 & 9.48 & 5.80 &  9.88 \\
 & Cascaded SI & 0.63 & 0.95 & 0.63 &  0.95 \\
\addlinespace
Passive scalar
 & Direct/Cascaded SI & -0.03 & 0.16 & -0.02 & 0.23 \\
 & EDM                & -0.02 & 0.12 & -0.03 & 0.14 \\
\bottomrule
\end{tabular}
\end{table}

On Euler, SQG, waves, and NLKG, Direct SI fits its training
conditions increasingly well but generalizes poorly to unseen
conditions. Cascaded SI achieves substantially lower test errors,
which remain relatively stable during further training.
EDM shows a similar generalization gap on Euler, SQG, and waves,
but does not fit its training conditions as well on NLKG.
On passive scalar, both direct methods generalize well.

These results explain the early checkpoint selection of the
direct models reported in Table \ref{tab:selected_settings}: continued training improves their fit to the
training conditions while validation performance deteriorates.
The test evaluations at 30k were diagnostic only and did not
affect model selection.

\subsubsection{Forecast horizon and cascade depth}
\label{app:horizon-depth}

We study how the benefit of cascading changes with the
forecasting horizon on Euler. We consider
$T\in\{0.2,0.4,0.6,0.8,1.0\}$ and, at each horizon,
all cascade depths $S$ that divide the $40T$ stored time
steps, giving 31 $(T,S)$ configurations.

We use the same Euler data, network architecture,
validation-based selection criterion, and csFID-10k
evaluation as in the main benchmark. For this experiment,
the hyperparameter search is wider: learning rates
$\{3\times10^{-5},10^{-4},3\times10^{-4},
10^{-3},3\times10^{-3}\}$ and SI noise scales from
$0.5$ to $64$. As in the benchmark protocol, if the best configuration lies on a boundary of this grid, the search is extended once in that direction (learning rate $10^{-5}$ or $10^{-2}$; noise scale $0.25$ and $0.125$, or $128$). Each configuration is evaluated with three training seeds, with hyperparameters selected by mean validation $\mathrm{ED}_U$ across seeds. Training
runs for 30,000 updates, with one extension to 60,000
if validation error is still improving (14 of 31
configurations). We use 15,840 supervised pairs
instead of 16,000 so that the pair budget is divisible
by every tested depth. The sampler-resolution ladder
is unchanged, with additional convergence checks.

\begin{figure}[h!]
\centering
\includegraphics[width=\linewidth]{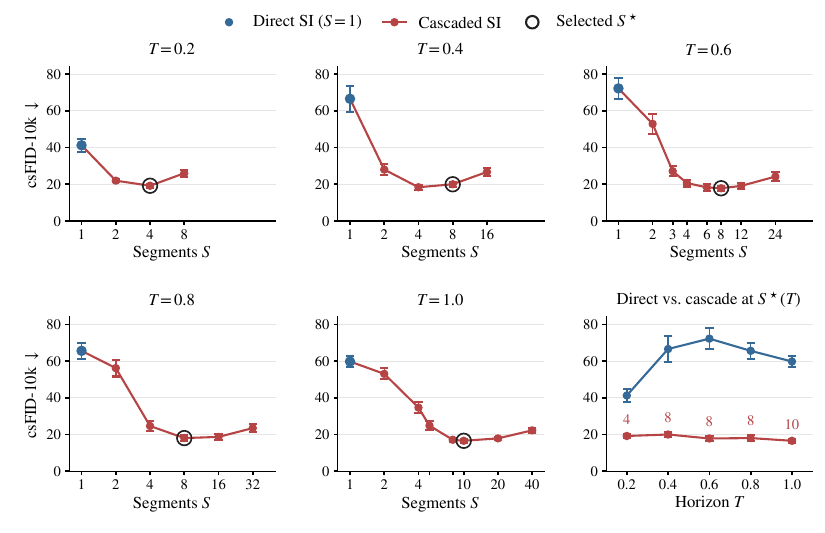}
\caption{Test csFID-10k is shown for five horizons, with Direct SI corresponding to $S=1$.
The black circles mark the validation-selected $S^\star$. The bottom-right panel compares
Direct SI with Cascaded SI at $S^\star(T)$ across
horizons. Values are means across three training seeds
and ten test conditions. Error bars show SEM across
conditions after averaging over seeds. Plotted numerical values are given in Table~\ref{tab:euler-horizon-depth}.}
\label{fig:euler-horizon-depth}
\end{figure}

\begin{table}[h!]
\centering
\caption{\textbf{Euler forecasting performance across
horizons.} Test csFID-10k (mean $\pm$ SEM across ten test conditions, each
averaged over three training seeds).}
\label{tab:euler-horizon-depth}
\small
\begin{tabular}{ccccc}
\toprule
$T$ & Direct SI & Cascaded SI & $S^\star$ & $T/S^\star$ \\
\midrule
0.2 & 41.3 $\pm$ 3.5 & \textbf{19.2} $\pm$ 0.9 & 4  & 0.050 \\
0.4 & 66.6 $\pm$ 7.0 & \textbf{20.0} $\pm$ 1.4 & 8  & 0.050 \\
0.6 & 72.3 $\pm$ 5.7 & \textbf{17.9} $\pm$ 1.5 & 8  & 0.075 \\
0.8 & 65.6 $\pm$ 4.4 & \textbf{18.1} $\pm$ 1.5 & 8  & 0.100 \\
1.0 & 59.8 $\pm$ 3.1 & \textbf{16.6} $\pm$ 1.0 & 10 & 0.100 \\
\bottomrule
\end{tabular}
\end{table}

Cascading improves performance at every tested horizon.
The advantage increases from approximately twofold at
$T=0.2$ to three- to fourfold at longer horizons.
The selected depth generally increases with $T$, from
$S^\star=4$ to $S^\star=10$, while the segment duration
$T/S^\star$ remains between 0.05 and 0.1.
The depth sweeps in Fig.~\ref{fig:euler-horizon-depth}
also show that performance deteriorates when too many
segments are used. The minima are relatively broad.

Absolute csFID values need not increase monotonically with $T$ (see performance of direct SI in Figure~\ref{fig:euler-horizon-depth}), since each horizon has its own reference ensembles, so csFID-10k compares methods at a fixed $T$, not one method across $T$.

For Direct SI at $T = 0.4$, the selected learning rate ($10^{-5}$) lies at the boundary of searched hyperparameters so a
wider search might improve this baseline. However, a large improvement is unlikely since validation performance is nearly flat in the learning rate (less than 2\% spread among the ten best
configurations, which cover $10^{-5}$ to $10^{-3}$), while the selected cascade is about six times better on the same criterion. 

\subsubsection{Anchoring the initial observation.}
\label{app:anchoring}
To test whether retaining the initial observation mitigates the deterioration
of Cascaded SI at large depths, we conduct the Euler and passive-scalar depth
sweeps with an \emph{anchored} cascade. The unanchored transition conditions
on the generated state $X_{t_i}$ while the anchored transition additionally
conditions on the original observation $X_0$ at every segment. We use the
same training data and U-Net architecture as in the corresponding benchmark,
with an additional conditioning input for the anchored model. Hyperparameters,
checkpoints, and sampler resolutions are selected separately using the
benchmark validation procedure. All curves are evaluated on the same ten
test conditions and conditional reference ensembles, with csFID-10k computed
from $10{,}000$ samples per condition.

\begin{figure}[h!]
\centering
\includegraphics[width=\linewidth]{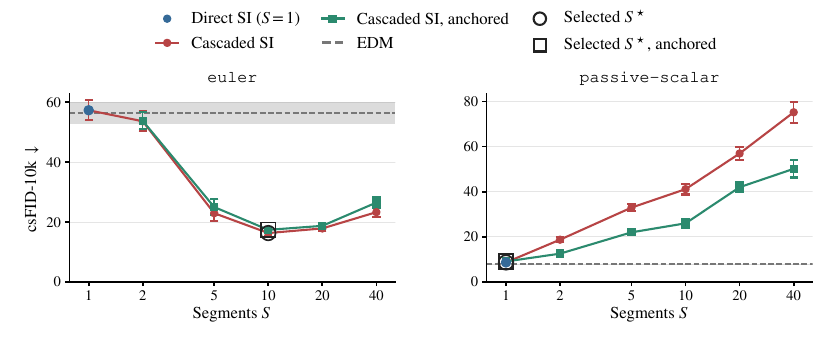}
\caption{\textbf{Retaining the initial observation.} Test csFID-10k against cascade depth for Euler
(left) and passive scalar (right). The anchored cascade receives the original observation $X_0$ at every segment, in addition to the generated current state. Error bars are SEM across ten test conditions; markers indicate the depth selected by validation for each method. Direct SI and EDM are shown as
references.}
\label{fig:anchored_depth}
\end{figure}

\begin{table}[h!]
\caption{\textbf{Anchored and unanchored depth sweeps.}
Test csFID-10k (mean $\pm$ SEM over ten conditions).
The anchored $S=1$ models were not test-scored.}
\label{tab:anchored_depth}
\centering
\footnotesize
\setlength{\tabcolsep}{4pt}
\begin{tabular}{@{}rcccc@{}}
\toprule
& \multicolumn{2}{c}{\euler} & \multicolumn{2}{c}{\scalar} \\
$S$ & Unanchored & Anchored & Unanchored & Anchored \\
\midrule
1  & $57.26\pm3.35$ & -- & $8.67\pm0.28$ & -- \\
2  & $53.60\pm3.28$ & $53.73\pm2.87$ & $18.75\pm1.17$ & $12.56\pm0.72$ \\
5  & $22.94\pm2.55$ & $24.99\pm2.61$ & $32.97\pm1.61$ & $22.01\pm0.98$ \\
10 & $16.32\pm1.14$ & $17.39\pm0.86$ & $41.14\pm2.36$ & $26.00\pm1.96$ \\
20 & $17.86\pm0.87$ & $18.73\pm1.01$ & $56.95\pm2.79$ & $42.01\pm2.20$ \\
40 & $23.29\pm1.50$ & $26.49\pm1.72$ & $75.21\pm4.74$ & $50.15\pm3.85$ \\
\bottomrule
\end{tabular}
\end{table}

The results differ substantially across systems (Fig.~\ref{fig:anchored_depth}).
On \euler, anchoring gives no improvement in csFID at any tested depth. Validation selects $S^\star=10$ for both
variants. In contrast, on \scalar, anchoring improves all tested depths $S\geq2$ on all ten test conditions. At $S=10$, csFID decreases from $41.14\pm2.36$ to $26.00\pm1.96$, and at $S=40$ from
$75.21\pm4.74$ to $50.15\pm3.85$. The anchored curve nevertheless
increases with depth, and validation selects $S^\star=1$ for both variants. 

To understand why anchoring helps on the passive-scalar dataset but not on the Euler dataset, we provide a theoretical analysis of learned anchored transitions that studies the distinction
between available information and learnable dependence in the rotation--OU model in Appendix~\ref{app:anchored-ou-theory} and
Appendix~\ref{app:ou-anchoring-experiment}.

\subsubsection{Sensitivity to the unresolved-scale prior}
\label{app:dependent-prior}

To test whether the advantage of Cascaded SI depends on
independently sampled observed and unresolved initial states,
we constructed an additional SQG dataset with a dependent
fine-scale prior. We kept the observed initial states, SQG
dynamics, and total hidden-state amplitude unchanged, but
modified how the unresolved Fourier modes are sampled at
$t=0$.

Specifically, we constructed a Markovian prior across
spatial scales. Starting from the observed coarse state
$X_0$ ($|k|\leq 8$), we generated two bands of hidden modes.
In the first band, $H_1$ ($8<|k|\leq16$), random fluctuations
have larger local variance where the observed coarse field
has strong gradients. In the second band, $H_2$
($16<|k|\leq32$), the local variance depends on the
fluctuations in $H_1$, rather than directly on $X_0$.
The construction therefore satisfies
\[
p(H_2\mid H_1,X_0)=p(H_2\mid H_1).
\]
Thus, the unresolved scales are statistically dependent
on the observed state, with dependence transmitted across
successive frequency bands. We used a modulation floor of
$0.5$ and retained the original hidden-state RMS of $0.1$.
The dependent prior is used only at initialization;
subsequent evolution follows the original SQG equations.

We repeated the benchmark protocol, independently tuning
Direct SI and Cascaded SI on the new dataset, with depths
$S\in\{1,2,5,10,20\}$. On the ten test conditions,
Cascaded SI achieves a csFID-10k of $53.18\pm 15.63$
at the validation-selected depth $S^\star=5$, compared
with $141.93\pm 23.46$ for Direct SI (mean $\pm$ SEM
across test conditions; one training seed). Cascaded SI
performs better on all ten conditions.

\section{How cascading changes forecast error}
\label{app:forecast-theory}

A forecast over a fixed horizon $T$ can be learned directly, in a single
transition, or through a cascade of shorter transitions. The experiments in
the main text show that shorter transitions can substantially improve the
conditional forecast, but that the improvement can stop as the cascade
becomes deeper. In this section, we study two mechanisms that can produce
this behavior. First, a fixed approximation class may represent a short
transition well while failing to represent the dependence that develops over
a longer interval. Composing several simple maps can then represent
dependence that a single map in the same class cannot. Second, under partial
observation, repeatedly conditioning only on the current observed state can
discard information carried by the unresolved state. We study these two
effects separately before combining them in the rotation model from the main text, in which a hidden Ornstein--Uhlenbeck process drives the dynamics.

To keep the analysis explicit, we use forecast classes built from polynomial
functions of the observed state. These provide a tractable setting in which
we can calculate when the dependence of the forecast on its input leaves a
fixed function class and how composing shorter fitted transitions changes
the resulting error. We then study exact short conditional kernels to isolate
the effect of repeated conditioning under partial observation. Finally, we analyze both effects together in the
rotation--OU model, giving an
example in which an intermediate cascade depth is optimal. The memory effect
that appears under partial observation is classical in optimal prediction
\cite{chorin2002optimal,ma2019}.

\textbf{Notation.} In this appendix, the full state is denoted by $Z_t$ and its unobserved initial component by $\eta$; in the main text, these are $Y_t$ and $Z_0$.

\subsection{When a fixed approximation class becomes insufficient}

Our first goal is to understand how approximation error appears when the
conditional forecast is restricted to a fixed class of functions. We begin
with a general model of a partially observed dynamical system and identify
the first term in the short-time expansion of the forecast that cannot be
represented by the chosen class.

Let $Z_t\in\R^d$ denote the full physical state. We assume that it evolves
according to the smooth autonomous system
\[
\dot Z_t=V(Z_t),
\]
where $V:\R^d\to\R^d$. We do not observe the full state $Z_t$. Instead, we
observe
\[
X_t=PZ_t\in\R^m,
\]
where $P:\R^d\to\R^m$ is a fixed observation map.

Let $\chi:\R^m\to\R^p$ specify the quantity that we wish to forecast from
the observed state. For example, $\chi$ may select a single observed Fourier
coefficient, a collection of observed coefficients, or the entire observed
state, in which case $\chi=\operatorname{Id}$. For each forecast horizon
$t$, define
\begin{equation}
(\mathcal T_t\chi)(x)
=
\E\!\left[\chi(X_t)\mid X_0=x\right].
\label{eq:conditional-forecast}
\end{equation}
Thus, $(\mathcal T_t\chi)(x)$ is the conditional mean of the future
quantity $\chi(X_t)$ given the observed initial state $X_0=x$.

Because only part of the physical state is observed, the operators
$\mathcal T_t$ need not satisfy the semigroup identity
\[
\mathcal T_{t+s}\chi
=
\mathcal T_t(\mathcal T_s\chi).
\]
Indeed, the unresolved variables may carry information from one time
interval to the next that cannot be recovered from the current observed
state alone. This issue will become important when we study cascades under
partial observation below.

We first examine how the conditional forecast changes over a short physical
time interval. For any smooth function $g:\R^d\to\R^p$ of the full state,
the chain rule gives
\[
\frac{\mathrm d}{\mathrm dt}g(Z_t)
=
Dg(Z_t)V(Z_t).
\]
We therefore define the derivative of $g$ along the full dynamics by
\[
(\mathcal L_Vg)(z)
=
Dg(z)V(z).
\]
Higher-order derivatives along the dynamics are defined recursively by
\[
\mathcal L_V^0g=g,
\qquad
\mathcal L_V^{j+1}g
=
\mathcal L_V(\mathcal L_V^jg).
\]
It follows that
\[
\left.
\frac{\mathrm d^j}{\mathrm dt^j}g(Z_t)
\right|_{t=0}
=
\mathcal L_V^jg(Z_0).
\]

We apply this identity to the full-state function
$g=\chi\circ P$. After conditioning on the observed initial state, define
\begin{equation}
a_{j,\chi}(x)
=
\E\!\left[
\mathcal L_V^j(\chi\circ P)(Z_0)
\mid X_0=x
\right].
\label{eq:conditional-lie}
\end{equation}
Each $a_{j,\chi}$ is a function of the observed initial state $x$.
However, the derivative inside the conditional expectation is computed
using the full state $Z_0$ and the full dynamics $V$. The conditional
expectation averages this derivative over the unresolved components of
$Z_0$.

The theorem below gives precise conditions under which the functions
$a_{j,\chi}$ are the coefficients in the short-time expansion of
$\mathcal T_t\chi$. It also controls the error made by truncating that
expansion.

Let
\[
\mu=\operatorname{Law}(X_0)
\]
denote the distribution of the observed initial state. Choose a closed
linear space
\[
\mathcal F\subset L^2(\mu;\R^p)
\]
representing the fixed class of forecasting functions under consideration.
The class $\mathcal F$ is chosen before selecting the forecast horizon and
before fitting a predictor to future targets. Since $\mathcal F$ is closed,
let $\Pi_{\mathcal F}$ denote the orthogonal projection from
$L^2(\mu;\R^p)$ onto $\mathcal F$.

We define $n_\star$ to be the first order for which the corresponding
coefficient cannot be represented by $\mathcal F$:
\[
n_\star
=
\min\left\{
j\geq0:
a_{j,\chi}\notin\mathcal F
\right\}.
\]
If every coefficient $a_{j,\chi}$ belongs to $\mathcal F$, we set
\[
n_\star=\infty.
\]
In particular, if $n_\star=n<\infty$, then
\[
a_{0,\chi},\ldots,a_{n-1,\chi}\in\mathcal F,
\qquad
a_{n,\chi}\notin\mathcal F.
\]
Thus, $a_{n,\chi}$ is the first coefficient in the short-time expansion
that cannot be represented by the chosen approximation class.

\begin{theorem}
\label{thm:first-escape}
Let $t_0>0$ and suppose that $n=n_\star<\infty$. Assume that, almost
surely, the trajectory $Z_t$ is defined on $[0,t_0]$ and the map
\[
t\longmapsto\chi(PZ_t)
\]
is $(n+1)$ times continuously differentiable on this interval. Assume also
that
\[
\mathbb E \left[ \left| \mathcal L_V^j(\chi\circ P)(Z_0)\right|^2 \right] <\infty
\qquad
\text{for every }j=0,\ldots,n.
\]
Define
\[
M_{n+1}
:=
\sup_{0\leq u\leq t_0}
\left(
\E\left[
\left|
\mathcal L_V^{n+1}(\chi\circ P)(Z_u)
\right|^2
\right]
\right)^{1/2},
\]
and assume that $M_{n+1}<\infty$. Finally, let
\[
c_n
=
\left\|
(I-\Pi_{\mathcal F})a_{n,\chi}
\right\|_{L^2(\mu)}.
\]
Then $c_n>0$ and, for every $t\in[0,t_0]$,
\begin{equation}
\left|
\operatorname{dist}_{L^2(\mu)}
\left(\mathcal T_t\chi,\mathcal F\right)
-
\frac{c_nt^n}{n!}
\right|
\leq
\frac{M_{n+1}t^{n+1}}{(n+1)!}.
\label{eq:first-escape}
\end{equation}
\end{theorem}

\begin{proof}[Proof of Theorem~\ref{thm:first-escape}]
Taylor's integral formula along each trajectory gives, with
$g=\chi\circ P$,
\[
 g(Z_t)=\sum_{j=0}^n\frac{t^j}{j!}\mathcal L_V^jg(Z_0)
       +\frac1{n!}\int_0^t(t-u)^n\mathcal L_V^{n+1}g(Z_u)\dd u.
\]
Condition on $X_0$ and use Minkowski's inequality and the $L^2$ contraction
of conditional expectation. The conditional remainder has norm at most
$M_{n+1}t^{n+1}/(n+1)!$. Applying $I-\Pi_{\mathcal F}$ removes all terms
before $n$ and does not increase the remainder. The reverse triangle
inequality proves \eqref{eq:first-escape}. Since $\mathcal F$ is closed
and $a_{n,\chi}\notin\mathcal F$, its distance $c_n$ is positive.
\end{proof}

\subsection{A quadratic dynamical model and a fixed forecast class}

We now apply the general result above to a concrete class of dynamical
systems. We consider quadratic dynamics, forecast the entire observed
state, and use quadratic functions of the initial observation as our fixed
forecast class. In the notation of the previous subsection, this means
that
\[
\chi=\operatorname{Id}
\qquad\text{and}\qquad
\mathcal F=\PP_2.
\]
Our goal is to show that, even when the full dynamics are quadratic, the
conditional mean can develop a nonzero cubic dependence on the observed
initial state.

This model isolates the nonlinear mechanism relevant to the active-scalar
systems in our benchmark. Finite-dimensional Galerkin versions of Euler,
for example, contain quadratic interactions between modes. These
interactions can make the observed forecast depend on increasingly
complicated combinations of the initial variables, and can transfer
information between observed and unresolved modes. The result below is
general for quadratic systems satisfying the stated assumptions. We also
give Lorenz--96 as a standard additional example, although it is not part
of the benchmark.

Let the full physical state satisfy
\begin{equation}
 \dot Z_t=V(Z_t),\qquad
 V(z)=Q(z)+Az+f,\qquad Z_t\in\R^d,
 \label{eq:primitive}
\end{equation}
where $Q$ is homogeneous and quadratic, $Az$ is linear, and $f$ is
constant. We assume that the quadratic term satisfies
\begin{equation}
 z^\top Q(z)=0
 \qquad
 \text{for every }z\in\R^d.
 \label{eq:quadratic-conservation}
\end{equation}
This means that the quadratic term alone does not change $|Z_t|^2$.
Indeed,
\[
\frac{\mathrm d}{\mathrm dt}|Z_t|^2
=
2Z_t^\top Q(Z_t)
=
0
\]
when the dynamics contains only $Q$. This property holds in the examples
that motivate the analysis: the quadratic term conserves enstrophy in a
Galerkin truncation of two-dimensional Euler and the squared norm in
Lorenz--96. The linear term $Az$ and the forcing term $f$ need not preserve
this quantity. We use Eq.~\eqref{eq:quadratic-conservation} both to control
the growth of the trajectories and to prove that the cubic term identified
below is nonzero.

As in the previous subsection, we observe
\[
X_t=PZ_t\in\R^m.
\]
We now assume that the rows of $P$ are orthonormal:
\[
PP^\top=I_m.
\]
Define
\[
E=P^\top.
\]
Then
\[
PE=I_m
\qquad\text{and}\qquad
|Ex|=|x|\quad\text{for every }x\in\R^m.
\]
Thus, $E$ maps an observed state $x$ back into the full state space while
preserving its observed coordinates and its norm.

We assume that the initial state can be written as
\begin{equation}
 Z_0=EX_0+\eta,\qquad
 P\eta=0,\qquad
 \E|Z_0|^8<\infty.
 \label{eq:initial-ensemble}
\end{equation}
Here, $EX_0$ is the observed part of the initial state and $\eta$ is the
unresolved part. The condition $P\eta=0$ means that $\eta$ is invisible
in the observation $X_0=PZ_0$. We assume that $X_0$ and $\eta$ are
independent. This matches the coarse--fine initialization used in the
Euler benchmark, where the observed and hidden Fourier modes are sampled
independently. Independence is assumed only at the initial time; the
dynamics may couple the two parts at later times. The eighth-moment
condition is used to control the remainder in the short-time expansion.

We keep the notation
\[
\mu=\operatorname{Law}(X_0)
\]
for the distribution of the observed initial state. We assume that $\mu$
has a density that is strictly positive on a nonempty open set. This
ensures that a polynomial which is nonzero as a function cannot vanish
$\mu$-almost everywhere.

Let $\Phi_t(z)$ denote the full state at time $t$ when the initial state is
$z$. Given an observed initial state $X_0=x$, define
\begin{equation}
K_t(x,\dd x')
=
\operatorname{Law}\!\left(P\Phi_t(Z_0)\mid X_0=x\right).
\label{eq:forecast-kernel}
\end{equation}
Thus, $K_t(x,\cdot)$ is the conditional distribution of the future
observed state $X_t$. Its conditional mean is
\begin{equation}
m_t(x)
=
\int x'K_t(x,\dd x')
=
\E[X_t\mid X_0=x].
\label{eq:forecast}
\end{equation}
Under squared error, $m_t$ is the best possible forecast of $X_t$ based
only on $X_0$. Since it predicts $X_t$ from $X_0$ over the full interval,
without using intermediate physical states, it is the direct conditional
mean forecast in this analysis.

We now restrict this direct forecast to a fixed class of quadratic
functions. Let $\PP_2$ be the space of vector-valued polynomials
$p:\R^m\to\R^m$ of total degree at most $2$, and let $\Pi_2$ denote the
orthogonal projection onto $\PP_2$ in $L^2(\mu)$. Define
\begin{equation}
 e_2(t)
 =
 \inf_{p\in\PP_2}
 \norm{m_t-p}_{L^2(\mu)}
 =
 \norm{m_t-\Pi_2m_t}_{L^2(\mu)}.
 \label{eq:e2}
\end{equation}
Thus, $e_2(t)$ measures how well a quadratic function of $X_0$ can
represent $m_t$ when the whole physical interval is predicted in a single
transition.

This is the quantity that appears in
Theorem~\ref{thm:first-escape} when
\[
\chi=\operatorname{Id}
\qquad\text{and}\qquad
\mathcal F=\PP_2.
\]
Indeed,
\[
(\mathcal T_t\operatorname{Id})(x)
=
\E[X_t\mid X_0=x]
=
m_t(x),
\]
and hence
\[
e_2(t)
=
\operatorname{dist}_{L^2(\mu)}
\left(\mathcal T_t\operatorname{Id},\PP_2\right).
\]

The square $e_2(t)^2$ also has a direct interpretation in terms of
prediction error. For any quadratic predictor $p\in\PP_2$,
\[
\E\!\left[|X_t-p(X_0)|^2\right]
=
\E\!\left[|X_t-m_t(X_0)|^2\right]
+
\|m_t-p\|_{L^2(\mu)}^2.
\]
Taking the infimum over $p\in\PP_2$ gives
\[
\inf_{p\in\PP_2}
\E\!\left[|X_t-p(X_0)|^2\right]
=
\E\!\left[|X_t-m_t(X_0)|^2\right]
+
e_2(t)^2.
\]
The first term is the prediction error that remains because $X_0$ does not
fully determine $X_t$. The second term is the additional error caused by
restricting the conditional mean forecast to quadratic functions.

The next subsection identifies the first part of the conditional mean that
lies outside $\PP_2$. We then apply
Theorem~\ref{thm:first-escape} to obtain an explicit short-time estimate
for $e_2(t)$.

\subsection{The cubic term that quadratic forecasts cannot represent}

We now identify the first coefficient in the short-time expansion of the
direct conditional mean $m_t$ that can lie outside $\PP_2$. This will
allow us to apply Theorem~\ref{thm:first-escape}.

For $\chi=\operatorname{Id}$, the zeroth-order coefficient is
\[
a_{0,\operatorname{Id}}(x)=x,
\]
which belongs to $\PP_2$. The first-order coefficient is
\[
a_{1,\operatorname{Id}}(x)
=
\E\!\left[
PV(Z_0)\mid X_0=x
\right].
\]
Since $V$ is quadratic and $Z_0=Ex+\eta$, this is a polynomial of degree
at most two in $x$. Here we use the independence of $X_0$ and $\eta$,
which ensures that the moments of $\eta$ appearing in the conditional
expectation do not depend on $x$. Therefore,
\[
a_{1,\operatorname{Id}}\in\PP_2.
\]

The second-order coefficient is
\[
a_{2,\operatorname{Id}}(x)
=
\E\!\left[
P\,DV(Z_0)V(Z_0)\mid X_0=x
\right].
\]
Unlike the previous coefficients, it can contain terms of degree three in
$x$. Its degree-three part comes from applying the quadratic term $Q$
twice and is given by
\begin{equation}
C_3(x)
=
P\,DQ(Ex)Q(Ex).
\label{eq:cubic-term}
\end{equation}
Here, $DQ(Ex)$ is the derivative of $Q$ evaluated at $Ex$, and the final
application of $P$ keeps only the observed coordinates. All the remaining
terms in $a_{2,\operatorname{Id}}$ have degree at most two. Therefore,
\[
a_{2,\operatorname{Id}}
=
C_3+b_2
\qquad
\text{for some }b_2\in\PP_2,
\]
and hence
\[
(I-\Pi_2)a_{2,\operatorname{Id}}
=
(I-\Pi_2)C_3.
\]

We measure the part of $C_3$ that cannot be represented by a quadratic
function by
\begin{equation}
 c_3
 =
 \norm{(I-\Pi_2)C_3}_{L^2(\mu)}.
 \label{eq:cubic-witness}
\end{equation}
Since the zeroth- and first-order coefficients belong to $\PP_2$, proving
that $c_3>0$ will show that $n_\star=2$. The subscript in $c_3$ refers to
the fact that $C_3$ has polynomial degree three, even though it appears at
second order in physical time.

\begin{theorem}
\label{thm:coarse-cubic}
Under the assumptions of the previous subsection, suppose that
$Q(Ex)$ is not identically zero as a function of $x$. Then
\[
n_\star=2
\qquad\text{and}\qquad
c_3>0.
\]
Moreover, for every fixed $t_0>0$, there is a finite and explicitly
bounded constant $M_3$ such that, for every $t\in[0,t_0]$,
\begin{equation}
\left|
e_2(t)-\frac{c_3t^2}{2}
\right|
\leq
\frac{M_3t^3}{6}.
\label{eq:cubic-error}
\end{equation}
Consequently,
\begin{equation}
e_2(t)^2
=
\frac{c_3^2}{4}t^4+O(t^5).
\label{eq:quartic-signature}
\end{equation}
The coefficient $c_3$ depends on the quadratic dynamics $Q$, the
observation map $P$, and the distribution $\mu$ of the observed initial
state. It does not depend on the distribution of the independent
unresolved initial state $\eta$. The unresolved variables can still
affect the lower-degree terms and the remainder.
\end{theorem}

The condition $Q(Ex)\not\equiv0$ has a simple meaning: an initial state
containing only observed modes must activate at least one quadratic
interaction. To see why this implies $c_3>0$, differentiate
\[
z^\top Q(z)=0
\]
in the direction $Q(z)$. This gives
\[
|Q(z)|^2+z^\top DQ(z)Q(z)=0.
\]
Setting $z=Ex$ and using $E=P^\top$ gives
\begin{equation}
x^\top C_3(x)
=
-|Q(Ex)|^2.
\label{eq:coarse-certificate}
\end{equation}
If $Q(Ex)$ is nonzero for some $x$, the right-hand side is strictly
negative at that point. Therefore, $C_3$ is not identically zero.

Finally, $C_3$ is a homogeneous polynomial of degree three, whereas every
element of $\PP_2$ has degree at most two. Since $\mu$ has a density that
is positive on a nonempty open set, a nonzero cubic polynomial cannot
agree with a quadratic polynomial $\mu$-almost everywhere. It follows
that
\[
c_3>0.
\]

Notice that the condition $Q(Ex)\not\equiv0$ does not require
\[
PQ(Ex)\not\equiv0.
\]
The first quadratic interaction may therefore lie entirely in the
unresolved coordinates. Identity~\eqref{eq:coarse-certificate} shows that,
after the dynamics acts again, this interaction still produces a nonzero
term in the observed forecast.

The conclusion $n_\star=2$ applies to the forecast of the entire observed
state. For a scalar quantity $\chi(x)=u^\top x$, the corresponding term is
$u^\top C_3(x)$. Therefore, the same conclusion holds only if
\[
u^\top C_3\not\equiv0.
\]
Thus, some individual coordinates, including conserved or uncoupled
quantities, may not contain this term.

For a general initial distribution, evaluating $c_3$ requires an
$L^2(\mu)$ projection. When the observed initial state is Gaussian, this
projection can be computed explicitly; see
Corollary~\ref{cor:gaussian-tensor} below. The argument below also shows
that the conclusion is unchanged by invertible affine changes of the
observed coordinates.

\begin{corollary}[An explicit coefficient for Gaussian conditioning states]
\label{cor:gaussian-tensor}
If $X_0\sim\N(a,\Sigma)$ with $\Sigma>0$, write
\begin{equation}
 [C_3(\Sigma^{1/2}z)]_k
   =\sum_{i,j,\ell=1}^m T_{kij\ell}z_i z_j z_\ell,
 \label{eq:tensor-proof}
\end{equation}
with $T$ symmetric in its last three indices. Then
\begin{equation}
 c_3^2=6\sum_{k,i,j,\ell}T_{kij\ell}^2,
 \qquad e_2(t)^2=\frac32\norm{T}_F^2t^4+O(t^5).
 \label{eq:tensor-coefficient-proof}
\end{equation}
For a fixed unit output direction $u$, the corresponding coefficient is
obtained by replacing $T_{kij\ell}$ by $\sum_k u_kT_{kij\ell}$.
\end{corollary}

\begin{proof}[Proof of Corollary~\ref{cor:gaussian-tensor}]
Write $X_0=a+\Sigma^{1/2}Z$, where $Z\sim\N(0,I_m)$. An invertible affine
coordinate change preserves polynomial degree. The shift by $a$ changes
$C_3$ only by terms of degree at most two after its leading homogeneous
part is extracted. For standard Gaussian inputs, the part of
$Z_iZ_jZ_\ell$ orthogonal to degrees at most two is
\[
 :Z_iZ_jZ_\ell:
   =Z_iZ_jZ_\ell-\delta_{ij}Z_\ell
                         -\delta_{i\ell}Z_j-\delta_{j\ell}Z_i.
\]
Their covariance is the sum of the six matchings between the three
indices in the first factor and those in the second factor. One direct
derivation differentiates the identity
\[
 \E\bigl[e^{s^\top Z-|s|^2/2}e^{t^\top Z-|t|^2/2}\bigr]=e^{s^\top t}
\]
three times in $s$ and three times in $t$, at zero. Contracting against
the symmetric tensor $T$ gives
\[
 \E\left|\sum_{i,j,\ell}T_{kij\ell}:Z_iZ_jZ_\ell:\right|^2
     =6\sum_{i,j,\ell}T_{kij\ell}^2.
\]
Summation in $k$ proves the first identity;
Theorem~\ref{thm:coarse-cubic} gives the stated leading term in
$e_2(t)^2$. Contracting the output index gives the directional version.
\end{proof}

\textbf{Affine changes of observed coordinates.}

Let $B$ be invertible, $\widetilde\mu=(B\cdot+b)_\#\mu$, and
$\widetilde f(u)=f(B^{-1}(u-b))$. Composition with $B^{-1}(\cdot-b)$ is
an isometry from $L^2(\mu)$ onto $L^2(\widetilde\mu)$ and a bijection of
degree-$r$ polynomial spaces. Taking the infimum over those spaces gives
exact equality of the approximation errors, including their leading
coefficients. If outputs are additionally transformed to $Cf+d$, with
$C$ invertible and $r\ge0$, candidates correspond as
$p\mapsto Cp+d$. The inequalities
\[
 \sigma_{\min}(C)|f-p|\le |C(f-p)|\le\sigma_{\max}(C)|f-p|
\]
give the bounds after integration and minimization. An isometric
output transformation gives equality. Thus membership of the vector
coefficients, and their first escape index, is unchanged. This argument
transports the input law and output metric consistently; arbitrary
rescaling while holding a different law fixed is a different comparison.

\textbf{Examples and comparisons.}

The results above apply differently to the systems considered in the
paper and to related examples:

\begin{itemize}

\item \textbf{Galerkin Euler.}
The quadratic vorticity interaction conserves enstrophy and therefore
satisfies Eq.~\eqref{eq:quadratic-conservation}. Any observed subspace for
which $Q(Ex)\not\equiv0$ has the nonzero cubic term described above.

\item \textbf{Surface quasi-geostrophic flow.}
A finite Galerkin truncation of SQG also has quadratic transport. The
conclusion of Theorem~\ref{thm:coarse-cubic} applies when its assumptions
hold, in particular when $Q\circ E\not\equiv0$ for the chosen observed
modes.

\item \textbf{Lorenz--96.}
The quadratic part conserves the squared Euclidean norm, while damping and
forcing appear in the lower-order terms. For $d\geq4$, the cubic term is
nonzero when the full state is observed. Under partial observation, the
result applies whenever $Q(Ex)\not\equiv0$. Lorenz--96 is included here as
a standard quadratic comparison rather than as one of our benchmark
systems.

\item \textbf{Conditional rotation.}
After including the conditioning variable in the state, the vector field is
\[
(0,\omega cJy),
\]
which is quadratic and preserves the norm of $y$. Its finite-time
conditional mean is
\[
R(\omega tc)y,
\]
which has nonpolynomial dependence on the conditioning variable $c$.
This example shows explicitly how repeated quadratic interactions can
produce a much more complicated finite-time forecast.

\item \textbf{Prescribed passive scalar.}
This benchmark provides a linear comparison. Its evolution is linear in
the initial scalar field, so it does not create the same higher-degree
dependence. Under independent coarse--fine initialization, its conditional
mean remains affine at every forecast horizon.
\end{itemize}

The heterogeneous-wave benchmark is also linear in the full physical
state, although partial observation of that state still makes the forecast
probabilistic. The nonlinear Klein--Gordon benchmark contains a cubic
nonlinearity and is therefore not covered by the quadratic theorem. The
general first-escape result in Theorem~\ref{thm:first-escape} can still be
applied whenever its regularity and moment assumptions hold.

The passive-scalar statement is made precise by the following proposition.

\begin{proposition}
\label{prop:linear-control}
Suppose that the full state evolves linearly according to
\[
Z_t=U_tZ_0,
\]
where $U_t:\R^d\to\R^d$ is the linear solution operator at time $t$.
Assume that $X_0$ and $\eta$ are independent. Then
\[
m_t(x)
=
PU_tEx+PU_t\,\E[\eta]
\]
for every forecast horizon $t$. Hence both the affine and quadratic
approximation errors are zero. If $\eta$ is Gaussian, then the conditional
distribution of the future observed state is Gaussian, with an affine mean
and a covariance that does not depend on $x$.
\end{proposition}

The result follows by substituting $Z_0=EX_0+\eta$ into
$Z_t=U_tZ_0$ and conditioning on $X_0=x$; see the proof below.

\textbf{Proof of the cubic estimate and the linear comparison.}

Under \eqref{eq:quadratic-conservation}, the radial derivative of a
solution of \eqref{eq:primitive} is bounded above by
$\|A\|_{\rm op}|Z|+|f|$. Applying the same argument to the time-reversed
equation gives, for every finite real $t$,
\begin{equation}
 |\Phi_t(z)|\le e^{a|t|}(|z|+b|t|),\qquad
 a=\|A\|_{\rm op},\quad b=|f|.
 \label{eq:growth-proof}
\end{equation}
One can justify the radial inequality at zero by differentiating
$\sqrt{|Z|^2+\varepsilon}$ and sending $\varepsilon$ to zero.
Local existence and continuation for a polynomial vector field then give
global existence. For $A=f=0$, the norm is exactly conserved.

Set $\mathcal A=DV\,V$ and $\mathcal B=D\mathcal A\,V$. They are
polynomials of degree at most three and four. Along a trajectory they are
the second and third time derivatives of the state. Therefore
\begin{equation}
 P\Phi_t(z)=Pz+tPV(z)+\tfrac12t^2P\mathcal A(z)
       +\int_0^t\tfrac12(t-s)^2P\mathcal B(\Phi_s(z))\dd s.
 \label{eq:finite-taylor-proof}
\end{equation}
For a fixed $t_0>0$, the following is one explicit permissible bound:
\begin{align}
 K_3&=\sup_{z\in\R^d}\frac{|P\mathcal B(z)|}{(1+|z|)^4}<\infty,
 &G&=e^{at_0}(1+bt_0),\notag\\
 M_3&=K_3G^4\bigl[\E(1+|Z_0|)^8\bigr]^{1/2}.
 \label{eq:M3-proof}
\end{align}
All quantities are determined by the polynomial coefficients, $P$, the
initial eighth moment, and $t_0$. Indeed,
$1+|\Phi_s(z)|\le G(1+|z|)$ on this interval. Minkowski's inequality in
\eqref{eq:finite-taylor-proof} bounds the remainder in $L^2$ by
$M_3t^3/6$. Conditional expectation is an $L^2$ contraction, so
\begin{equation}
 \begin{aligned}
 m_t(x)&=x+t a_1(x)+\tfrac12t^2a_2(x)+\mathcal E_t(x),\\
 a_1(x)&=\E[PV(Z_0)\mid X_0=x],\\
 a_2(x)&=\E[P\mathcal A(Z_0)\mid X_0=x],
 \end{aligned}
 \label{eq:coarse-taylor-proof}
\end{equation}
with $\|\mathcal E_t\|_{L^2(\mu)}\le M_3t^3/6$.

\begin{proof}[Proof of Theorem~\ref{thm:coarse-cubic}]
Independence in \eqref{eq:initial-ensemble} implies $a_1\in\PP_2$.
The homogeneous cubic part of $DV(z)V(z)$ is $DQ(z)Q(z)$; all remaining
terms have degree at most two. Substituting $z=Ex+\eta$, every term with
at least one hidden factor has at most two visible factors. Its expectation
is therefore a polynomial of degree at most two. Thus
\[
 a_2(x)=C_3(x)+p_2(x),\qquad p_2\in\PP_2.
\]
Differentiate $z^\top Q(z)=0$ in direction $Q(z)$:
$|Q(z)|^2+z^\top DQ(z)Q(z)=0$. At $z=Ex$ this is
\eqref{eq:coarse-certificate}, because $E=P^\top$.
Hence $Q\circ E\not\equiv0$ implies $C_3\not\equiv0$.
A nonzero homogeneous cubic cannot agree with a quadratic polynomial on
an open set. Positivity of the density there and closedness of the
finite-dimensional polynomial subspace imply $c_3>0$.

Applying $I-\Pi_2$ to \eqref{eq:coarse-taylor-proof} leaves
\[
 (I-\Pi_2)m_t=\tfrac12t^2(I-\Pi_2)C_3+(I-\Pi_2)\mathcal E_t.
\]
The projection is a contraction; the reverse triangle inequality proves
\eqref{eq:cubic-error}. Squaring proves \eqref{eq:quartic-signature}.
For example $e_2(t)\ge c_3t^2/4$ whenever $M_3t\le3c_3/2$.
\end{proof}

\begin{proof}[Proof of Proposition~\ref{prop:linear-control}]
Substitute $Z_0=EX_0+\eta$ into $Z_t=U_tZ_0$ and condition on $X_0=x$.
Independence gives $\E[\eta\mid X_0=x]=\E\eta$. If $\eta$ is Gaussian,
$PU_t\eta$ is Gaussian with covariance
$PU_t\operatorname{Cov}(\eta)U_t^\top P^\top$, which is independent of
$x$.
\end{proof}

\textbf{Verification for the quadratic examples.}

For a finite Fourier--Galerkin truncation of two-dimensional Euler, the
vorticity field obeys $\partial_t\omega=-u\cdot\nabla\omega$ and
$\nabla\cdot u=0$. Consequently,
\[
\frac{\dd}{\dd t}\frac12\|\omega\|_{L^2}^2
=-\frac12\int u\cdot\nabla(\omega^2)\dd x=0.
\]
In orthonormal real Fourier coordinates, the Galerkin vector field is
homogeneous quadratic and satisfies $z^\top Q(z)=0$. The condition
$Q\circ E\not\equiv0$ still needs to be checked for the chosen observed
subspace. For example, observed modes $(1,1)$ and $(2,0)$ interact to
produce mode $(3,1)$ with a nonzero coefficient in a truncation that
retains this mode. The output of this first interaction may itself be
unresolved: Theorem~\ref{thm:coarse-cubic} only requires it to be nonzero
in the full Galerkin state.

For Lorenz--96 with cyclic indices $i=1,\ldots,d$, $d\geq4$, write
\[
V_i(z)=(z_{i+1}-z_{i-2})z_{i-1}-z_i+F,
\qquad Q_i(z)=(z_{i+1}-z_{i-2})z_{i-1}.
\]
Relabeling indices gives
\[
\sum_i z_iQ_i(z)
=\sum_i z_{i-1}z_i z_{i+1}
-\sum_i z_{i-2}z_{i-1}z_i=0.
\]
Since $Q\not\equiv0$, the certificate
\eqref{eq:coarse-certificate} proves that its full-state cubic term is
nonzero. A particular coarse projection requires the additional check
$Q\circ E\not\equiv0$. The damping and forcing have lower degree and
cannot cancel that term. For the conditional rotation example,
$Q(c,y)=(0,\omega cJy)$ is likewise quadratic and satisfies
$(c,y)^\top Q(c,y)=0$.

\subsection{Why a cascade can reduce approximation error}

We now isolate the first effect in the cascading trade-off: shorter
transitions can reduce approximation error. The previous subsection showed
that a single quadratic function may fail to represent the conditional
mean over the full forecast horizon. We now show that the same quadratic
class can give a more accurate endpoint forecast when it is fitted over a
shorter interval and then composed repeatedly.

Partial observation creates a separate difficulty because each stage of a
cascade may discard information carried by unresolved variables. We set
that issue aside in this subsection and assume that the full state is
observed. Thus there is no unresolved component, and the exact forecast
over an interval of length $h$ is the deterministic flow map
\[
z\longmapsto\Phi_h(z).
\]

Let
\[
B_R=\{z\in\R^d:|z|\leq R\},
\]
and let $\nu$ be a probability distribution on $B_R$ whose density,
relative to the uniform distribution, is bounded above and below by
positive constants. We use $\nu$ to measure approximation errors in this
subsection.

Let $\Pi_2^\nu$ denote the orthogonal projection in $L^2(\nu)$ onto the
space $\PP_2(\R^d;\R^d)$ of quadratic functions from the full state space
to itself. For a physical transition of length $h$, define
\[
F_h^{(2)}
=
\Pi_2^\nu\Phi_h.
\]
Thus, $F_h^{(2)}$ is the best quadratic approximation of the exact flow map
$\Phi_h$ under the distribution $\nu$. Its approximation error is
\[
e_2^{\mathrm{full}}(h)
=
\norm{(I-\Pi_2^\nu)\Phi_h}_{L^2(\nu)}.
\]
The superscript ``full'' distinguishes this error from $e_2(t)$, which was
defined earlier for a forecast based only on the observed state.

Fix a final forecast horizon $T$. A direct quadratic forecast uses the
single map
\[
F_T^{(2)}.
\]
A cascade of depth $S$ divides the same physical horizon into intervals of
length
\[
\delta=\frac{T}{S}
\]
and repeatedly applies the short-transition approximation
$F_\delta^{(2)}$. Its endpoint map is therefore
\[
\left(F_\delta^{(2)}\right)^{\circ S}.
\]
The following theorem compares these two forecasts. It shows that a single
quadratic map over the full horizon has a nonzero approximation error,
while the endpoint error obtained by composing the fitted short-transition
map converges to zero as the transition length decreases.

\begin{theorem}
\label{thm:composition}
Suppose first that
\[
V=Q\not\equiv0
\]
and that $Q$ satisfies Eq.~\eqref{eq:quadratic-conservation}. Then, for
every $T>0$,
\[
e_2^{\mathrm{full}}(T)>0.
\]
Therefore, the exact flow map $\Phi_T$ cannot be represented exactly by a
quadratic function.

There are constants $D_2,\Lambda_2<\infty$, independent of $S$, such that,
for all sufficiently large $S$,
\begin{equation}
\sup_{z\in B_R}
\left|
\left(F_\delta^{(2)}\right)^{\circ S}(z)-\Phi_T(z)
\right|
\leq
D_2
\frac{e^{\Lambda_2T}-1}{\Lambda_2}
\delta.
\label{eq:composition-bound}
\end{equation}
When $\Lambda_2=0$, the quotient is understood to be $T$.
Consequently, for sufficiently large $S$, the cascade of short quadratic
fits has smaller endpoint approximation error than every direct quadratic
function.

For the more general dynamics
\[
V(z)=Q(z)+Az+f,
\]
the same convergence bound holds on a forward-invariant ball. If
$Q\not\equiv0$, the direct approximation error is positive for every
sufficiently small nonzero $T$.
\end{theorem}

The bound in \eqref{eq:composition-bound} tends to zero as the transition
length $\delta$ tends to zero. Thus, composing short quadratic fits can
approach the exact endpoint map even though no single quadratic function
can represent that map exactly.

The same result holds for every fixed polynomial degree $r\geq2$, with an
endpoint-error bound of order $\delta^{r-1}$. The proof of this more
general statement, together with a finite-sample result for quadratic
fitting, is given below in the proof of
Theorem~\ref{thm:composition}.

\textbf{Proof of Theorem~\ref{thm:composition}.}

Write $A_n=\mathcal L_V^n\Id$, with $A_0(z)=z$.
Since $\deg V\le2$, $\deg A_n\le n+1$. The homogeneous component of
degree $n+1$ is $\mathcal L_Q^n\Id$.

\begin{lemma}
\label{lem:all-orders}
If $Q\not\equiv0$ is homogeneous quadratic and norm preserving, then
$\mathcal L_Q^n\Id\not\equiv0$ for every integer $n\ge1$.
\end{lemma}

\begin{proof}
Otherwise the $n$th time derivative of every $Q$-trajectory would vanish
identically. Each trajectory would be a polynomial in time of degree at
most $n-1$. It exists for all real time and has constant norm. Every
bounded real polynomial is constant, so all trajectories would be
constant. Their derivative at zero would imply $Q\equiv0$, a
contradiction.
\end{proof}

On a forward-invariant ball, Taylor's formula gives
\[
 \Phi_t=\sum_{n=0}^{r}\frac{t^n}{n!}A_n+\mathcal E_t,
 \qquad \sup_{B_R}|\mathcal E_t|
       \le\frac{t^{r+1}}{(r+1)!}\sup_{B_R}|A_{r+1}|.
\]
For $n<r$, $A_n\in\PP_r$. Lemma~\ref{lem:all-orders} shows that $A_r$
has a nonzero homogeneous component of degree $r+1$. Consequently
\begin{equation}
 \left|e_r^{\rm full}(t)-\frac{c_rt^r}{r!}\right|
 \le\frac{t^{r+1}}{(r+1)!}\sup_{B_R}|A_{r+1}|,
 \quad c_r=\norm{(I-\Pi_r^\nu)A_r}_{L^2(\nu)}>0.
 \label{eq:degree-r-gap-proof}
\end{equation}
This establishes the small-time claim for $V=Q+Az+f$, including
Lorenz--96. It uses a finite Taylor remainder, not convergence of an
infinite Lie series at the terminal horizon.

For the pure field $V=Q$, the direct gap is positive at every $T>0$.
Indeed, if $e_r^{\rm full}(T)=0$, the finite-dimensional space is closed,
so $\Phi_T=p$ for a polynomial $p$ on $B_R$, first almost everywhere and
then everywhere by continuity and the density assumption. Norm
conservation implies the polynomial identity $|p(z)|^2=|z|^2$ on all of
$\R^d$. If $p$ had degree greater than one, its leading squared norm could
not vanish. Hence $p$ is affine. But $\Phi_T(0)=0$ and
$D\Phi_T(0)=I$, so $p=\Id$. The second variational equation at the
equilibrium zero is
\[
 D^2\Phi_T(0)[a,b]=T D^2Q(0)[a,b]\not\equiv0,
\]
which contradicts $p=\Id$.

\textbf{Convergence of the fitted short-transition rule.}

Finite-dimensional norm equivalence gives a constant $C_{\Pi,r}$ with
\begin{equation}
 \|p\|_{C^1(B_{2R})}\le C_{\Pi,r}\|p\|_{L^2(\nu)},
 \qquad p\in\PP_r.
 \label{eq:polynomial-norm-proof}
\end{equation}
Here the norm on the right is a genuine norm on polynomial coefficients.
Let $M_r=\sup_{B_R}|A_r|$. The Taylor polynomial
\[
 T_{r-1,\delta}=\sum_{n=0}^{r-1}\frac{\delta^n}{n!}A_n
\]
belongs to $\PP_r$. The best $L^2(\nu)$ approximation satisfies
\[
 F_\delta^{(r)}=T_{r-1,\delta}
      +\Pi_r^\nu(\Phi_\delta-T_{r-1,\delta}),\qquad
 \|\Phi_\delta-T_{r-1,\delta}\|_{L^2(\nu)}
      \le M_r\delta^r/r!.
\]
For $0<\delta\le1$, permissible constants are
\begin{align}
 D_r&=(1+C_{\Pi,r})M_r/r!,\notag\\
 \Lambda_r&=\sum_{n=1}^{r-1}\frac1{n!}
               \sup_{B_{2R}}\|DA_n\|_{\rm op}
                   +C_{\Pi,r}M_r/r!.
 \label{eq:composition-constants-proof}
\end{align}
They yield
\[
 \sup_{B_R}|F_\delta^{(r)}-\Phi_\delta|\le D_r\delta^r,
 \qquad \Lip(F_\delta^{(r)}|_{B_{2R}})\le1+\Lambda_r\delta.
\]
Let $z_i=\Phi_{i\delta}(z_0)$ and
$\widehat z_i=(F_\delta^{(r)})^{\circ i}(z_0)$, starting in $B_R$.
As long as $\widehat z_i\in B_{2R}$,
\[
 |\widehat z_{i+1}-z_{i+1}|
 \le(1+\Lambda_r\delta)|\widehat z_i-z_i|+D_r\delta^r.
\]
Summing the recursion gives the bound in
Theorem~\ref{thm:composition}, with $r=2$; the same argument gives order
$\delta^{r-1}$ for any fixed $r\geq2$. If its right-hand side is at most
$R/2$, every iterate lies in $B_{3R/2}$, so induction justifies all the
estimates. This condition holds for sufficiently large $S$ because
$r\ge2$. Combining with the positive direct gaps above proves
Theorem~\ref{thm:composition}.

\textbf{Fitting error and a finite-sample bound.}

For $r=2$, let $\widehat F_\delta\in\PP_2$ and
$\eta_\delta=\|\widehat F_\delta-F_\delta^{(2)}\|_{L^2(\nu)}$.
If $\Lip(\widehat F_\delta|_{B_{2R}})\le1+\widehat\Lambda\delta$,
the same proof gives
\begin{equation}
 \sup_{B_R}|\widehat F_\delta^{\circ S}-\Phi_T|
 \le\frac{e^{\widehat\Lambda T}-1}{\widehat\Lambda}
       \left(D_2\delta+\frac{C_{\Pi,2}\eta_\delta}{\delta}\right),
 \label{eq:learned-composition}
\end{equation}
provided the right-hand side is at most $R/2$. For example,
$\eta_\delta\le C\delta^2$ implies the required stability with a constant
$\widehat\Lambda$ independent of $\delta\le1$, by
\eqref{eq:polynomial-norm-proof}. A small fitting error alone is therefore not enough to justify
\eqref{eq:learned-composition}: we also need to control the Lipschitz
constant of the fitted map.

We next bound $\eta_\delta$ when the map is fitted from finitely many
observations. Let $p=\binom{d+2}{2}$, and let
$\varphi_1,\ldots,\varphi_p$ be an orthonormal basis of the scalar
quadratic polynomials in $L^2(\nu)$. Write
$\varphi=(\varphi_1,\ldots,\varphi_p)^\top$ and set
$\kappa_\varphi=\sup_{B_R}|\varphi|^2$.
Suppose that we observe $n$ independent pairs
\[
 Z_i\sim\nu,\qquad U_i=\Phi_\delta(Z_i)+\zeta_i,
 \quad \E[\zeta_i\mid Z_i]=0,\quad
 \E[|\zeta_i|^2\mid Z_i]\le v_{\rm obs}^2.
\]
For every $\varepsilon\in(0,1)$, with probability at least
$1-4\kappa_\varphi p/n-\varepsilon$, the empirical Gram matrix is
invertible and least squares satisfies
\begin{equation}
 \eta_\delta\le
 2\sqrt{\frac{\kappa_\varphi(D_2^2\delta^4+v_{\rm obs}^2)}
 {n\varepsilon}}.
 \label{eq:finite-sample}
\end{equation}
To verify this, let $G_n=n^{-1}\sum_i\varphi_i\varphi_i^\top$.
Independence gives
$\E\|G_n-I\|_F^2\le\kappa_\varphi p/n$, so Markov's inequality bounds
the failure of $\|G_n-I\|_{\rm op}\le1/2$ by
$4\kappa_\varphi p/n$. The mean-zero matrix
\[
 Z_n=\frac1n\sum_i
 [\Phi_\delta(Z_i)-F_\delta^{(2)}(Z_i)+\zeta_i]\varphi_i^\top
\]
has second moment at most
$\kappa_\varphi(D_2^2\delta^4+v_{\rm obs}^2)/n$.
On the Gram event the coefficient error is $Z_nG_n^{-1}$, of norm at
most $2\|Z_n\|_F$. Orthonormality identifies its Frobenius norm with
the $L^2(\nu)$ norm of the prediction error. A second Markov inequality proves
\eqref{eq:finite-sample}.

In the noiseless case the fitting error is proportional to
$\delta^2/\sqrt n$ at fixed confidence. With fixed observation noise,
the propagated contribution can instead grow like
$v_{\rm obs}/(\delta\sqrt n)$. This result concerns independent
full-state regression pairs. Several transitions drawn from one trajectory
do not constitute independent pairs, and physical hidden-state uncertainty
is not identified with this observation-noise model.

\subsection{Connection to the stochastic-interpolant learner}

The previous subsection compared direct and cascaded forecasts at the
level of their endpoint maps. The experiments in the main text implement
these forecasts using stochastic interpolants. We now connect the
approximation results above to a specific stochastic-interpolant learner.
In the deterministic full-state setting of
Theorem~\ref{thm:composition}, this gives an explicit setting in which
Cascaded SI has smaller endpoint approximation error than Direct SI.

We begin by describing the stochastic-interpolant training problem. Fix a
physical transition length $h$. Given the initial observation $X_0=x$,
let
\[
C=X_h,
\qquad
C\mid X_0=x\sim K_h(x,\cdot),
\]
denote the future state that we wish to generate. The physical transition
length $h$ is different from the interpolation variable
$\tau\in[0,1]$ used by the stochastic interpolant. We use the following
Wiener stochastic interpolant:
\begin{equation}
Y_\tau
=
(1-\tau)x+\tau C
+\sigma_b(1-\tau)\sqrt{\tau}\,\xi,
\qquad
R_\tau
=
C-x-\sigma_b\sqrt{\tau}\,\xi,
\label{eq:wiener}
\end{equation}
where $\xi\sim\N(0,I)$ is independent of $(X_0,C)$ and $\sigma_b>0$
controls the amount of noise. At $\tau=0$, the interpolant starts from
$x$, while it approaches the endpoint $C$ as $\tau$ approaches $1$.

The pair $(Y_\tau,R_\tau)$ defines the regression problem used to train
the stochastic interpolant. Let $\pi$ denote the distribution from which
the interpolation time $\tau$ is sampled during training, independently
of $(X_0,C,\xi)$. For a candidate drift $b$, define the training objective
by
\begin{equation}
\mathcal J(b)
=
\frac12
\E\!\left[
\left|
b(Y_\tau,\tau;X_0)-R_\tau
\right|^2
\right],
\label{eq:si-training-loss}
\end{equation}
where the expectation is taken over $X_0$, $C$, $\xi$, and
$\tau\sim\pi$. We assume that the expectations appearing below are
finite.

Under this objective, the optimal drift is
\[
b^\star(y,\tau;x)
=
\E\!\left[
R_\tau\mid Y_\tau=y,X_0=x
\right].
\]
Since
\[
R_\tau
=
\frac{C-Y_\tau}{1-\tau},
\]
the optimal drift can also be written as
\begin{equation}
b^\star(y,\tau;x)
=
\frac{
\E[C\mid Y_\tau=y,X_0=x]-y
}{1-\tau}.
\label{eq:bayes-drift}
\end{equation}
Moreover, for any candidate drift $b$,
\begin{equation}
\mathcal J(b)-\mathcal J(b^\star)
=
\frac12
\E\!\left[
\left|
b(Y_\tau,\tau;X_0)
-
b^\star(Y_\tau,\tau;X_0)
\right|^2
\right].
\label{eq:excess-risk}
\end{equation}
Thus, the difference from the minimum training loss is exactly one half
of the mean-square error between the candidate drift $b$ and the optimal
drift $b^\star$.

The identities above follow directly from the stated interpolant. For
$\tau<1$,
\[
C-Y_\tau=(1-\tau)(C-x-\sigma_b\sqrt\tau\,\xi)
=(1-\tau)R_\tau.
\]
Conditional expectation gives \eqref{eq:bayes-drift}. Since $b^\star$
is the conditional expectation of $R_\tau$ given
$(Y_\tau,\tau,X_0)$, the cross term in
\[
|b-R_\tau|^2
=
|b-b^\star|^2+|b^\star-R_\tau|^2
+2(b-b^\star)^\top(b^\star-R_\tau)
\]
has zero expectation. This proves \eqref{eq:excess-risk}.

Before relating this objective to Direct SI and Cascaded SI, we connect
the approximation of the conditional mean to the error in the generated
endpoint distribution.

\begin{proposition}
\label{prop:mean-endpoint}
Let $\widehat K_h(x,\cdot)$ be any approximation of the true conditional
distribution $K_h(x,\cdot)$. Define the mean of this approximate
distribution by
\[
\widehat m_h(x)
=
\int c\,\widehat K_h(x,\dd c).
\]
Assume that
\[
\int_{\R^m}
\int_{\R^m}
|c|^2
\left[
K_h(x,\dd c)+\widehat K_h(x,\dd c)
\right]
\mu(\dd x)
<
\infty.
\]
Then
\begin{equation}
\int_{\R^m}
W_2^2\!\left(
K_h(x,\cdot),
\widehat K_h(x,\cdot)
\right)
\mu(\dd x)
\geq
\left\|
m_h-\widehat m_h
\right\|_{L^2(\mu)}^2.
\label{eq:mean-lower}
\end{equation}
\end{proposition}

\begin{proof}[Proof of Proposition~\ref{prop:mean-endpoint}]
For each $x$, let $(C,\widehat C)$ be any coupling of $K_h(x,\cdot)$
and $\widehat K_h(x,\cdot)$. Jensen's inequality gives
$\E|C-\widehat C|^2\geq|m_h(x)-\widehat m_h(x)|^2$.
Take the infimum over couplings and integrate over $x$.
\end{proof}

Proposition~\ref{prop:mean-endpoint} shows that an error in the conditional
mean necessarily produces an error in the generated conditional
distribution. In particular, if $\widehat m_h \in \PP_2$, then
\[
\int
W_2^2\!\left(
K_h(x,\cdot),
\widehat K_h(x,\cdot)
\right)
\mu(\dd x)
\geq
e_2(h)^2.
\]
Therefore, the approximation error identified in
Theorem~\ref{thm:coarse-cubic} is also a lower bound on the endpoint
distribution error of any learner whose conditional mean is restricted to
be quadratic.

This first result concerns only the endpoint distributions. It does not
yet connect the approximation error $e_2(h)$ to the stochastic-interpolant
training objective $\mathcal J$. To obtain such a relation, we consider a
particular family of conditional distributions for which both the optimal
drift and the generated endpoint distribution can be calculated exactly.

\begin{proposition}
\label{prop:gaussian-transfer}
Suppose that
\[
C\mid X_0=x
\sim
\N\!\left(m_h(x),q_hI\right),
\qquad
q_h\geq0.
\]
Assume that the interpolation-time law $\pi$ is supported on
$[0,\tau_*]$ for some $\tau_*<1$. Consider the family of
stochastic-interpolant drifts $b_{h,p}$ obtained by keeping the exact
Gaussian dependence on $y$ and $\tau$ and replacing the true conditional
mean $m_h(x)$ with a candidate function $p(x)$. There is a constant
$w_h>0$ such that
\begin{equation}
\mathcal J(b_{h,p})-\mathcal J(b_{h,m_h})
=
\frac12w_h
\left\|
p-m_h
\right\|_{L^2(\mu)}^2.
\label{eq:gaussian-transfer}
\end{equation}
When $b_{h,p}$ is integrated exactly, the generated endpoint distribution
is
\[
\N\!\left(p(x),q_hI\right).
\]
Consequently,
\begin{equation}
W_2^2\!\left(
\N(m_h(x),q_hI),
\N(p(x),q_hI)
\right)
=
|m_h(x)-p(x)|^2.
\label{eq:gaussian-endpoint-error}
\end{equation}
\end{proposition}

Thus, for this family, approximating the conditional mean controls both
the stochastic-interpolant training loss and the endpoint distribution
error. In particular, choosing $p=\Pi_2m_h$ gives
\[
\mathcal J(b_{h,\Pi_2m_h})-\mathcal J(b_{h,m_h})
=
\frac12w_he_2(h)^2,
\]
and
\[
\int
W_2^2\!\left(
\N(m_h(x),q_hI),
\N(\Pi_2m_h(x),q_hI)
\right)
\mu(\dd x)
=
e_2(h)^2.
\]

We can now connect this result to the direct and cascaded forecasts from
the previous subsection. There, the full state is observed and the
dynamics are deterministic. In that setting, the exact endpoint is
\[
C=\Phi_h(x),
\]
and hence
\[
m_h=\Phi_h.
\]
There is no conditional randomness in the endpoint. This is the case
$q_h=0$ in Proposition~\ref{prop:gaussian-transfer}; the zero covariance
simply means that the endpoint distribution is concentrated at
$\Phi_h(x)$.

Since $w_h>0$, the best quadratic function $p$ minimizes
\[
\left\|p-\Phi_h\right\|_{L^2(\nu)}^2.
\]
The optimal quadratic function is therefore precisely
\[
p
=
\Pi_2^\nu\Phi_h
=
F_h^{(2)}.
\]
Moreover, the stochastic interpolant associated with $p$ produces the
deterministic endpoint $p(x)$. The fitted quadratic maps studied in
Theorem~\ref{thm:composition} are therefore exactly the endpoint maps
produced by this stochastic-interpolant learner.

We can now state the resulting comparison between Direct SI and
Cascaded SI.

\begin{corollary}
\label{cor:si-composition}
Under the assumptions of Theorem~\ref{thm:composition}, consider the
stochastic-interpolant learner in
Proposition~\ref{prop:gaussian-transfer} with $q_h=0$ and with
$p\in\PP_2(\R^d;\R^d)$. The endpoint map produced by Direct SI is
\[
F_T^{(2)}.
\]
If the horizon is divided into $S$ intervals of length
\[
\delta=\frac{T}{S},
\]
the endpoint map produced by Cascaded SI is
\[
\left(F_\delta^{(2)}\right)^{\circ S}.
\]
For all sufficiently large $S$, Cascaded SI has smaller endpoint
approximation error than every Direct SI whose endpoint map is quadratic.
\end{corollary}

Corollary~\ref{cor:si-composition} shows, in this specified full-state
setting, that the approximation benefit proved in
Theorem~\ref{thm:composition} carries directly to the corresponding
stochastic-interpolant learners. This result isolates approximation
error. It does not yet include the information loss that can occur when
the state passed between stages of a cascade is only partially observed.

\textbf{Proof of Proposition~\ref{prop:gaussian-transfer}.}

Suppose $C\mid x\sim\N(m(x),qI)$, $q\ge0$. Define
\[
 D_\tau=\tau q+\sigma_b^2(1-\tau)^2,\quad
 K_\tau=q/D_\tau,\quad H_\tau=\sigma_b^2(1-\tau)/D_\tau,
\]
and, for a candidate $p(x)$, set
\begin{equation}
 b_p(y,\tau;x)=\frac{K_\tau-1}{1-\tau}y-K_\tau x+H_\tau p(x).
 \label{eq:gaussian-drift-proof}
\end{equation}
At $q=0$, this is $(p(x)-y)/(1-\tau)$. For $q>0$, Gaussian
conditioning proves $b_m=b^\star$, since
\[
 \E[C\mid y,x]
 =m(x)+K_\tau[y-(1-\tau)x-\tau m(x)].
\]
The same formula holds directly for $q=0$. Thus
$b_p-b_m=H_\tau(p-m)$.

Take an interpolation-time law $\pi$ independent of $(x,C,\xi)$ with
$0<w_q:=\int H_\tau^2\pi(\dd\tau)<\infty$ and finite regression losses.
One common law supported in $[0,\tau_*]$, $\tau_*<1$, suffices for all
$q\ge0$. The analytic drift formula is specified on all of $[0,1)$;
only $p$ is fitted. Equation~\eqref{eq:gaussian-transfer} follows with
$w_h=w_{q_h}$.

The affine SDE
\begin{equation}
 \dd\overline Y_\tau=b_p(\overline Y_\tau,\tau;x)\dd\tau
       +\sigma_b(1-\tau)\dd W_\tau,\qquad \overline Y_0=x,
 \label{eq:gaussian-sampler-proof}
\end{equation}
has, on every compact subinterval of $[0,1)$, a unique Gaussian solution.
Its mean and covariance are
\[
 (1-\tau)x+\tau p(x),\qquad
 [\tau^2q+\sigma_b^2\tau(1-\tau)^2]I.
\]
Substitution into the linear mean and covariance equations verifies both
expressions and their initial conditions. Taking $\tau\uparrow1$ gives
the endpoint law $\N(p(x),qI)$. Equal-covariance Gaussians are translates:
the translation coupling has cost $|p-m|^2$, and Jensen's inequality gives
the matching lower bound. This proves
Proposition~\ref{prop:gaussian-transfer}. When $q>0$, the corresponding
conditional KL is $|p(x)-m(x)|^2/(2q)$; when $q=0$, differing point masses
have infinite KL and Wasserstein distance remains the appropriate finite
metric.

\subsection{What repeated conditioning in a cascade forgets}
\label{app:forgets}

Corollary~\ref{cor:si-composition} shows that a cascade of shorter
transitions can reduce approximation error when the full state is
observed. We now return to the partially observed setting. Here cascading
introduces a second effect: at the end of each stage, the next transition
is conditioned only on the current observed state. Information carried by
the unresolved variables, and by their relation to earlier observations,
may therefore be lost.

We first study this effect using exact short-transition distributions.
There is therefore no learning error in the analysis below. Any difference
between the direct conditional law and the cascade is caused only by
repeated conditioning on the observed state.

\textbf{Exact short-transition distributions.}

We extend the notation introduced above by defining, for
$0\leq s<t\leq T$,
\[
K_{s,t}(x,\cdot)
=
\operatorname{Law}(X_t\mid X_s=x),
\qquad
\mu_t
=
\operatorname{Law}(X_t).
\]
Thus,
\[
K_{0,h}=K_h
\qquad\text{and}\qquad
\mu_0=\mu.
\]
The kernel $K_{s,t}(x,\cdot)$ is the exact conditional distribution of the
observed state at time $t$, given only the observed state $X_s=x$.

For two conditional distributions $K$ and $L$, we use the notation
\[
\mathfrak D_\mu(K,L)
=
\left[
\int
W_2^2\!\left(
K(x,\cdot),L(x,\cdot)
\right)
\mu(\dd x)
\right]^{1/2}.
\]
Thus, $\mathfrak D_\mu(K,L)$ is their average conditional Wasserstein
distance when the input state has distribution $\mu$.

Consider a partition
\[
0=t_0<t_1<\cdots<t_S=T,
\]
and write
\[
K_i
=
K_{t_i,t_{i+1}},
\qquad
i=0,\ldots,S-1.
\]
The cascade built from the exact short-transition distributions is
\begin{equation}
K^{(S)}
=
K_0K_1\cdots K_{S-1}.
\label{eq:exact-local-composition}
\end{equation}
Starting from $X_0=x$, this composition first samples from
$K_0(x,\cdot)$ and then repeatedly samples the next observed state
conditional only on the current one. We measure the difference between
this exact-kernel cascade and the direct conditional distribution
$K_{0,T}$ by
\begin{equation}
C_S
=
\mathfrak D_{\mu_0}
\left(
K_{0,T},K^{(S)}
\right).
\label{eq:memory-cost}
\end{equation}
Because all the kernels in $K^{(S)}$ are exact, $C_S$ contains no learning
or approximation error. It measures only the error caused by replacing the
true conditional dependence from $X_0$ to $X_T$ with repeated conditioning
on the current observed state.

The short-transition kernels always satisfy
\[
\mu_sK_{s,t}=\mu_t.
\]
Consequently, their composition produces the correct unconditional
distribution $\mu_{t_i}$ at every time in the partition. However, it need
not produce the correct distribution conditional on $X_0=x$. In other
words, all one-time marginals can be correct even when the dependence
between the initial and final observations is wrong~\citep{Brunick_2013}.

The following theorem makes this loss of information precise. As the
cascade becomes finer, the composition of the exact observed-state
conditional transitions approaches a deterministic evolution driven by
the conditional mean observed velocity. Consequently, the memory error
$C_S$ approaches the error made by replacing the true observed trajectory
with this deterministic evolution.

The result does not require the quadratic dynamics considered in the
previous subsections. It applies to any smooth partially observed
deterministic system satisfying the conditions below.

\begin{theorem}
\label{thm:memory-limit}
Suppose that the full state evolves according to
\[
\dot Z_t=V(Z_t)
\]
on a compact invariant set $D\subset\R^d$, and that $V$ is twice
continuously differentiable on a neighborhood of $D$. As before, let
\[
X_t=PZ_t.
\]
Define the conditional mean of the observed velocity by
\[
\overline v(t,x)
=
\E\!\left[
PV(Z_t)\mid X_t=x
\right].
\]
Assume that $\overline v$ has a bounded extension to
$[0,T]\times\R^m$ that is Lipschitz continuous in $x$ with constant $L$
and Lipschitz continuous in $t$ with constant $H$. Let
$\psi_{s,t}(x)$ denote the solution at time $t$ of
\[
\frac{\mathrm d}{\mathrm dt}\psi_{s,t}(x)
=
\overline v\!\left(t,\psi_{s,t}(x)\right),
\qquad
\psi_{s,s}(x)=x.
\]

Define
\[
M
=
\sup_{z\in D}|PV(z)|,
\qquad
A_0
=
\frac12\sup_{z\in D}|P\,DV(z)V(z)|,
\]
and
\[
B_0
=
A_0+\frac12(H+LM).
\]
For the equally spaced partition with
\[
\delta=\frac{T}{S},
\]
let
\begin{equation}
r_S
=
e^{LT}
\left(
2M\sqrt{T\delta}
+
B_0T\delta
\right).
\label{eq:rs}
\end{equation}
Then
\begin{equation}
\mathfrak D_{\mu_0}
\left(
K^{(S)},
\delta_{\psi_{0,T}(\cdot)}
\right)
\leq
r_S,
\label{eq:closure-limit}
\end{equation}
where
$\delta_{\psi_{0,T}(\cdot)}(x,\cdot)$ is the point mass at
$\psi_{0,T}(x)$. Moreover,
\begin{equation}
\left|
C_S-\Gamma_T
\right|
\leq
r_S,
\qquad
\Gamma_T^2
=
\E\!\left[
|X_T-\psi_{0,T}(X_0)|^2
\right].
\label{eq:memory-limit}
\end{equation}
The deterministic flow generated by $\overline v$ has the correct
one-time marginals:
\[
(\psi_{0,t})_\#\mu_0=\mu_t
\qquad
\text{for every }t\in[0,T].
\]

Finally, define
\[
v_0
=
\E\!\left[
\operatorname{tr}
\operatorname{Cov}
\left(
PV(Z_0)\mid X_0
\right)
\right].
\]
Then
\begin{equation}
C_S
\geq
\left(
T\sqrt{v_0}
-
A_0T^2
-
r_S
\right)_+,
\label{eq:primitive-memory-lower}
\end{equation}
where $(a)_+=\max\{a,0\}$.
\end{theorem}

The main conclusion follows from Eq.~\eqref{eq:memory-limit}. Since
$\delta=T/S$, we have $r_S\to0$ as $S\to\infty$, and therefore
\[
C_S\longrightarrow\Gamma_T.
\]
Thus, as the cascade becomes increasingly fine, the error caused by
discarding unresolved information approaches the error of replacing the
true observed trajectory by the deterministic conditional-mean flow
\[
x\longmapsto\psi_{0,T}(x).
\]

Equation~\eqref{eq:closure-limit} explains why. The composition of
increasingly short observed-state conditional transitions converges to
the point mass
\[
\delta_{\psi_{0,T}(x)}.
\]
Conditional on the initial observation $X_0=x$, the infinitely fine
exact-kernel cascade therefore becomes deterministic. It follows the
average observed velocity $\overline v(t,x)$ and loses the variability
caused by the unresolved state.

The unresolved variables persist along the true physical trajectory.
Their effects over different physical intervals are therefore correlated.
By contrast, after every stage of the cascade, the exact observed-state
kernel retains only the new observed state. It forgets which unresolved
state produced that transition and averages again over all unresolved
states compatible with the current observation.

For a smooth deterministic system, the unresolved contribution to one
short increment has size of order $\delta$. After $S$ such independently
reconditioned transitions, the accumulated variance is of order
\[
S\delta^2=T\delta,
\]
which converges to zero as $\delta\to0$. This is why the random effect of
the unresolved variables disappears and only the conditional mean velocity
remains.

Importantly, the limiting deterministic flow still has the correct
one-time distributions:
\[
(\psi_{0,t})_\#\mu_0=\mu_t.
\]
It can therefore reproduce the distribution of $X_t$ at every individual
time while failing to preserve the correct dependence between $X_0$ and
$X_T$. The error $\Gamma_T$ measures this lost dependence.

The quantity $v_0$ measures how much the observed velocity can vary among
full states with the same initial observation. If $v_0>0$, the unresolved
state affects the initial observed velocity. The lower bound in
Eq.~\eqref{eq:primitive-memory-lower} then shows that, for sufficiently
small fixed $T>0$ and sufficiently fine cascades, the memory error remains
strictly positive. Thus, arbitrarily short transitions do not in general
recover the correct conditional endpoint distribution.

\begin{proof}[Proof of Theorem~\ref{thm:memory-limit}]
The extension of $\bar v$ can be assumed to have norm at most $M$:
project any given extension onto the closed Euclidean ball of radius $M$.
This changes no conditional velocity on its support and does not increase
its spatial or temporal Lipschitz constants.

Along a physical trajectory,
$|X_{t+\delta}-X_t|\le M\delta$. Taylor's formula followed by conditioning
gives, for $\mu_{t_i}$-almost every $x$,
\begin{align}
 \int(x'-x)K_i(x,\dd x')&=\delta\bar v(t_i,x)+r_i(x),
       &|r_i(x)|&\le A_0\delta^2,\notag\\
 \int|x'-x|^2K_i(x,\dd x')&\le M^2\delta^2.
 \label{eq:increment-proof}
\end{align}
Construct a Markov chain $\widetilde X_i$ with these kernels, starting
from $\mu_0$. Its marginal is $\mu_{t_i}$ at every step, so all
almost-everywhere bounds hold at the states it visits. It has the recursion
\[
 \widetilde X_{i+1}=\widetilde X_i+\delta\bar v(t_i,\widetilde X_i)
       +r_i(\widetilde X_i)+\xi_{i+1},
\]
where $\xi_{i+1}$ is a martingale difference with conditional second
moment at most $M^2\delta^2$.

For $x_i=\psi_{0,t_i}(\widetilde X_0)$, the deterministic flow has the
one-step remainder bound
\[
 |x_{i+1}-x_i-\delta\bar v(t_i,x_i)|
       \le\tfrac12(H+LM)\delta^2.
\]
Sum the two recursions and apply discrete Gronwall to the running maximum.
The result is
\[
 \max_{i\le S}|\widetilde X_i-x_i|
 \le e^{LT}\left[\max_{i\le S}\left|\sum_{j=1}^i\xi_j\right|
                                      +B_0T\delta\right].
\]
Doob's $L^2$ inequality and orthogonality of the increments give
\[
 \left\|\max_{i\le S}\left|\sum_{j=1}^i\xi_j\right|\right\|_{L^2}
       \le2M\sqrt S\,\delta=2M\sqrt{T\delta}.
\]
Distance to a point mass is the second moment of displacement from that
point. The coupling uses the same initial state for the chain and the
deterministic flow, so it proves \eqref{eq:closure-limit}. The chain
endpoint has law $\mu_T$ for every $S$; passing to the limit proves
$(\psi_{0,T})_\#\mu_0=\mu_T$. Apply the same argument on $[0,t]$ for
each earlier time.

The reverse triangle inequality gives \eqref{eq:memory-limit}. Also
\begin{equation}
 \Gamma_T^2=\E\tr\Cov(X_T\mid X_0)
                 +\E|m_T(X_0)-\psi_{0,T}(X_0)|^2.
 \label{eq:gamma-decomposition-proof}
\end{equation}
Finally $X_T=X_0+TPV(Z_0)+\mathcal E_T$, with
$\|\mathcal E_T\|_{L^2}\le A_0T^2$. Conditional centering is an
orthogonal projection in $L^2$, so
\[
 \sqrt{\E\tr\Cov(X_T\mid X_0)}
 \ge\bigl(T\sqrt{v_0}-A_0T^2\bigr)_+.
\]
Combine with \eqref{eq:gamma-decomposition-proof} and
\eqref{eq:memory-limit} to obtain
\eqref{eq:primitive-memory-lower}.
\end{proof}

\textbf{Retaining the initial observation.}

The theorem also identifies a way to prevent this particular loss of
information. Instead of conditioning each short transition only on the
current observed state, one can retain the initial observation throughout
the cascade.

For a fixed initial observation $a$, define
\[
K_i^a(x,\cdot)
=
\operatorname{Law}
\left(
X_{t_{i+1}}
\mid
X_{t_i}=x,\,
X_0=a
\right)
\]
and
\[
\mu_i^a
=
\operatorname{Law}(X_{t_i}\mid X_0=a).
\]
The kernel $K_i^a$ therefore receives both the current observed state $x$
and the initial observation $a$.

\begin{proposition}
\label{prop:anchoring}
For any Euclidean-valued process and any finite partition, the exact
kernels defined above satisfy
\begin{equation}
\delta_a
K_0^a
K_1^a
\cdots
K_{S-1}^a
=
\operatorname{Law}(X_T\mid X_0=a)
\label{eq:anchored-exact}
\end{equation}
for $\mu_0$-almost every $a$.

Now let $\widehat K_i^a$ be approximations of $K_i^a$. Suppose that
\[
W_2\!\left(
\widehat K_i^a(x,\cdot),
\widehat K_i^a(x',\cdot)
\right)
\leq
L_i|x-x'|
\]
for every $a$, and define
\[
\varepsilon_i^2
=
\E\!\left[
W_2^2\!\left(
K_i^{X_0}(X_{t_i},\cdot),
\widehat K_i^{X_0}(X_{t_i},\cdot)
\right)
\right].
\]
If the required second moments and the quantity on the right-hand side
below are finite, then
\begin{equation}
\left[
\E_{a\sim\mu_0}
W_2^2\!\left(
\operatorname{Law}(X_T\mid X_0=a),
\delta_a
\widehat K_0^a
\widehat K_1^a
\cdots
\widehat K_{S-1}^a
\right)
\right]^{1/2}
\leq
\sum_{i=0}^{S-1}
\varepsilon_i
\prod_{j>i}L_j.
\label{eq:anchored-learned}
\end{equation}
\end{proposition}

\begin{proof}[Proof of Proposition~\ref{prop:anchoring}]
For $\mu_0$-almost every $a$, the tower property gives
\[
\mu_i^aK_i^a=\mu_{i+1}^a.
\]
Starting from $\mu_0^a=\delta_a$ proves
\eqref{eq:anchored-exact} by induction.

For the learned bound, compare two compositions at a time, keeping
the first $i$ kernels exact and all subsequent kernels learned. The
input to the replacement at step $i$ has law $\mu_i^a$. Its conditional
$W_2$ distance changes by at most
\[
e_i(a)=
\left[
\int
W_2^2\bigl(
K_i^a(x,\cdot),
\widehat K_i^a(x,\cdot)
\bigr)
\mu_i^a(\dd x)
\right]^{1/2}.
\]
Each later learned kernel enlarges this distance by at most its
Wasserstein Lipschitz constant. Sum over the replacement steps,
then apply Minkowski's inequality over $a$. Since
$\E_a[e_i(a)^2]=\varepsilon_i^2$, this gives
\eqref{eq:anchored-learned}.
\end{proof}

The first part of Proposition~\ref{prop:anchoring} shows that retaining
$X_0$ removes this endpoint memory error when the short-transition kernels
are exact. This does not mean that the resulting process reproduces the
full joint law of the physical trajectory. It guarantees only the correct
conditional law at the final time.

This anchored construction is a separate variant of the cascade. The
Cascaded SI evaluated in the main benchmark conditions each stage on the
generated current state, as described in Section~\ref{sec:methods}; the
result above shows what changes theoretically if the initial observation
is retained as an additional conditioning variable.

\textbf{Adding learning error.}

We now consider the additional error caused by learning the short
transitions. There are three conditional endpoint distributions to
distinguish:
\[
K_{0,T},
\qquad
K^{(S)}=K_0K_1\cdots K_{S-1},
\qquad
\widehat K_0\widehat K_1\cdots\widehat K_{S-1}.
\]
The first is the true transition from time $0$ to time $T$. The second is
the cascade of exact short transitions, each conditioned only on the
current observed state. The third is the learned cascade, in which each
exact transition $K_i$ is replaced by an approximation $\widehat K_i$.

The distance between the first two distributions is the memory error
$C_S$ studied above. We now bound the distance between the second and
third distributions.

For each segment, define
\[
\epsilon_i^2
=
\int
W_2^2\!\left(
K_i(x,\cdot),
\widehat K_i(x,\cdot)
\right)
\mu_{t_i}(\dd x).
\]
Thus, $\epsilon_i$ measures how well the learned transition
$\widehat K_i$ approximates the exact transition $K_i$, averaged over the
states that occur at time $t_i$.

To control how this error affects later stages, assume that
\[
W_2\!\left(
\widehat K_i(x,\cdot),
\widehat K_i(x',\cdot)
\right)
\leq
L_i|x-x'|.
\]
The constant $L_i$ measures how much the $i$th learned transition can
enlarge a difference between two input states. An error introduced at
step $i$ passes through all later learned transitions and can therefore be
enlarged by the factor
\[
\prod_{j>i}L_j.
\]
This leads to the accumulated learning-error bound
\[
B_S
=
\sum_{i=0}^{S-1}
\epsilon_i
\prod_{j>i}L_j,
\]
where the product is equal to one when $i=S-1$. In particular,
\[
\mathfrak D_{\mu_0}
\left(
K^{(S)},
\widehat K_0\widehat K_1\cdots\widehat K_{S-1}
\right)
\leq
B_S.
\]
Thus, $B_S$ bounds how much replacing the exact short transitions with
their learned approximations changes the final conditional distribution.

The total endpoint distribution error of the learned cascade is
\[
E_S
=
\mathfrak D_{\mu_0}
\left(
K_{0,T},
\widehat K_0\widehat K_1\cdots\widehat K_{S-1}
\right).
\]
Recall that
\[
C_S
=
\mathfrak D_{\mu_0}
\left(
K_{0,T},
K^{(S)}
\right)
\]
is the error caused by repeatedly conditioning only on the current
observed state, even when all short transitions are exact. The triangle
inequality and the reverse triangle inequality give
\[
|E_S-C_S|
\leq
B_S.
\]
Equivalently,
\begin{equation}
\left(C_S-B_S\right)_+
\leq
E_S
\leq
C_S+B_S.
\label{eq:band}
\end{equation}

These inequalities do not say that the total error is the sum of a memory
error and a learning error. The learned approximations may move the
endpoint distribution either closer to or farther from the true
distribution. The result says only that learning can change the
exact-kernel cascade error $C_S$ by at most $B_S$.

The bounds also do not by themselves show that Cascaded SI is better than
Direct SI. The quantity $e_2(T)$ is a lower bound on the endpoint
distribution error of any direct forecast whose conditional mean is
quadratic, whereas
\[
E_S\leq C_S+B_S
\]
is an upper bound on the error of a learned cascade. Therefore, the
condition
\[
C_S+B_S<e_2(T)
\]
would be sufficient to prove that the cascade is better than every direct
forecast with a quadratic conditional mean. Whether this condition holds
depends on the system and on the cascade depth $S$. In particular, these
bounds alone do not show that the endpoint error is minimized at a finite
depth $S>1$.

The results above therefore isolate the two mechanisms in the cascading
trade-off. Shorter transitions can reduce approximation error, but repeated
conditioning on a partial observation can discard information carried by
the unresolved state. The bounds keep these effects separate; they do not
determine where their balance is optimal. We next turn to the
rotation--OU model, where the endpoint error of a specified learner can be
computed explicitly and a finite optimal cascade depth can be proved.

\textbf{Proof of the learned-kernel bound.}

A kernel satisfying the $W_2$ Lipschitz bound with constant $L_i$ sends
input laws at distance $d$ to output laws at distance at most $L_id$:
couple the inputs, couple the conditional outputs, and take infima.
Its compositions have the product Lipschitz bound.

Compare the exact and learned products through hybrid products with an
exact prefix and a learned suffix. At replacement index $i$, the common
prefix, averaged over $X_0$, has law $\mu_{t_i}$. Couple that prefix
identically, use an optimal coupling of $K_i(x,\cdot)$ and
$\widehat K_i(x,\cdot)$, and propagate through the learned suffix.
The integrated conditional cost of this replacement is at most
$\epsilon_i\prod_{j>i}L_j$. Triangle inequalities over the hybrids give
\[
 \D_{\mu_0}(K^{(S)},\widehat K_0\cdots\widehat K_{S-1})\le B_S.
\]
The triangle and reverse triangle inequalities with $K_{0,T}$ prove
\eqref{eq:band}. In particular, when Theorem~\ref{thm:memory-limit}
applies, $|E_S-\Gamma_T|\le B_S+r_S$. Neither $B_S$ nor $C_S$ is
presumed monotone. A mean-only fitting error cannot replace $\epsilon_i$,
which measures an entire conditional distribution.

\subsection{Direct and cascaded forecasts in the rotation--OU model:
an explicit trade-off}
\label{sec:rotation-ou-tradeoff}

The previous results identify two opposing effects of cascading. Shorter
transitions can be easier to approximate, but repeated conditioning on the
observed state can discard information carried by unresolved variables.
Those general results do not yet show whether these effects can produce an
optimal intermediate cascade depth.

We now exhibit this behavior in a model for which the generated endpoint
distribution can be calculated explicitly. This is the model introduced
in Section~\ref{sec:rotation-model}. We proceed in three steps. First, we
construct a family of forecasting problems with the same target endpoint
distribution but different amounts of hidden memory. Second, we compute
the endpoint distribution generated by Direct SI and Cascaded SI for a
specified quadratic learner. Finally, we compare direct forecasting, fine
cascades, and an intermediate cascade.

\textbf{A fixed endpoint task with tunable hidden memory.}

Let
\[
c\sim\operatorname{Unif}[-1,1],
\qquad
Y_0\sim\N(0,s_0^2I_2),
\]
independently, and let the hidden state $H_t\in\R^2$ and observed state
$Y_t\in\R^2$ satisfy
\begin{align}
\dd H_t
&=
-\kappa H_t\dd t+\sqrt{2\kappa}\dd B_t,
&
H_0&\sim\N(0,I_2),
\notag\\
\dd Y_t
&=
\omega cJY_t\dd t
+\beta_\rho R(\omega ct)H_t\dd t
+\sigma_\rho\dd W_t,
\label{eq:rotation-ou-model}
\end{align}
where
\[
J=
\begin{pmatrix}
0&-1\\
1&0
\end{pmatrix},
\qquad
R(\theta)=e^{\theta J},
\]
and all initial variables and Brownian motions are independent. The
conditioning variable $c$ is retained throughout the forecast. By
contrast, $H_t$ is unresolved: every stage is conditioned on the current
value of $Y_t$ and on $c$, but not on $H_t$.

For fixed $T,q,\kappa,s_0>0$, define
\begin{equation}
V_H(t)
=
2\left[
\frac{t}{\kappa}
-
\frac{1-e^{-\kappa t}}{\kappa^2}
\right],
\qquad
\beta_\rho^2
=
\frac{\rho q}{V_H(T)},
\qquad
\sigma_\rho^2
=
\frac{(1-\rho)q}{T},
\label{eq:rotation-ou-parameters}
\end{equation}
where $\rho\in[0,1]$. Thus, the integrated OU process contributes
$\rho qI_2$ to the terminal covariance, while the white noise contributes
$(1-\rho)qI_2$.

To remove the deterministic rotation, introduce
\[
Z_t=R(-\omega ct)Y_t.
\]
The dynamics of $Z_t$ contain no rotational drift, and
\[
Z_T\mid(Y_0=y,c)\sim\N(y,qI_2).
\]
Transforming back to the original coordinates gives
\begin{equation}
Y_T\mid(Y_0=y,c)
\sim
\N\!\left(R(\omega cT)y,qI_2\right)
\label{eq:rotation-ou-endpoint}
\end{equation}
for every $\rho\in[0,1]$. Consequently, changing $\rho$ changes the amount
of information carried through time by the hidden state without changing
the target conditional endpoint distribution.

\textbf{Exact short transitions.}

Although the endpoint forecasting task is independent of $\rho$, the
short conditional transitions are not. For $0\leq s\leq t$, set
\begin{align}
I_\kappa(s,t)
&=
\frac{2s}{\kappa}
-
\frac{(1-e^{-\kappa s})(1+e^{-\kappa(t-s)})}{\kappa^2},
\notag\\
v_s
&=
s_0^2+\beta_\rho^2V_H(s)+\sigma_\rho^2s,
\notag\\
b_{s,t}
&=
s_0^2+\beta_\rho^2I_\kappa(s,t)+\sigma_\rho^2s,
\qquad
r_{s,t}
=
\frac{b_{s,t}}{v_s}.
\label{eq:rotation-ou-covariances}
\end{align}
Consider the equal partition $t_i=i\delta$, where $\delta=T/S$, and write
\[
r_i=r_{t_i,t_{i+1}},
\qquad
q_i=v_{t_{i+1}}-r_i^2v_{t_i}.
\]
Gaussian conditioning gives
\begin{equation}
Y_{t_{i+1}}\mid(Y_{t_i}=y,c)
\sim
\N\!\left(
r_iR(\omega c\delta)y,
q_iI_2
\right).
\label{eq:rotation-ou-local-kernel}
\end{equation}

When $\rho>0$, composing these exact short transitions does not recover
Eq.~\eqref{eq:rotation-ou-endpoint}. At every reconditioning step, the
cascade retains $Y_{t_i}$ and $c$ but forgets the unresolved state
$H_{t_i}$. A cascade can therefore incur endpoint error even before its
short transitions are approximated.

\textbf{A fixed quadratic learner for Direct SI and Cascaded SI.}

We next introduce approximation error while keeping the comparison fully
specified. We take $0\leq\rho<1$, so that every $q_i$ is positive, and use
the Gaussian stochastic-interpolant family of
Proposition~\ref{prop:gaussian-transfer}. The exact scalar coefficients
$r_i$ and $q_i$ are supplied to the learner. The only fitted quantity is a
shared function $p(c,y)$ of total degree at most two, used to approximate
the rotation $R(\omega c\delta)y$ in the conditional mean.

Set
\[
\nu=\omega\delta,
\qquad
j_0(\nu)=\frac{\sin\nu}{\nu},
\qquad
j_1(\nu)=\frac{\sin\nu-\nu\cos\nu}{\nu^2},
\]
with $j_0(0)=1$ and $j_1(0)=0$. Minimizing the expected
stochastic-interpolant training objective over the quadratic function
class gives, at every stage,
\begin{equation}
p_\nu(c,y)
=
M_\nu(c)y,
\qquad
M_\nu(c)
=
j_0(\nu)I_2+3j_1(\nu)cJ.
\label{eq:rotation-ou-quadratic-fit}
\end{equation}
Centered Gaussian symmetry eliminates terms that are nonlinear in $y$.
The fitted map in Eq.~\eqref{eq:rotation-ou-quadratic-fit} is therefore
the $L^2$ projection of the rotation onto functions that are linear in
$y$ and affine in $c$.

The role of the cascade is now explicit. Direct SI uses
$\nu=\omega T$ and must approximate the full rotation in one transition.
Cascaded SI uses $\nu=\omega T/S$ and composes $S$ simpler fitted
rotations.

\textbf{Exact endpoint distribution error.}

Define
\[
d_\nu(c)
=
j_0(\nu)^2+9j_1(\nu)^2c^2,
\qquad
A_S
=
\prod_{i=0}^{S-1}r_i,
\]
and let
\begin{equation}
\widehat q_0(c)=0,
\qquad
\widehat q_{i+1}(c)
=
r_i^2d_\nu(c)\widehat q_i(c)+q_i.
\label{eq:rotation-ou-learned-covariance}
\end{equation}
After $S$ learned transitions, the generated conditional distribution is
Gaussian, with mean $A_SM_\nu(c)^Sy$ and covariance
$\widehat q_S(c)I_2$.

We measure its discrepancy from the target through the average squared
conditional Wasserstein error
\begin{equation}
\mathcal E_S(\rho)
=
\frac12
\int_{-1}^1
\left[
s_0^2
\left\|
R(\omega Tc)-A_SM_\nu(c)^S
\right\|_F^2
+
2\left(
\sqrt q-\sqrt{\widehat q_S(c)}
\right)^2
\right]
\dd c.
\label{eq:rotation-ou-endpoint-error}
\end{equation}
The first term measures the error in the conditional mean, while the
second measures the error in the conditional covariance.
Equation~\eqref{eq:rotation-ou-endpoint-error} gives the complete endpoint
error of the specified learner. It is not obtained by adding a separately
defined approximation error and memory error.

\textbf{Direct forecasting and the infinitely deep cascade.}

Before comparing all values of $S$, it is useful to identify the two
extremes. For Direct SI, $S=1$. Writing $\Omega=\omega T$, we obtain
\begin{equation}
\mathcal E_1(\rho)
=
2s_0^2
\left[
1-j_0(\Omega)^2-3j_1(\Omega)^2
\right]
>
0
\label{eq:rotation-ou-direct-error}
\end{equation}
whenever $\omega\neq0$. The generated covariance is exact, but one
quadratic map cannot reproduce the full dependence of
$R(\omega Tc)y$ on $c$. Since the endpoint task is independent of
$\rho$, this direct approximation error is also independent of $\rho$.

At the opposite extreme, let $S\to\infty$ and define
\begin{align}
a_\rho(t)
&=
\frac{\beta_\rho^2(1-e^{-\kappa t})}{\kappa v_t},
&
A_\infty
&=
\exp\!\left(\int_0^T a_\rho(t)\dd t\right),
\notag\\
q_\infty
&=
v_T-s_0^2A_\infty^2,
&
\mathcal C_{\rho,\infty}
&=
2s_0^2(A_\infty-1)^2
+2\left(\sqrt q-\sqrt{q_\infty}\right)^2.
\label{eq:rotation-ou-fine-limit}
\end{align}
If $\rho=0$, there is no persistent hidden input and arbitrarily short
transitions recover the endpoint distribution. If $0<\rho<1$, however,
$\mathcal C_{\rho,\infty}>0$: an increasingly deep cascade repeatedly
forgets information carried by the hidden OU state.

The comparison therefore contains two competing effects. Direct SI has a
fixed approximation error from fitting the full rotation in one
transition. Increasing the cascade depth reduces this error by replacing
that transition with simpler rotations. But in the presence of persistent
hidden memory, a sufficiently deep cascade approaches a nonzero error
caused by repeated reconditioning on the observed state. The following
theorem shows that, when the hidden-memory contribution is sufficiently
weak but nonzero, neither extreme is optimal.

\begin{theorem}[An intermediate cascade can be optimal]
\label{thm:rotation-ou-interior}
Fix $T,\kappa,q,s_0>0$ and $\omega\neq0$. For the centered rotation--OU
model and the quadratic stochastic-interpolant learner defined above,
Eq.~\eqref{eq:rotation-ou-endpoint-error} has the following properties.
\begin{enumerate}
\item \emph{No persistent hidden memory.} If $\rho=0$, then
\[
\mathcal E_S(0)=O(S^{-2})
\qquad\text{as }S\to\infty.
\]
Thus, in the absence of the persistent hidden OU contribution,
progressively finer cascades recover the target endpoint distribution.

\item \emph{Persistent hidden memory.} For every fixed $0<\rho<1$, there
is a constant $\Lambda_\rho>0$ such that
\begin{equation}
\mathcal E_S(\rho)
=
\mathcal C_{\rho,\infty}
-
\Lambda_\rho\frac{T}{S}
+
O(S^{-2}).
\label{eq:rotation-ou-large-S}
\end{equation}
Thus, once the cascade becomes sufficiently deep, increasing $S$ moves
the endpoint error upward toward the nonzero limit
$\mathcal C_{\rho,\infty}$.

\item \emph{Intermediate optimum.} There exists $\rho_*>0$ such that, for
every $0<\rho<\rho_*$, the minimum over positive integers $S$ is attained
at a finite $S^\star\geq2$, and
\begin{equation}
\mathcal E_{S^\star}(\rho)
<
\min\left\{
\mathcal E_1(\rho),
\mathcal C_{\rho,\infty}
\right\}.
\label{eq:rotation-ou-strict-interior}
\end{equation}
Hence the best finite cascade is strictly better than both Direct SI and
the infinitely deep cascade limit.
\end{enumerate}
\end{theorem}

The theorem makes the cascading trade-off explicit. Dividing the horizon
into shorter transitions eventually reduces the error from approximating
the rotation, although not necessarily at every step from $S=1$: at
$\rho=0$, $s_0^2=q=T=1$, and $\omega=2\pi$, for instance, the error
rises from $1.85$ at $S=1$ to $4.27$ at $S=4$ before decaying.
Once the transitions become too short, however, the benefit of fitting
simpler rotations is outweighed by the information lost through repeated
conditioning without $H_t$. The minimum can therefore occur at an
intermediate cascade depth.

\textbf{Without rotation, cascading only loses memory.}
Theorem~\ref{thm:rotation-ou-interior} assumes $\omega\neq0$. If $\omega=0$, then $\nu=0$, $M_0(c)=I_2$, and $d_0(c)=1$, so every transition mean is fitted exactly. For Direct SI, $r_0=1$ and $\widehat q_1=q_0=q$, so $\mathcal E_1(\rho)=0$: the direct forecast is exact. For $i\geq1$, Eq.~\eqref{eq:rotation-ou-covariances} gives
\[
r_i-1
=
\frac{\beta_\rho^2}{\kappa^2v_{t_i}}
\bigl(1-e^{-\kappa t_i}\bigr)\bigl(1-e^{-\kappa\delta}\bigr),
\]
which is positive when $\rho>0$. Given $Y_{t_i}$ alone, a transition cannot tell which part of the state was produced by the persistent hidden process and which part is the initial state, and it extrapolates a fraction of the whole state. Hence $A_S>1$ for every $S\geq2$, and Eq.~\eqref{eq:rotation-ou-endpoint-error} gives
\[
\mathcal E_S(\rho)\geq2s_0^2(A_S-1)^2>0=\mathcal E_1(\rho).
\]
When the conditional mean lies in the approximation class, cascading thus has nothing to gain, and every cascade is worse than the direct forecast. In numerical evaluations of Eq.~\eqref{eq:rotation-ou-endpoint-error}, $\mathcal E_S(\rho)$ moreover increases with $S$, as does the error of the passive-scalar forecasts (Figure~\ref{fig:depth_sweep}), whose conditional mean is affine in the observation (Proposition~\ref{prop:linear-control}).

\textbf{How the optimal cascade depth depends on memory.}

The previous theorem establishes the existence of an intermediate
optimum. We now determine how its location changes as the hidden-memory
contribution becomes small. Define
\begin{align}
u(t)
&=
\frac{q(1-e^{-\kappa t})}
{\kappa V_H(T)(s_0^2+qt/T)},
&
U
&=
\int_0^T u(t)\dd t,
&
D
&=
\frac12\left[u(T)+\kappa U\right],
\notag\\
\lambda_2
&=
4s_0^2\left(1+\frac{s_0^2}{q}\right)UD,
&
B_{\mathrm{fit}}
&=
\frac{\omega^4T^2(4s_0^2+q)}{90}.
\label{eq:rotation-ou-scaling-constants}
\end{align}
Both $\lambda_2$ and $B_{\mathrm{fit}}$ are strictly positive.

The origin of the scaling can be seen by writing $\delta=T/S$. For small
$\rho$ and small $\delta$, the endpoint error satisfies
\begin{equation}
\mathcal E_S(\rho)
=
\mathcal C_{\rho,\infty}
-\lambda_2\rho^2\delta
+B_{\mathrm{fit}}\delta^2
+
O\!\left(
\rho^3\delta+\rho\delta^2+\delta^3
\right).
\label{eq:rotation-ou-joint-expansion}
\end{equation}
The positive term $B_{\mathrm{fit}}\delta^2$ is the remaining error from
approximating the short rotation. The negative term
$-\lambda_2\rho^2\delta$ describes the improvement obtained by keeping a
finite transition length instead of taking the infinitely deep
reconditioning limit. Balancing these two leading terms gives
$\delta^\star\asymp\rho^2$, or equivalently
$S^\star\asymp\rho^{-2}$. The following theorem makes this statement
precise.

\begin{theorem}[The optimal cascade depth grows as memory weakens]
\label{thm:rotation-ou-scaling}
Under the assumptions of
Theorem~\ref{thm:rotation-ou-interior}, every global minimizer over
positive integers satisfies
\begin{equation}
\rho^2S^\star(\rho)
\longrightarrow
\frac{2B_{\mathrm{fit}}T}{\lambda_2}
\qquad
\text{as }\rho\downarrow0.
\label{eq:rotation-ou-optimal-segments}
\end{equation}
Moreover,
\begin{equation}
\min_{S\geq1}\mathcal E_S(\rho)
=
\mathcal C_{\rho,\infty}
-
\frac{\lambda_2^2}{4B_{\mathrm{fit}}}\rho^4
+
o(\rho^4).
\label{eq:rotation-ou-optimal-error}
\end{equation}
\end{theorem}

Thus, weaker hidden memory permits a deeper useful cascade:
\[
S^\star(\rho)\asymp\rho^{-2}.
\]

The rotation--OU model therefore exhibits the same qualitative trade-off
that motivates the empirical depth study in the main paper: the direct
transition can have substantial approximation error, a moderate cascade
can reduce it, and a sufficiently deep cascade can lose part of that
improvement through repeated conditioning on an incomplete state. The
theorem is a statement about this specified Gaussian process and
quadratic SI learner. It does not claim that hidden OU-like memory is the
mechanism responsible for the depth curves of the PDE benchmarks, nor
does it predict the value of $S^\star$ selected in those experiments.

The conditioning calculations, the uniform expansion in
Eq.~\eqref{eq:rotation-ou-joint-expansion}, and the proofs of
Theorems~\ref{thm:rotation-ou-interior}--\ref{thm:rotation-ou-scaling} are given in Section~\ref{app:subsec:proofs_OU}.

\subsection{Learning anchored transitions in the rotation--OU model}
\label{app:anchored-ou-theory}
Conditioning every transition on $Y_0$ removes the memory error when the
transition kernels are exact (Proposition~\ref{prop:anchoring}). However, it also
changes the conditional mean that must be learned. To examine this change
in the same model, let
\[
m_i^{\mathrm{anc}}(c,y,a)
=\E[Y_{t_{i+1}}\mid Y_{t_i}=y,Y_0=a,c].
\]
At each stage, approximate this mean by its population least-squares
projection onto the vector-valued polynomials of total degree at most two
in $(c,y,a)$, under the true joint law of $(c,Y_{t_i},Y_0)$. As above, supply
the exact conditional covariance and sample the resulting Gaussian kernel
exactly. Denote the endpoint law by $\widehat K_S^{\mathrm{anc}}(a,c)$ and set
\[
\mathcal E_S^{\mathrm{anc}}(\rho,\omega)
=\E_{Y_0,c}W_2^2\!\left(
K_T^{\mathrm{OU}}(Y_0,c),\widehat K_S^{\mathrm{anc}}(Y_0,c)
\right).
\]
The polynomial degree is unchanged, but the learner now receives the
initial observation as an additional input.

\begin{theorem}[Anchoring with quadratic transition means]
\label{thm:anchored-ou}
Fix $T,\kappa,q,s_0>0$ and $0<\rho<1$ in~\eqref{eq:ou-dynamics}. Define
\begin{align*}
w_t&=2\beta_\rho^2\left[
\frac{t}{\kappa}-\frac{1-e^{-\kappa t}}{\kappa^2}
\right]+\sigma_\rho^2t,\\
k_t&=\frac{\beta_\rho^2(1-e^{-\kappa t})}{\kappa},
\qquad\lambda_t=\frac{k_t}{w_t}\quad(t>0),
\end{align*}
with $\lambda_0=\beta_\rho^2/\sigma_\rho^2$. Thus $w_T=q$. Let
$z_{\mathrm{un}}$ solve
\[
\dot z_{\mathrm{un}}(t)
=\frac{2k_t}{s_0^2+w_t}z_{\mathrm{un}}(t)+\sigma_\rho^2,
\qquad z_{\mathrm{un}}(0)=0;
\]
this is the conditional variance per coordinate of the unanchored
infinite-depth forecast.
\begin{enumerate}[itemsep=2pt,topsep=2pt]
\item Composing the exact anchored kernels recovers
$K_T^{\mathrm{OU}}(a,c)$ at every depth. If $\omega=0$, their means are
affine, so the quadratic anchored forecast is also exact at every depth.
\item For every fixed $\omega\ne0$, the quadratic anchored forecasts
converge in averaged conditional $W_2$ as $S\to\infty$ to the Gaussian law
$\N(F_T(c)a,z_{\mathrm{anc}}(T)I_2)$, whose variance satisfies
\[
z_{\mathrm{un}}(T)<z_{\mathrm{anc}}(T)<q.
\]
Consequently, the limiting error satisfies
\begin{equation}
\mathcal C_{\rho,\infty}^{\mathrm{anc}}(\omega)
:=\lim_{S\to\infty}\mathcal E_S^{\mathrm{anc}}(\rho,\omega)
\geq2\bigl(\sqrt q-\sqrt{z_{\mathrm{anc}}(T)}\bigr)^2>0.
\label{eq:anchored-ou-positive-limit}
\end{equation}
\item Let $D_t$ quantify the accumulation and propagation of the
leading-order approximation error in the conditional mean, through
\[
\dot D_t=\lambda_t(D_t+t^2),\qquad D_0=0.
\]
With the other parameters fixed, as $\omega\to0$,
\begin{equation}
\mathcal C_{\rho,\infty}^{\mathrm{anc}}(\omega)
=\frac{2s_0^2D_T^2}{45}\,\omega^4+O(\omega^6).
\label{eq:anchored-ou-small-rotation}
\end{equation}
In particular, for all sufficiently small $|\omega|$, this error is
strictly smaller than the unanchored limiting error
$\mathcal C_{\rho,\infty}$.
\end{enumerate}
\end{theorem}

The proof is given in Section~\ref{app:subsec:proofs_OU}.

Thus retaining the initial observation removes the error of exact
reconditioning, but it introduces a dependence on $R(\omega ct)Y_0$
into the mean of each short transition. For nonzero rotation, the
quadratic class cannot represent this dependence even in the
infinitesimal conditional mean. The remaining infinite-depth error is
therefore an approximation error. Anchoring reduces the limiting variance error for every nonzero rotation and the total limiting error for sufficiently small rotation.

\subsection{Learning the anchor dependence in the rotation--OU model}
\label{app:ou-anchoring-experiment}
The previous analysis shows that retaining the initial observation
removes memory error for exact transitions, but can increase the
approximation difficulty for learned transitions. We test this
representation effect by training three Wiener stochastic interpolants on the rotation--OU model: (A) unanchored SI, conditioned on the current state $Y_{t_i}$; (B) raw-anchored SI, additionally conditioned on $Y_0$; and (C) rotated-anchor SI, supplied with $R(\omega c t_{i+1})Y_0$ instead of $Y_0$. All three receive the observed parameter $c$ and relevant times. The anchor in (C) is an invertible, known transformation of the
anchor in (B), so the two contain identical information; the network in (C) need not learn this rotation. To match input dimensions, model (A) receives a zero-filled anchor channel.

We use $T=1$, $S=8$, $\kappa=1$, $\beta=0.5$, $\sigma=0.1$,
$Y_0\sim\mathcal N(0,0.25^2I_2)$, $c\sim\mathrm{Unif}[-1,1]$, and
$\omega\in\{0,12\}$. Each model is a four-hidden-layer MLP of width
128, trained on 16,000 transition pairs for 30,000 updates. For each of three paired training seeds, the SI training-noise scale is selected from $\{0.5,1,2\}$ by validation energy distance; checkpoints are selected separately. The sampling-step doubling check changes mean energy distance by less than $0.001$.

For each of 100 held-out pairs $(c^{(i)},y_0^{(i)})$, we generate
4,000 forecast endpoints and estimate their mean $\widehat m_i$ and
covariance $\widehat\Sigma_i$. As the exact endpoint law is Gaussian,
we report the conditional Gaussian moment-matched distance
\begin{equation}
\label{eq:ou-anchoring-w2}
\frac1{100}\sum_{i=1}^{100}
W_2^2\!\left(
\mathcal N(\widehat m_i,\widehat\Sigma_i),\,
\mathcal N\!\left(R(\omega c^{(i)}T)y_0^{(i)},\sigma_T^2I_2\right)
\right).
\end{equation}
Distances are evaluated per conditioning pair, in original physical
coordinates, before averaging. This metric compares fitted Gaussian
moments, not necessarily the full generated endpoint laws.

\begin{table}[h!]
\caption{\textbf{Rotation--OU anchoring ablation.}
Conditional Gaussian moment-matched $W_2^2$ averaged across 100 held-out conditioning pairs. Mean $\pm$ standard error over
three training seeds.}
\label{tab:ou-anchoring}
\centering
\small
\begin{tabular}{@{}cccc@{}}
\toprule
$\omega$ & Unanchored & Raw anchor & Rotated anchor \\
\midrule
$0$  & $0.1504\pm0.0011$ & $0.0050\pm0.0005$ & $\mathbf{0.0050\pm0.0005}$ \\
$12$ & $0.0945\pm0.0030$ & $0.0934\pm0.0049$ & $\mathbf{0.0116\pm0.0008}$ \\
\bottomrule
\end{tabular}
\end{table}

At $\omega=0$, raw anchoring strongly improves the forecast
(Table~\ref{tab:ou-anchoring}). At $\omega=12$, the selected raw-anchored
model performs almost identically to the unanchored model, whereas
pre-rotating the same information substantially reduces the error.

\subsection{Proofs for the rotation--OU model}
\label{app:subsec:proofs_OU}
\textbf{Exact OU conditioning.}

Conditionally on $c$, rotate the process in
\eqref{eq:rotation-ou-model} by
$Z_t=R(-\omega ct)Y_t$. The visible rotation cancels and
\begin{equation}
 Z_t=Y_0+\beta_\rho\int_0^tH_u\dd u+\sigma_\rho\widetilde W_t.
 \label{eq:co-rotating-proof}
\end{equation}
The rotated Brownian motion has the standard law and is independent of
$H,Y_0$ conditionally on $c$. This entire conditional law is independent
of $c$ and $\omega$.

Since $H$ is stationary with covariance
$\E[H_sH_t^\top]=e^{-\kappa|t-s|}I_2$, for $0\le s\le t$ define
\begin{align}
 I_\kappa(s,t)&=\int_0^s\int_0^t e^{-\kappa|r-u|}\dd u\dd r
   =\frac{2s}{\kappa}
    -\frac{(1-e^{-\kappa s})(1+e^{-\kappa(t-s)})}{\kappa^2},
    \notag\\
 v_s&=s_0^2+\beta_\rho^2V_H(s)+\sigma_\rho^2s,\notag\\
 b_{s,t}&=s_0^2+\beta_\rho^2 I_\kappa(s,t)+\sigma_\rho^2s,
 \qquad r_{s,t}=b_{s,t}/v_s.
 \label{eq:rotation-ou-covariances-aux}
\end{align}
Then $\Cov(Z_s)=v_sI_2$ and $\Cov(Z_s,Z_t)=b_{s,t}I_2$.
Gaussian conditioning gives
\begin{equation}
 Z_t\mid Z_s=z\sim
 \N\bigl(r_{s,t}z,[v_t-r_{s,t}^2v_s]I_2\bigr).
 \label{eq:ou-kernel-proof}
\end{equation}
At $s=0,t=T$, $r_{0,T}=1$ and
$v_T-s_0^2=\beta_\rho^2V_H(T)+\sigma_\rho^2T=q$.
This proves the endpoint law \eqref{eq:rotation-ou-endpoint}.

For a partition write
\[
 r_i=r_{t_i,t_{i+1}},\qquad
 q_i=v_{t_{i+1}}-r_i^2v_{t_i},\qquad
 A_S=\prod_{i=0}^{S-1}r_i,\qquad
 \widetilde q_S=v_T-s_0^2A_S^2.
\]
Composition of \eqref{eq:ou-kernel-proof} gives
\begin{equation}
 \widetilde Z_T\mid Y_0=y\sim
 \N(A_Sy,\widetilde q_SI_2).
 \label{eq:ou-product-law-proof}
\end{equation}
To verify the covariance, start $w_0=0$ and use
$w_{i+1}=r_i^2w_i+q_i$. Induction yields
$w_i=v_{t_i}-s_0^2\prod_{j<i}r_j^2$, so
$\widetilde q_S\ge0$.

The average squared conditional Wasserstein cost is
\begin{equation}
 \mathcal C_{\rho,S}
 =2s_0^2(A_S-1)^2+2(\sqrt q-\sqrt{\widetilde q_S})^2.
 \label{eq:ou-cost-proof}
\end{equation}
This is a squared distance, unlike the root distance $C_S$ in
\eqref{eq:memory-cost}. To derive it, the conditional mean discrepancy
in the co-rotating frame is $(A_S-1)y$, whose squared norm has
expectation $2s_0^2(A_S-1)^2$. Isotropic Gaussian covariances contribute
$2(\sqrt q-\sqrt{\widetilde q_S})^2$. Terminal rotation is an isometry,
so the result is independent of $\omega$.

For strict monotonicity in $\rho$, compute
\[
 D_{s,t}:=I_\kappa(s,t)-V_H(s)
       =\frac{(1-e^{-\kappa s})(1-e^{-\kappa(t-s)})}{\kappa^2}.
\]
Then
\begin{equation}
 r_{s,t}=1+
 \frac{\rho qD_{s,t}/V_H(T)}
 {s_0^2+qs/T+\rho q[V_H(s)/V_H(T)-s/T]}.
 \label{eq:r-monotone-proof}
\end{equation}
Its derivative in $\rho$ is
$(qD_{s,t}/V_H(T))(s_0^2+qs/T)$ divided by the squared denominator,
strictly positive for $0<s<t$. The first partition interval has
$r_0=1$, and each later one has $r_i>1$ at $\rho>0$.
Hence $A_S$ increases strictly from one and $\widetilde q_S$ decreases
from $q$ for $S\ge2$. Both terms of \eqref{eq:ou-cost-proof} increase
strictly as a function of $\rho$. At $S=1$ or $\rho=0$ the cost is zero.
Continuity covers the endpoint $\rho=1$. This proves monotonicity in the
memory parameter for a fixed partition; it does not assert monotonicity
under every change of cascade depth.

\textbf{Optimal quadratic fit and endpoint law.}

Keep the centered initial law, take $0\le\rho<1$, and use equal intervals
of length $\delta=T/S$. Set $\Omega=\omega T$ and
$\nu=\omega\delta$.
The marginal $Y_{t_i}\sim\N(0,v_{t_i}I_2)$ is independent of $c$, and
\[
 Y_{t_{i+1}}\mid(Y_{t_i}=y,c)\sim
                 \N(r_iR(\nu c)y,q_iI_2).
\]
Here $q_i>0$ because the physical white-noise increment has positive
variance for $\rho<1$. The parameter $c$ is retained as a conditioning
input; it is not interpolated.

Supply the known $r_i,q_i$ and fit a shared polynomial $p(c,y)$ of total
degree at most two, with candidate local mean $r_ip(c,y)$.
The Gaussian SI reduction multiplies each stage's mean-squared loss by a
positive scalar depending on $q_i$. Centered Gaussian symmetry in $y$
shows that only terms linear in $y$ can correlate with the target. Their
coefficients can have degree at most one in $c$. Therefore the unique
function that minimizes the expected squared error fitted at every stage is
\begin{equation}
 p_\nu(c,y)=M_\nu(c)y,\qquad
 M_\nu(c)=j_0(\nu)I_2+3j_1(\nu)cJ.
 \label{eq:Mnu-proof}
\end{equation}
Indeed, $\E_c\cos(\nu c)=j_0(\nu)$, while orthogonal projection of
$\sin(\nu c)$ onto the span of $c$ has coefficient
$3\E_c[c\sin(\nu c)]=3j_1(\nu)$. Every positive weighted average of the
stage losses has this same minimizer: after restricting to the
linear-in-$y$ part, the only additional stage factors are scalar
$r_i^2v_{t_i}$. Thus one shared fit suffices.

Let $d_\nu(c)=j_0(\nu)^2+9j_1(\nu)^2c^2$, so
$M_\nu M_\nu^\top=d_\nu I_2$. The exact learned Gaussian kernels
compose to a Gaussian endpoint conditional on $(y,c)$, with mean
$A_SM_\nu(c)^Sy$ and scalar covariance $\widehat q_S(c)I_2$, where
\begin{equation}
 \widehat q_0(c)=0,\qquad
 \widehat q_{i+1}(c)=r_i^2d_\nu(c)\widehat q_i(c)+q_i.
 \label{eq:qhat-proof}
\end{equation}
Define
\begin{align*}
d_\rho&=\frac12\left[a_\rho(T)+
\int_0^T(\kappa a_\rho(t)+a_\rho(t)^2)\dd t\right],\\
\mathcal C(A)&=2s_0^2(A-1)^2+
2\left(\sqrt q-\sqrt{v_T-s_0^2A^2}\right)^2.
\end{align*}
Then $\mathcal C(A_\infty)=\mathcal C_{\rho,\infty}$.

\begin{proof}[Proof of Theorem~\ref{thm:rotation-ou-interior}]
At $\rho=0$, $r_i=1$ and $q_i=q/S$. Uniformly in $c\in[-1,1]$,
\[
 M_{\omega\delta}(c)-R(\omega\delta c)=O(\delta^2),\qquad
 \|M_{\omega\delta}(c)\|_{\rm op}\le1+C\delta^2.
\]
A telescoping product gives
$M_{\omega\delta}^S-R(\Omega c)=O(\delta)$. Also
$d_{\omega\delta}=1+O(\delta^2)$, so
\[
 \widehat q_S(c)=\frac qS\sum_{j=0}^{S-1}d_{\omega\delta}(c)^j
       =q+O(\delta).
\]
Since $q>0$, both squared errors in
\eqref{eq:rotation-ou-endpoint-error} are $O(\delta^2)$. Together with
\eqref{eq:rotation-ou-direct-error}, this proves part (i).

Next fix $0<\rho<1$. Equation~\eqref{eq:r-monotone-proof} is equivalently
\[
 r_i=1+a_\rho(t_i)\frac{1-e^{-\kappa\delta}}{\kappa}.
\]
All following Taylor estimates are uniform in $t_i\in[0,T]$ and
$c\in[-1,1]$, with constants for the fixed physical parameters.
Expanding the logarithm gives
\[
 \log r_i=a_\rho(t_i)\delta
   -\tfrac12[\kappa a_\rho(t_i)+a_\rho(t_i)^2]\delta^2+O(\delta^3).
\]
The left Riemann sum for the smooth function $a_\rho$, with
$a_\rho(0)=0$, yields
\begin{equation}
 \log A_S=\log A_\infty-d_\rho\delta+O(\delta^2),\qquad
 A_S=A_\infty(1-d_\rho\delta)+O(\delta^2).
 \label{eq:AS-expansion-proof}
\end{equation}
Here $d_\rho>0$ and $A_\infty>1$.

To justify differentiating the covariance cost, note that
$v_t'=2a_\rho(t)v_t+\sigma_\rho^2$. An integrating factor gives
\begin{equation}
 q_\infty=\sigma_\rho^2A_\infty^2
    \int_0^T\exp\left(-2\int_0^t a_\rho(s)\dd s\right)\dd t>0.
 \label{eq:qinf-positive-proof}
\end{equation}
Therefore $\mathcal C$ is smooth near $A_\infty$, and
\begin{equation}
 \mathcal C'(A)=4s_0^2(A-1)
       +4s_0^2A
       \left(
       \frac{\sqrt q}{\sqrt{v_T-s_0^2A^2}}-1
       \right)>0
 \label{eq:C-derivative-proof}
\end{equation}
for $1<A\le A_\infty$. It follows that the exact-kernel squared cost is
\begin{equation}
 \mathcal C_{\rho,S}
 =\mathcal C_{\rho,\infty}
 -\Lambda_\rho\delta+O(\delta^2),\qquad
 \Lambda_\rho=A_\infty d_\rho\mathcal C'(A_\infty)>0.
 \label{eq:C-expansion-proof}
\end{equation}

It remains to show that fitting the polynomial rotation does not change
this first-order coefficient. Identifying $uI+vJ$ with $u+iv$, Taylor
expansion near one gives
\[
 \log\bigl(j_0(\omega\delta)+3ij_1(\omega\delta)c\bigr)
 =i\omega\delta c
 +\tfrac12\omega^2\delta^2(c^2-1/3)
 +O(\delta^3).
\]
Multiplying by $S=T/\delta$ and exponentiating gives
\begin{align}
 M_{\omega\delta}(c)^S
 &=R(\Omega c)[I+\delta\alpha(c)I+O(\delta^2)],
 &\alpha(c)&=\tfrac12\omega^2T(c^2-1/3),\notag\\
 d_{\omega\delta}(c)
 &=1+\omega^2\delta^2(c^2-1/3)+O(\delta^4),
 &\E_c\alpha(c)&=0.
 \label{eq:rotation-expansion-proof}
\end{align}
The contribution to the mean risk linear in $\delta$ is a scalar
multiple of $\E_c\alpha(c)$, so it vanishes.

For the covariance, define nonnegative weights
$w_i=q_i\prod_{j=i+1}^{S-1}r_j^2$. They satisfy
$\sum_iw_i=\widetilde q_S\le q$, and
\eqref{eq:rotation-ou-learned-covariance} gives
\[
 \widehat q_S(c)=
 \sum_iw_i d_{\omega\delta}(c)^{S-1-i}
 =
 \widetilde q_S
 +\omega^2\delta^2(c^2-1/3)
 \sum_i(S-1-i)w_i
 +O(\delta^2).
\]
The displayed correction is $O(\delta)$ and has mean zero in $c$.
The remainder is uniform, since $S\delta=T$ and $\sum_iw_i\le q$.
As $\widetilde q_S\to q_\infty>0$, the function
$w\mapsto2(\sqrt q-\sqrt w)^2$ has bounded second derivative in a
neighborhood of all these covariances for large $S$. Taylor expansion
and the zero-mean correction therefore give an expected change
$O(\delta^2)$. The same conclusion for the mean term was just proved.
Consequently
\begin{equation}
 \mathcal E_S(\rho)=\mathcal C_{\rho,S}+O(\delta^2).
 \label{eq:risk-versus-cost-proof}
\end{equation}
Combining \eqref{eq:C-expansion-proof} and
\eqref{eq:risk-versus-cost-proof} proves the second claim. This step is a
calculation from the fitted endpoint law, not an assumption that learning
and memory errors add.

Finally $a_\rho=O(\rho)$ uniformly on $[0,T]$ as $\rho\downarrow0$,
since $v_t\ge s_0^2$. Thus $A_\infty=1+O(\rho)$ and
$\mathcal C_{\rho,\infty}=O(\rho^2)\to0$. The strictly positive direct
risk \eqref{eq:rotation-ou-direct-error} is independent of $\rho$.
Choose $\rho_*>0$ so that
$\mathcal C_{\rho,\infty}<\mathcal E_1$ for
$0<\rho<\rho_*$. Part (ii) supplies a sufficiently large finite
$S_0$ with
\[
 \mathcal E_{S_0}(\rho)
 <
 \mathcal C_{\rho,\infty}
 <
 \mathcal E_1(\rho).
\]
Because $\mathcal E_S(\rho)\to\mathcal C_{\rho,\infty}$, every
sufficiently large index has risk strictly larger than
$\mathcal E_{S_0}(\rho)$. The minimum is therefore attained in a finite
set of depths and cannot occur at one. This proves the final claim.
\end{proof}

\textbf{Joint expansion and optimal cascade depth.}

The fixed-$\rho$ remainder above is insufficient to optimize jointly over
$S$ and $\rho$. We first justify a uniform second-order expansion directly
from the finite recurrences that define the learned endpoint law.

\begin{lemma}[Expansion of a smooth affine recurrence]
\label{lem:recurrence-expansion}
Let $\theta$ range over a compact parameter set. Suppose the scalar real
or complex functions $F(t,h,\theta)$ and $G(t,h,\theta)$ are smooth near
$[0,T]\times\{0\}$, uniformly in $\theta$, with
$F(t,0,\theta)=1$ and $G(t,0,\theta)=0$. For $h=T/S$, let
\[
 y_{i+1}=F(ih,h,\theta)y_i+G(ih,h,\theta),
 \qquad y_0=y_{\rm init}(\theta),
\]
where the initial value is smooth. Then smooth functions
$y^{[j]}(t,\theta)$, $j=0,1,2$, exist such that
\[
 y_S=y^{[0]}(T,\theta)+h y^{[1]}(T,\theta)
                           +h^2y^{[2]}(T,\theta)+O(h^3),
\]
with a remainder uniform in $\theta$.
\end{lemma}

\begin{proof}
Expand
$F=1+hf_1+h^2f_2+h^3f_3+O(h^4)$ and
$G=hg_1+h^2g_2+h^3g_3+O(h^4)$, uniformly on the compact set.
Writing $y_j=y^{[j]}$, choose the functions by the linear equations
\begin{align*}
 y_0'&=f_1y_0+g_1,\\
 y_1'&=f_1y_1+f_2y_0+g_2-\tfrac12y_0'',\\
 y_2'&=f_1y_2+f_2y_1+f_3y_0+g_3-\tfrac12y_1''-\tfrac16y_0''',
\end{align*}
with $y_0(0)=y_{\rm init}$ and $y_1(0)=y_2(0)=0$. Successive solution
gives smooth functions with bounded derivatives on the compact set.
Taylor expansion shows that $y_0(t)+hy_1(t)+h^2y_2(t)$ satisfies one
step of the recurrence with defect $O(h^4)$. Since $|F|\le1+Ch$, the
terminal defect is bounded by
\[
Ch^4\sum_{i=0}^{S-1}(1+Ch)^i=O(h^3),
\]
uniformly in $\theta$.
\end{proof}

\begin{lemma}[Joint small-memory and fine-cascade expansion]
\label{lem:joint-risk}
For some fixed $0<\rho_0<1$ and all sufficiently large $S$, uniformly for
$0\le\rho\le\rho_0$ and $h=T/S$,
\begin{equation}
 \mathcal E_S(\rho)=\mathcal C_{\rho,\infty}
       -\lambda_2\rho^2h+B_{\rm fit}h^2
       +O(\rho^3h+\rho h^2+h^3),
 \label{eq:joint-risk-proof}
\end{equation}
with the strictly positive constants in
\eqref{eq:rotation-ou-scaling-constants}.
\end{lemma}

\begin{proof}
Write $a=a_\rho$ and
\[
r(t,h)=1+a(t)(1-e^{-\kappa h})/\kappa.
\]
The complex mean multiplier obeys the recurrence in
Lemma~\ref{lem:recurrence-expansion} with
\[
 F=r(t,h)[j_0(\omega h)+3ij_1(\omega h)c],\quad
 G=0,\quad
 y_{\rm init}=1.
\]
The covariance obeys it with
\[
 F=r(t,h)^2d_{\omega h}(c),\quad
 G=v_{t+h}-r(t,h)^2v_t,\quad
 y_{\rm init}=0.
\]
These coefficients are smooth jointly in $(t,h,\rho,c)$, with removable
singularities of $j_0,j_1$ filled in at zero. Their leading endpoint
values are $A_\infty e^{i\Omega c}$ and $q_\infty$, respectively, because
the leading differential equations are
$y'=(a+i\omega c)y$ and
$w'=2aw+\sigma_\rho^2$, with the stated initial values.

On the compact set $\rho\in[0,\rho_0]$, $q_\infty$ is bounded below by a
positive constant by \eqref{eq:qinf-positive-proof}. Thus the square-root
cost is smooth uniformly near the limiting covariances. Applying the
lemma to both recurrences, substituting in
\eqref{eq:rotation-ou-endpoint-error}, and integrating over $c$ yields
\begin{equation}
 \mathcal E_S(\rho)
 =
 \mathcal C_{\rho,\infty}
 -\Lambda_\rho h
 +B(\rho)h^2
 +O(h^3),
 \label{eq:smooth-risk-expansion-proof}
\end{equation}
where $B$ is smooth and the remainder is uniform. The already calculated
first-order coefficient is $-\Lambda_\rho$ for $\rho>0$, and continuity
extends that identity to zero. This deduction uses the endpoint
recurrences, not an assumed decomposition of squared errors.

To calculate $B(0)$, use $r_i=1$, $q_i=q/S$, and set
\[
\alpha(c)=\omega^2T(c^2-1/3)/2.
\]
Uniformly in $c$,
\begin{align*}
 M_{\omega h}(c)^S&=
 R(\Omega c)[I+h\alpha(c)I+O(h^2)],\\
 \widehat q_S(c)&=
 \frac qS\sum_{j=0}^{S-1}d_{\omega h}(c)^j
 =
 q[1+h\alpha(c)]+O(h^2).
\end{align*}
The conditional mean contributes
$2s_0^2h^2\alpha(c)^2+O(h^3)$ to the risk.
The covariance contributes
$(q/2)h^2\alpha(c)^2+O(h^3)$. Since
$\E_c(c^2-1/3)^2=4/45$,
\[
 B(0)
 =
 (2s_0^2+q/2)\E_c\alpha(c)^2
 =
 \frac{\omega^4T^2(4s_0^2+q)}{90}
 =
 B_{\rm fit}.
\]
Smoothness gives $B(\rho)=B_{\rm fit}+O(\rho)$.

Finally, uniformly for $t\in[0,T]$,
$a_\rho(t)=\rho u(t)+O(\rho^2)$, so
\[
 A_\infty=1+\rho U+O(\rho^2),
 \qquad
 d_\rho=\rho D+O(\rho^2).
\]
Writing $A=1+\epsilon$ in the definition of $\mathcal C(A)$ above gives
\[
 \mathcal C(1+\epsilon)
 =
 2s_0^2(1+s_0^2/q)\epsilon^2
 +O(\epsilon^3).
\]
Therefore
\[
 \Lambda_\rho
 =
 A_\infty d_\rho\mathcal C'(A_\infty)
 =
 4s_0^2(1+s_0^2/q)UD\,\rho^2
 +O(\rho^3)
 =
 \lambda_2\rho^2+O(\rho^3).
\]
Substitution in \eqref{eq:smooth-risk-expansion-proof} proves
\eqref{eq:rotation-ou-joint-expansion}. Positivity follows from
$u(t)>0$ for $t>0$ and $\omega\ne0$.
\end{proof}

\begin{proof}[Proof of Theorem~\ref{thm:rotation-ou-scaling}]
Theorem~\ref{thm:rotation-ou-interior} gives a finite global minimizer for
every sufficiently small positive $\rho$. Let
\[
x_\star=\lambda_2/(2B_{\rm fit})>0
\]
and choose an integer $S_\rho$ nearest to
$T/(x_\star\rho^2)$. Then
$T/S_\rho=x_\star\rho^2+O(\rho^4)$, and
\eqref{eq:rotation-ou-joint-expansion} gives
\begin{equation}
 \mathcal E_{S_\rho}(\rho)-\mathcal C_{\rho,\infty}
       =
 -\frac{\lambda_2^2}{4B_{\rm fit}}\rho^4
 +o(\rho^4).
 \label{eq:optimizer-candidate-proof}
\end{equation}
In particular this candidate has strictly smaller risk than the
infinitely deep cascade limit.

Every sequence of global minimizers satisfies
$S^\star(\rho)\to\infty$. Otherwise a subsequence has one fixed finite
index $S$, while the candidate risk tends to zero and
$\mathcal E_S(\rho)\to\mathcal E_S(0)>0$.
The last strict inequality follows already from the mean term:
\[
\left(
j_0(\omega T/S)+3ij_1(\omega T/S)c
\right)^S
\]
is a finite polynomial in $c$ and cannot equal
$e^{i\Omega c}$ on an interval for $\Omega\ne0$.

Put $h^\star=T/S^\star$. The uniform expansion and
$h^\star\to0$ imply, for sufficiently small $\rho$,
\[
 \mathcal E_{S^\star}(\rho)-\mathcal C_{\rho,\infty}
 \ge
 -C\rho^2h^\star
 +\tfrac12B_{\rm fit}(h^\star)^2.
\]
The left side is negative by \eqref{eq:optimizer-candidate-proof}, so
$h^\star\le C'\rho^2$. For bounded $x=h/\rho^2$, the same expansion is
uniformly
\[
 \frac{
 \mathcal E_S(\rho)-\mathcal C_{\rho,\infty}
 }{\rho^4}
 =
 -\lambda_2x+B_{\rm fit}x^2+o(1).
\]
The quadratic has the unique minimum at $x_\star$. Comparison with
\eqref{eq:optimizer-candidate-proof} forces every minimizing sequence to
satisfy
\[
\frac{h^\star}{\rho^2}\to x_\star,
\]
proving the cascade-depth limit
\eqref{eq:rotation-ou-optimal-segments}. Evaluation of that minimum gives
\eqref{eq:rotation-ou-optimal-error}.
\end{proof}

\subsection{Proof of the anchoring comparison}
\label{app:anchored-ou-proof}

\begin{proof}[Proof of Theorem~\ref{thm:anchored-ou}]
Write $R_t(c)=R(\omega ct)$ and suppress $c$ when unambiguous.
Conditionally on $c$, the rotating coordinates satisfy
\[
R_t^{-1}Y_t=Y_0+U_t,
\qquad
U_t=\beta_\rho\int_0^tH_r\,\dd r
      +\sigma_\rho\widetilde W_t.
\]
Here $U$ is centered Gaussian and independent of $Y_0$, with covariance
$w_tI_2$. For $s>0$ and $h>0$, the anchored conditional is Gaussian with
mean $m_{s,h}^{\mathrm{anc}}$ and covariance $\eta_{s,h}I_2$, where
\begin{align}
\alpha_{s,h}
&=1+\lambda_s\frac{1-e^{-\kappa h}}{\kappa},\notag\\
\eta_{s,h}&=w_{s+h}-\alpha_{s,h}^2w_s,\notag\\
m_{s,h}^{\mathrm{anc}}(c,y,a)
&=\alpha_{s,h}R_hy+(1-\alpha_{s,h})R_{s+h}a.
\label{eq:anchored-ou-exact-step}
\end{align}
The first kernel is $\N(R_ha,w_hI_2)$. If an exact anchored cascade has
mean $R_sa$ and covariance $w_sI_2$, the next kernel has mean
$\alpha_{s,h}R_hR_sa+(1-\alpha_{s,h})R_{s+h}a=R_{s+h}a$ and covariance
$(\alpha_{s,h}^2w_s+\eta_{s,h})I_2=w_{s+h}I_2$. Induction proves the
first claim. When $\omega=0$, all these means are affine in $(y,a)$.

We next identify the limit of the quadratic fits. Let $\Pi_t$ denote
least-squares projection under the true law of $(c,Y_t,Y_0)$. Centering,
rotational symmetry, and the symmetry under simultaneous reflection of
the vectors and replacement of $c$ by $-c$ give
\begin{equation}
\Pi_t\bigl[y-R_ta\bigr]=P_t(c)y+Q_t(c)a,
\qquad
P_t(c)=u_tI_2+\zeta_tcJ,\quad
Q_t(c)=d_tI_2+e_tcJ.
\label{eq:anchored-ou-residual-fit}
\end{equation}
Set $\theta=\omega t$ and
\[
j_0=j_0(\theta),\qquad j_1=j_1(\theta),\qquad
m_2=\E_c[c^2\cos(\theta c)].
\]
Eliminating the coefficients multiplying $a$ in the normal equations gives
\begin{gather}
d_t=-j_0u_t+j_1\zeta_t,
\qquad e_t=-3j_1u_t-3m_2\zeta_t,\notag\\
\begin{pmatrix}
w_t+s_0^2h_0&s_0^2h_1\\
s_0^2h_1&w_t/3+s_0^2h_2
\end{pmatrix}
\begin{pmatrix}u_t\\\zeta_t\end{pmatrix}
=\begin{pmatrix}w_t\\0\end{pmatrix},
\label{eq:anchored-ou-normal-equations}\\
h_0=1-j_0^2-3j_1^2,\qquad
h_1=j_1(j_0-3m_2),\qquad
h_2=\frac13-j_1^2-3m_2^2.\notag
\end{gather}
Indeed, in complex notation, let $\mathsf P$ be the $L^2([-1,1],\dd c/2)$
projection onto the real span of $1$ and $ic$. Then $(h_{jk})_{j,k=0}^1$,
with entries $h_{00}=h_0$, $h_{01}=h_1$, and $h_{11}=h_2$, is the real
Gram matrix of
\[
(I-\mathsf P)e^{i\theta c},
\qquad (I-\mathsf P)(ic e^{i\theta c}).
\]
Thus $h_1^2\leq h_0h_2$. Also $0<h_0<1$ when $\theta\ne0$:
$e^{i\theta c}$ is not affine in $c$, and $j_0(\theta)$ and $j_1(\theta)$
cannot both vanish. Solving~\eqref{eq:anchored-ou-normal-equations} yields
\[
u_t=\frac{w_t}{w_t+s_0^2\chi_t},
\qquad
\chi_t=h_0-\frac{s_0^2h_1^2}{w_t/3+s_0^2h_2}.
\]
The Gram inequality gives
$0<h_0w_t/(w_t+3s_0^2h_2)\leq\chi_t\leq h_0<1$. Hence
\begin{equation}
\frac{w_t}{s_0^2+w_t}<u_t<1\qquad(t>0,\ \omega\ne0).
\label{eq:anchored-ou-coefficient-bounds}
\end{equation}

Expanding~\eqref{eq:anchored-ou-exact-step} and projecting its mean gives
\begin{align*}
\Pi_t m_{t,h}^{\mathrm{anc}}(c,y,a)
&=y+h\bigl[\omega cJy+\lambda_t(P_t(c)y+Q_t(c)a)\bigr]
  +O(h^2)(|y|+|a|),\\
\eta_{t,h}&=\sigma_\rho^2h+O(h^2).
\end{align*}
The coefficient bounds in these expansions are uniform for
$h\leq t\leq T-h$. To check the endpoint $t\downarrow0$, note that
$w_t\geq\sigma_\rho^2t$, while the two residual functions defining the
Gram matrix have norms $O(t^2)$ and $O(t)$. After elimination of the
coefficients multiplying $a$, the normal-equation matrix is bounded
below by $(w_t/3)I_2$. The remainder in its right-hand side is
$O(h^2)(w_t+t)$, so solving for the coefficients preserves the uniform
$O(h^2)$ remainder. The coefficients extend smoothly to
$P_0=I_2$, $Q_0=-I_2$. The first fitted kernel separately has mean
$a+h\omega cJa+O(h^2)|a|$ and covariance
$(\sigma_\rho^2h+O(h^2))I_2$.

The Gaussian mean and variance recursions are therefore consistent Euler
recursions, with uniformly bounded coefficients, for
\begin{align}
\dot F_t(c)
&=[\omega cJ+\lambda_tP_t(c)]F_t(c)+\lambda_tQ_t(c),
\notag\\
\dot z_{\mathrm{anc}}(t)
&=2\lambda_tu_tz_{\mathrm{anc}}(t)+\sigma_\rho^2,
\label{eq:anchored-ou-limit-equations}
\end{align}
with $F_0(c)=I_2$ and $z_{\mathrm{anc}}(0)=0$.
Discrete Gronwall gives convergence uniformly in $c$. In the variance
recursion, the coefficient of $cJ$ contributes only at order $h^2$;
thus the limiting variance is independent of $c$. The limiting risk is
\begin{equation}
\begin{aligned}
\mathcal C_{\rho,\infty}^{\mathrm{anc}}(\omega)
&=s_0^2\E_c\|F_T(c)-R_T(c)\|_F^2\\
&\quad+2\bigl(\sqrt q-\sqrt{z_{\mathrm{anc}}(T)}\bigr)^2.
\end{aligned}
\label{eq:anchored-ou-limit-risk}
\end{equation}
Since $\dot w_t=2k_t+\sigma_\rho^2=2\lambda_tw_t+\sigma_\rho^2$,
the strict coefficient inequalities
\[
\frac{k_t}{s_0^2+w_t}<\lambda_tu_t<\lambda_t
\]
and the variation-of-constants formula give
$z_{\mathrm{un}}(T)<z_{\mathrm{anc}}(T)<w_T=q$.
Equation~\eqref{eq:anchored-ou-limit-risk} then proves
\eqref{eq:anchored-ou-positive-limit}.

Finally, as $\omega\to0$, the normal equations give, uniformly for
$t\in[0,T]$,
\[
u_t=1+O(\omega^4),\quad \zeta_t=O(\omega^3),\quad
d_t=-1+\frac{\omega^2t^2}{6}+O(\omega^4),\quad
e_t=-\omega t+O(\omega^3).
\]
These estimates also hold at $t=0$ because $w_t\geq\sigma_\rho^2t$.
Expanding $R_t$ therefore gives
\[
P_t(c)R_t(c)+Q_t(c)
=-\frac{\omega^2t^2}{2}\left(c^2-\frac13\right)I_2
 +O(\omega^3).
\]
Subtracting $\dot R_t=\omega cJR_t$ from the first equation
in~\eqref{eq:anchored-ou-limit-equations} and using variation of constants
yields
\[
F_t(c)-R_t(c)
=-\frac{\omega^2D_t}{2}\left(c^2-\frac13\right)I_2
 +O(\omega^3).
\]
The variance equation gives $z_{\mathrm{anc}}(T)=q+O(\omega^4)$, so its
contribution to~\eqref{eq:anchored-ou-limit-risk} is $O(\omega^8)$.
Using $\E_c(c^2-1/3)^2=4/45$ and $\|I_2\|_F^2=2$ gives the leading
term $2s_0^2D_T^2\omega^4/45$. The coefficients are smooth in $\omega$,
and reflection symmetry makes the risk even in $\omega$, giving the
$O(\omega^6)$ remainder. The unanchored limit
$\mathcal C_{\rho,\infty}>0$ is independent of $\omega$, whereas the
anchored limit tends to zero. This proves the final comparison.
\end{proof}
\end{document}